\documentclass[10pt, sigconf]{acmart}
\renewcommand\footnotetextcopyrightpermission[1]{} 
\setcopyright{none}

 \pdfoutput=1
\include{psfig}  
\usepackage{epsfig}
\usepackage{algorithm }
\usepackage{xspace}
\usepackage{amsmath}
\usepackage{algpseudocode}
\usepackage{float}
\usepackage{graphicx}
\usepackage[update,prepend]{epstopdf}
\usepackage{flushend}
\usepackage{color}
\usepackage{url}
\usepackage{array}
\usepackage{subfig}
\usepackage{enumitem}
\usepackage{wrapfig}
\usepackage{multirow}
\usepackage[utf8]{inputenc}

\floatstyle{plain}
\newfloat{deneme}{t!}

\newcommand \be {\begin{equation}}
\newcommand \ee {\end{equation}}

\algnewcommand{\LineComment}[1]{\State \(\triangleright\) #1}

\newcommand{\system}{\textsc{RoVaR}\xspace}

\begin{document}

\hyphenation{thro-ugh-put o-f-d-m-a plan-ned program-mable draf-ted wi-max two-di-men-sion-al para-digm}

\title{{\LARGE \system}: Robust Multi-agent Tracking through Dual-layer Diversity in Visual and RF Sensor Fusion}


\author{\Large Mallesham Dasari$^{1, *, \dagger}$, Ramanujan K Sheshadri$^2$, Karthikeyan Sundaresan$^{3, *}$, Samir R. Das$^1$}
\affiliation{
  \institution{$^1$Stony Brook University, $^2$NEC Labs America, $^3$Georgia Tech}
  \authornote{Work done while the authors were at NEC Labs America. \\ $\dagger$Current affiliation is Carnegie Mellon University.}
}




%


\begin{abstract}
The plethora of sensors in our commodity devices provides a rich substrate
for sensor-fused tracking. Yet, today's solutions are unable to deliver robust and high tracking
accuracies across multiple agents in practical, everyday environments -- a feature central 
to the future of immersive and collaborative applications.
This can be attributed to the limited scope of diversity leveraged by these fusion solutions, 
preventing them from catering to the multiple dimensions of {\em accuracy}, {\em robustness} 
(diverse environmental conditions) and {\em scalability} (multiple agents) simultaneously.

In this work, we take an important step towards this goal by introducing the 
notion of {\em dual-layer diversity} to the problem of sensor fusion in multi-agent tracking. 
We demonstrate that the fusion of complementary tracking modalities, 
-- passive/relative (e.g. visual odometry) and active/absolute tracking (e.g.infrastructure-assisted RF localization)
offer a key first layer of diversity that brings scalability 
while the second layer of diversity lies in the {\em methodology} of fusion, where we 
bring together the complementary strengths of algorithmic (for robustness) and data-driven (for accuracy) approaches. \system is an embodiment of such a dual-layer diversity approach that intelligently {\em attends} to cross-modal information using algorithmic and data-driven techniques that jointly share the burden of accurately tracking multiple agents in the wild. Extensive evaluations reveal \system's multi-dimensional benefits in terms of tracking accuracy (median of $\approx$15cm), robustness (in unseen environments), light weight (runs in real-time on mobile platforms such as Jetson Nano/TX2), to enable practical multi-agent immersive applications in everyday environments. 
\end{abstract}

\pagenumbering{arabic}

\maketitle

\section{Introduction}
\label{INTRO}

Tracking  multiple agents (humans and robots) in real life within a given space is a foundational problem in immersive and interactive interactive applications. The growing availability of visual (cameras), inertial (IMUs), and RF (WiFi, BLE, UWB) sensors on our everyday devices provides a rich substrate for effective tracking. However, tracking solutions that are robust, accurate and cost-effective in realistic environments remain elusive.

\noindent {\bf Limitations of Today's Tracking Solutions.} Existing tracking solutions can be broadly categorized as: passive and active (infrastructure-assisted) tracking. Passive solutions~\cite{campos2021orb, bloesch2015robust, zhang2014loam, solin2018inertial} are largely odometry-based relative tracking methods that commonly use cameras and inertial sensors. While these passive/relative solutions are cost-effective and can track to within few tens of {\em cm} accuracy under favorable conditions, they are significantly vulnerable (1m+ error) when facing everyday environmental conditions like dim-light, texture-less surfaces, etc.~\cite{feigl2020localization, saputra2018visual}. Further, their reliance on  {\em relative} tracking prevents them from robustly recovering from unfavorable events, while also limiting the use 
of multi-agent tracking within one global frame of reference. 
On the other hand, active tracking approaches~\cite{ni2017accurate, palacios2017jade, ibrahim2018verification, ruiz2017comparing} use planned deployment of one or more anchors to locate agents. They enable multi-agent tracking through {\em absolute} localization which also eliminates error accumulation. However, these active/absolute tracking solutions often face a tradeoff in operational range and accuracy, and typically offer lower accuracies than passive tracking. Therefore, most of the current generation commercial solutions (e.g., ARCore~\cite{lanham2018learn} and ARKit~\cite{arkitapple}) continue to adopt passive tracking and face aforementioned limitations when deployed in practice. 
  
\begin{figure}[t]
\centering
      \includegraphics[width=\linewidth]{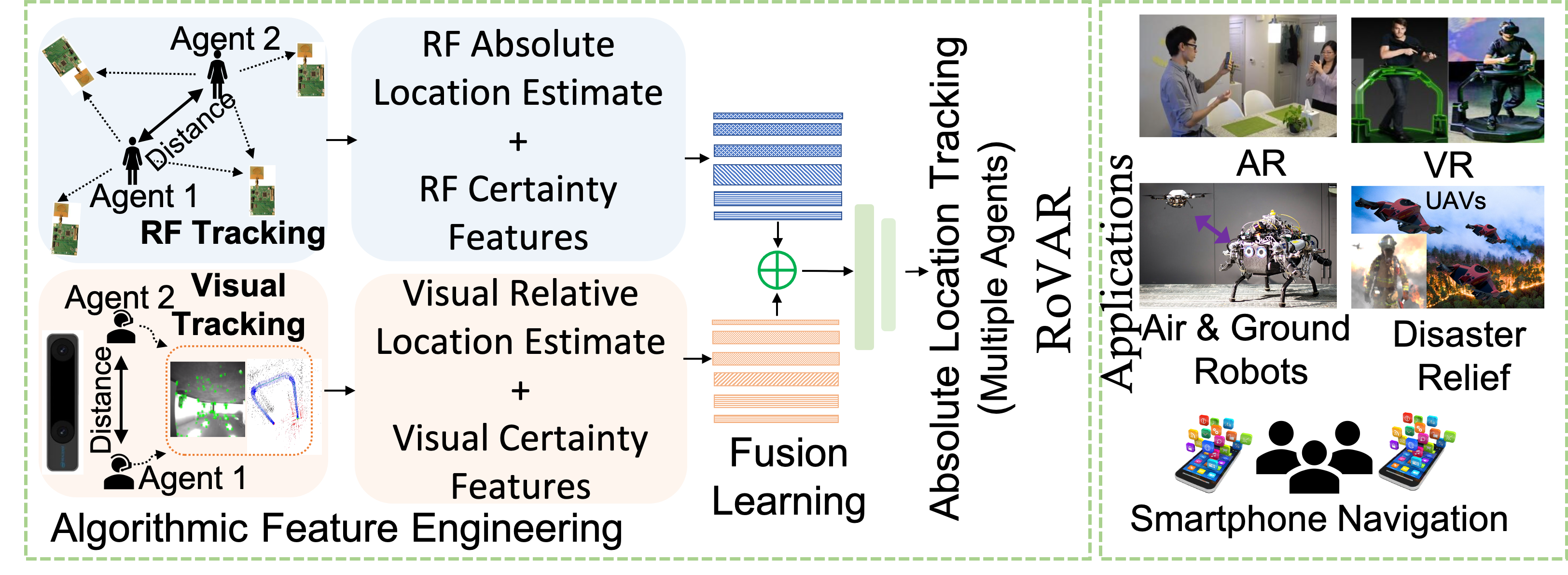}
      \vspace{-0.6cm}
      \caption{Proposed system and its applications.}
     \label{fig:systemaandapplications}
\vspace{-0.5cm}
\end{figure} 
  
\noindent {\bf Need for Dual-layer Diversity.} This work advocates the need for a hybrid tracking approach for immersive and interactive applications, that can effectively bring together the complementary benefits of \underline{\textit{active (absolute, multi-agent)}} and \underline{\textit{passive (relative, high resolution)}} tracking to enable {\em scalable and accurate multi-agent} tracking.\footnote{We consider 2D position tracking in this work 
to focus on the fusion methodology. Details on extending it to 3D tracking can be found in~\cite{tech-report}.} 
While this forms the first layer that leverages 
diversity in the sensor tracking {\em modalities}, this is not sufficient. Conventional
{\em algorithmic} sensor-fusion approaches (e.g. Kalman~\cite{venkatnarayan2019enhancing, corrales2008hybrid, wan2000unscented} and Particle filters~\cite{wang2017imu}) that offer this first layer of diversity are robust to operate in 
various environments in the wild, but are unable to effectively sift out the 
erroneous artifacts of individual sensors and fail to provide effective 
fusion that can deliver high accuracies sustainably. On the other hand, recent {\em data-driven} approaches (e.g. milliEgo~\cite{lu2020milliego}, VINET~\cite{clark2017vinet}) 
show significant promise for much better fusion, however, they have thus far targeted only passive/relative tracking in a small area, with a black-box learning approach that often face nontrivial challenges when deployed in the real-world (\S\ref{sec:dual-layer-fusion}). Hence, the fundamental challenge in realizing our vision for a \underline{\textit{robust, accurate, and scalable}} tracking solution lies in not only leveraging the diversity offered by the two modality of tracking (active and passive), but also the diversity offered by
the methodology of fusion – in particular, figuring out when and how to leverage statistical learning in combination with closed-form algorithms.  

\noindent {\bf \system:} We propose \system (Fig.~\ref{fig:systemaandapplications}): an
accurate and robust multi-agent tracking solution sufficiently lightweight to run 
in real-time on a low-end portable GPU device. 
\system embodies effective fusion through diversity 
at two layers: it brings together the complementary benefits of active (RF - WiFi/UWB) and passive 
(visual) tracking modalities through an intelligent combination of both algorithmic 
and data-driven techniques. 
In \system,  agents carry a tracking device (e.g., headset, smart device) that
houses an embedded camera (either stereo or monocular)
and an RF interface. While the camera enables visual odometry-based relative tracking, the RF interface 
enables absolute tracking by estimating ranges (and angles, AoA) to one or more access points (anchors)
in the environment. RF technologies like WiFi, and more so UWB, offer a good balance between tracking 
resolution (160 MHz with 802.11mc, 500 MHz with UWB) and robust NLoS operation, and are already deployed
in many commodity access points and smartphones (Apple iPhone12, Samsung Galaxy S21, etc.)~\cite{appleu1chip, sammobileuwbchip}. 
While \system employs UWB for RF tracking modality in this work, its approach is equally applicable to 
other RF technologies (e.g., WiFi).

\system leverages algorithmic solutions to estimate the absolute location estimate from RF 
(e.g. multi-lateration~\cite{cheung2006constrained, hua2014geometrical} or 
range+AoA~\cite{herath2013optimal} approaches) and the relative translation estimate from 
visual (e.g. ORB-SLAM3~\cite{campos2021orb}), while data-driven model assists in data filtering,
feature composition and the fusion itself.
\system allows individual sensor algorithms to provide estimates 
along with their certainties to the fusion model. This leverages the physics and geometry inherent to these 
localization problems and also relieves the fusion model from trying to learn
purely from raw data alone, allowing it to focus more  on the effective fusion. 
This also reduces the compute and latency requirements 
enabling \system to run in real-time even on 
resource-constrained devices, compared to compute-intensive deep learning/blackbox approaches. 
More importantly, these pure blackbox approaches are also limited in their ability to generalize. In contrast, \system's approach contributes to 
a robust model that that generalizes well for deployments in untrained environments.    
	
\system is built as a hand-held prototype comprising a UWB beacon and Intel T265 stereo-camera, experimented in diverse environments spanning 4500 sq.ft. area at multiple locations for a total of 232,000 tracked data points. Comprehensive evaluations over diverse settings reveal the importance
of \system's proposed dual-layer diversity framework, where active
RF tracking eliminates the accumulation of errors, while leveraging
the high resolution offered by the visual sensor, offering a median
tracking accuracy of 15 cm. This is in contrast to 40 and 32 cm
accuracy offered by RF and visual tracking solutions respectively
in isolation. \system\ requires small memory footprint of
\texttt{5MB}, 60\% less training data, and a \texttt{5X} latency reduction, \texttt{3X} less power consumption compared to the next best alternative (blackbox solution), enabling real-time operation on mobile platforms like Jetson Nano/TX2~\cite{jetsontx2}. In summary, our contributions in this work are the following:

{\bf (1)} We advocate the need for dual-layer diversity in both {\em tracking modality} 
(active + passive) and {\em fusion methodology} (algorithmic + data-driven) for accurate, 
robust, and scalable tracking in everyday environments (\S\ref{MOTIV}).

{\bf (2)} We design and build \system as an instantiation of our lightweight, yet robust and accurate framework that embodies the 
dual-layer diversity approach to fuse visual and RF sensing (\S\ref{DESIGN}).

{\bf (3)} We comprehensively evaluate and showcase the benefits of \system-style fusion models that enable both robustness and real-time operation\footnote{A real-time demo of \system is given here~\cite{video-link}. We also plan to release our dataset (UWB, Camera and RFID) to foster this line of research in Tracking.} (\S\ref{EVAL}).

\section{Background and Motivation}
\label{MOTIV}
\subsection{Related Work}
\label{sec:background}
{\bf Active/Absolute tracking:}
Active tracking techniques use pre-deployed anchors at known locations to continuously localize and track a device. Popular solutions include RF multi-lateration techniques~\cite{cheung2006constrained, hua2014geometrical} or recent deep learning methods~\cite{ayyalasomayajula2020deep}. There are other works using optical
(e.g., IR beacons~\cite{kumar2014improving, jung2004ubitrack}, 
VLC~\cite{luo2017indoor}) and acoustic sensors ~\cite{chunyi2007, lollmann2018locata, constandache2014daredevil}, however, these solutions are either limited to LoS scenarios or highly sensitive to ambient noise~\cite{jia2014soundloc}, 
material attenuation (e.g., wood, concrete)~\cite{tarzia2011indoor, constandache2014daredevil}, and certain environmental conditions 
(e.g., temperature, humidity) limiting them to a room-level applications.
In comparison, RF signals are more robust in LoS and NLoS indoor environments. Consequently, prior works mainly leverage RF solutions like WiFi~\cite{kotaru2015spotfi}, UWB~\cite{neuhold2019hipr} and mmWave~\cite{zeng2016human, palacios2018communication} for active tracking.\\ 
{\bf Passive/Relative tracking:}
Passive odometry-based solutions do not require pre-deployed infrastructure.
Visual and inertial odometry are perhaps the most popular in this category. 
Visual Odometry (VO) algorithms rely on changes in texture, color and shape 
in successive camera images of a static environment to track the motion with high-precision (cm-scale). Popular VO solutions include ORBSLAM3~\cite{campos2021orb}, and recent deep learning methods (e.g., DeepSLAM~\cite{li2020deepslam}, VINET~\cite{clark2017vinet}). However, majority of these solutions suffer in environments with poor lighting, 
lack of texture or in conditions that enable perceptual aliasing~\cite{song2015high, taketomi2017visual}. 
\\
{\bf Sensor fusion for tracking:}
\label{sec:sensor-fusion}
Sensor fusion has been long studied in the past to compensate for the noise originating from individual sensors. Solutions in this space include either algorithmic (e.g., WiFi+Camera~\cite{hashemifar2019augmenting}, IMU+WiFi~\cite{venkatnarayan2019measuring}) or deep learning methods (e.g., milliEgo~\cite{lu2020milliego}, Camera+IMU~\cite{clark2017vinet,chen2019selective}, DeepTIO~\cite{saputra2020deeptio}).
While the algorithmic approaches are robust to diverse environments but struggle to deliver high
tracking accuracies, the deep learning methods often fail to robustly generalize to untrained environments while offering high accuracy in trained environments (see \S\ref{sec:dual-layer-fusion}). 

\subsection{Choice of Tracking Techniques}
\label{sec:tracking-mod}
Although \system's design is applicable to other RF and visual tracking modalities, we rationalize the following choices.   

\noindent {\bf UWB-based active tracking:}  
UWBs, compared to other RF technologies, offer a good balance between coverage 
(35-40m in LoS and 25-30m in NLoS) and ranging accuracy (Table~\ref{tab:choiceofactivetracking}).
Consequently, there has been a spurt of consumer devices (e.g., IPhone 12, Samsung Galaxy S21 etc.)
using UWB chips for indoor tracking. Future UWB chips~\cite{nxpsinglechip} are expected to offer
multi-antenna (single anchor) solutions that enable both ranging and AoA estimation. \system can easily accommodate WiFi FTM~\cite{ibrahim2018verification} (in sub-6 GHz or mmWave band) as well. \\
\noindent {\bf ORB-SLAM3 for passive tracking:} ORB-SLAM3~\cite{campos2021orb}
is a feature-based SLAM algorithm, robust to motion clutter and can work with monocular,
stereo or RGB-D images. The ORB-features~\cite{rublee2011orb} allow for fast matching across camera frames to enable real-time tracking. 
Previous studies show that ORB-SLAM algorithms can significantly outperform other popular SLAM algorithms in odometry detection~\cite{ragot2019benchmark, campos2021orb}. Nonetheless, in \S\ref{DESIGN} we discuss how \system can just as easily support other SLAM/VO algorithms.
 
\begin{table}[t]
  \vspace{-0.2cm}
  \centering
     \caption{Choice of RF Solutions - Range Vs. Accuracy.}
     \vspace{-0.4cm}
       \scalebox{0.83}{
    \begin{tabular}{|l|c|c|c|c|}
  \hline
  \textbf{ } & \textbf{LTE~\cite{ni2017accurate}} & \textbf{WiFi~\cite{ibrahim2018verification}} & \textbf{UWB~\cite{ruiz2017comparing}} & \textbf{mmWave~\cite{palacios2017jade}} \\
    \hline
    Accuracy & 20m & 5m & 25cm & 1cm \\ \hline
    Range & 1km & 40m & 30m & 10m \\ \hline
  \end{tabular}   \label{tab:choiceofactivetracking}
  }
  \vspace{-0.3cm}
\end{table}

\begin{figure}[!h]
\centering
\vspace{-0.2cm}
        \subfloat[RF-poor ranging in NLOS]{%
      \includegraphics[width=0.43\linewidth]{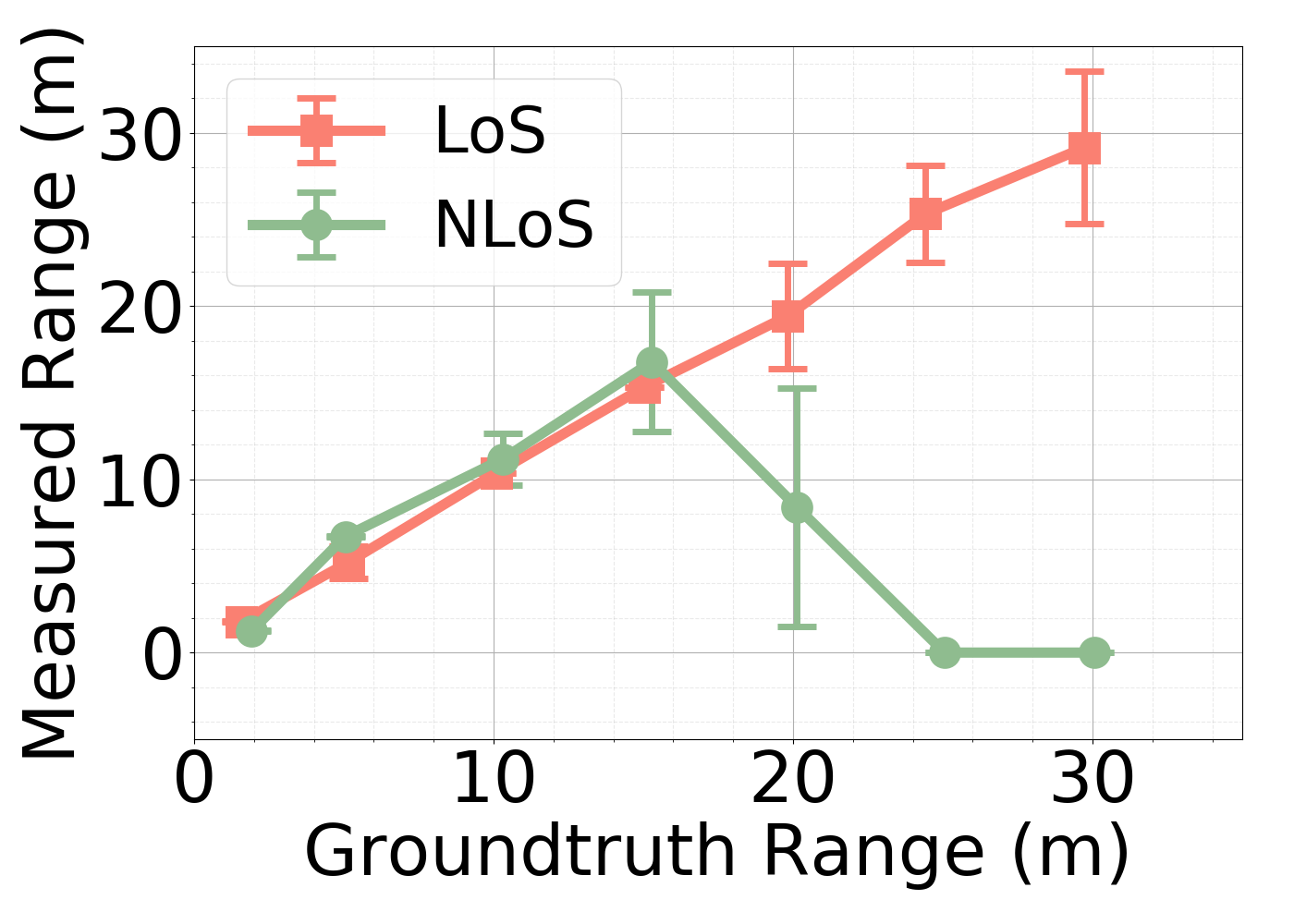}\label{fig:nlosvsranging}
     }
        \subfloat[VO-drifting in dim-light]{%
      \includegraphics[width=0.43\linewidth]{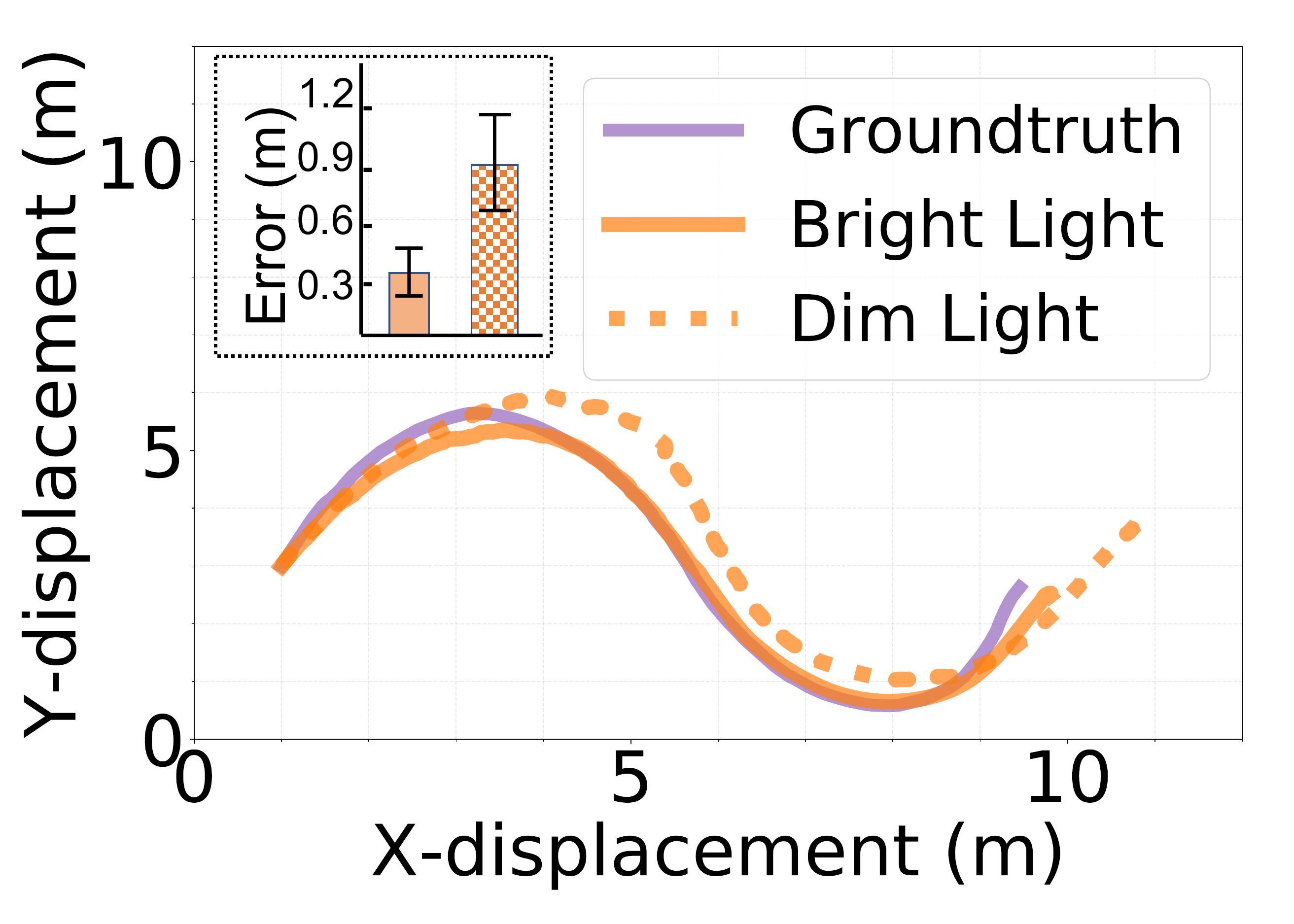} \label{fig:vo-limitations}
     }

    \vspace{-0.3cm}
           \subfloat[Complementary benefits of VO and RF]{
                 \includegraphics[width=0.85\linewidth]{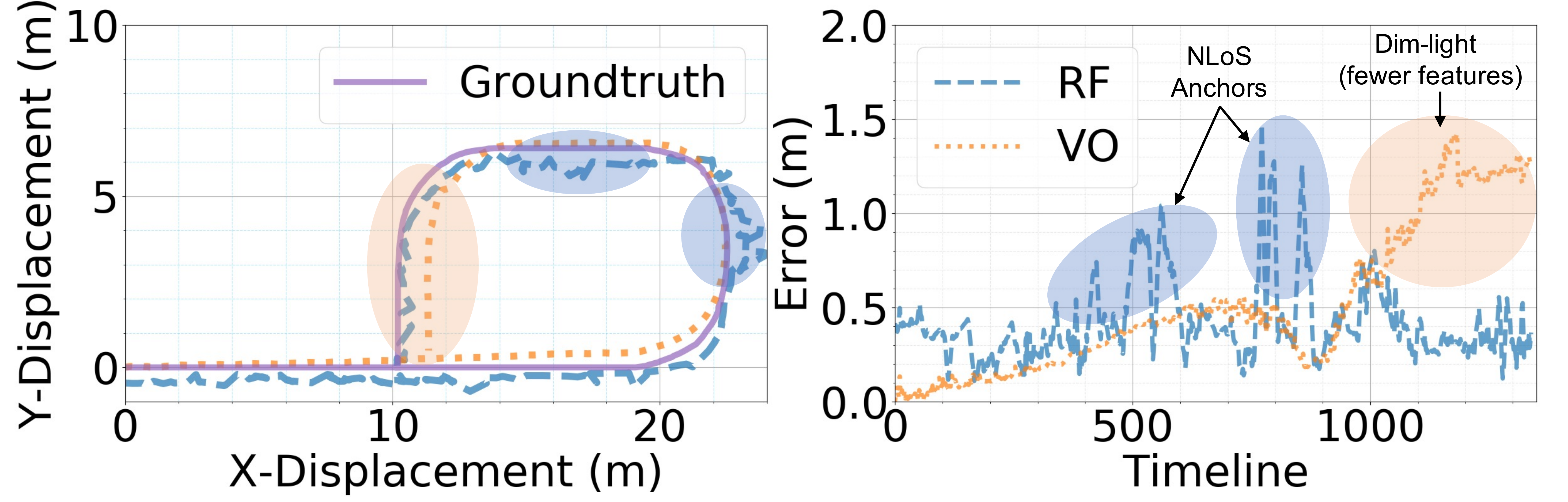} \label{fig:fusioncase}
           }
      \vspace{-0.3cm}
      \caption{Deficiencies of Active and Passive Tracking: a) RF suffering from NLoS conditions, b) VO suffering from poor lighting, c) Example trajectory with groundtruth and tracking error of RF and VO.}
     \label{fig:limitations1}
\vspace{-0.4cm}
\end{figure}

\begin{figure*}
\centering
\vspace{-0.3cm}
      \subfloat[Trained Environment (left: trajectory, right: error)]{%
      \includegraphics[width=0.5\linewidth]{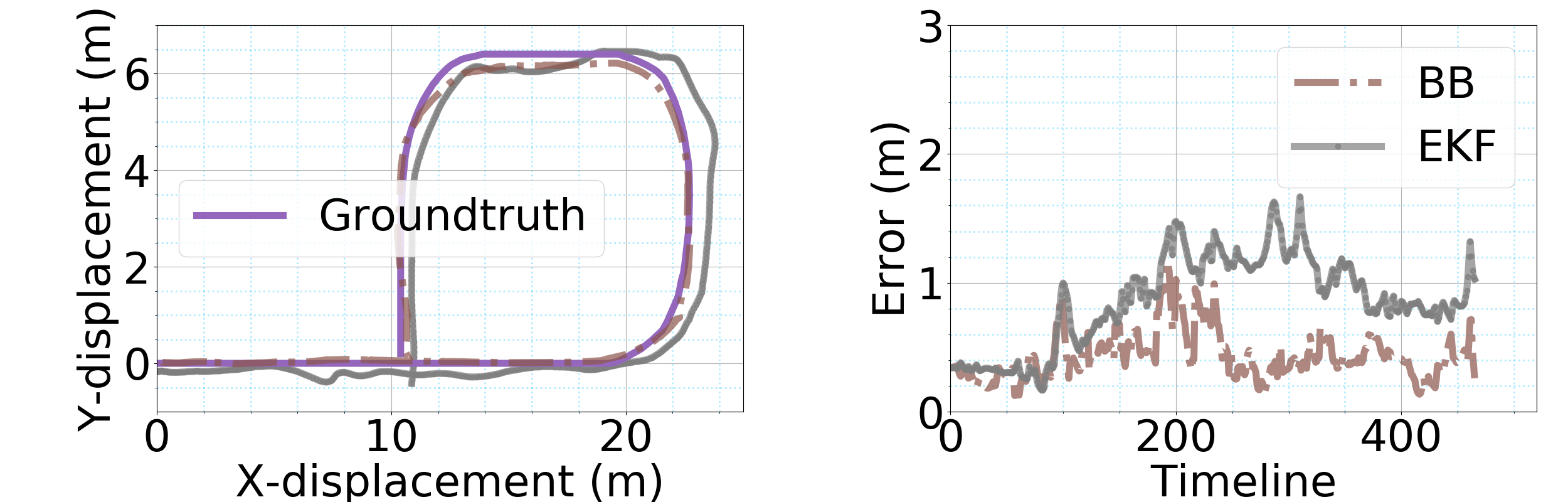} \label{efk-limitation}
     }
    \subfloat[Unseen Environment (left: trajectory, right: error)]{%
      \includegraphics[width=0.5\linewidth]{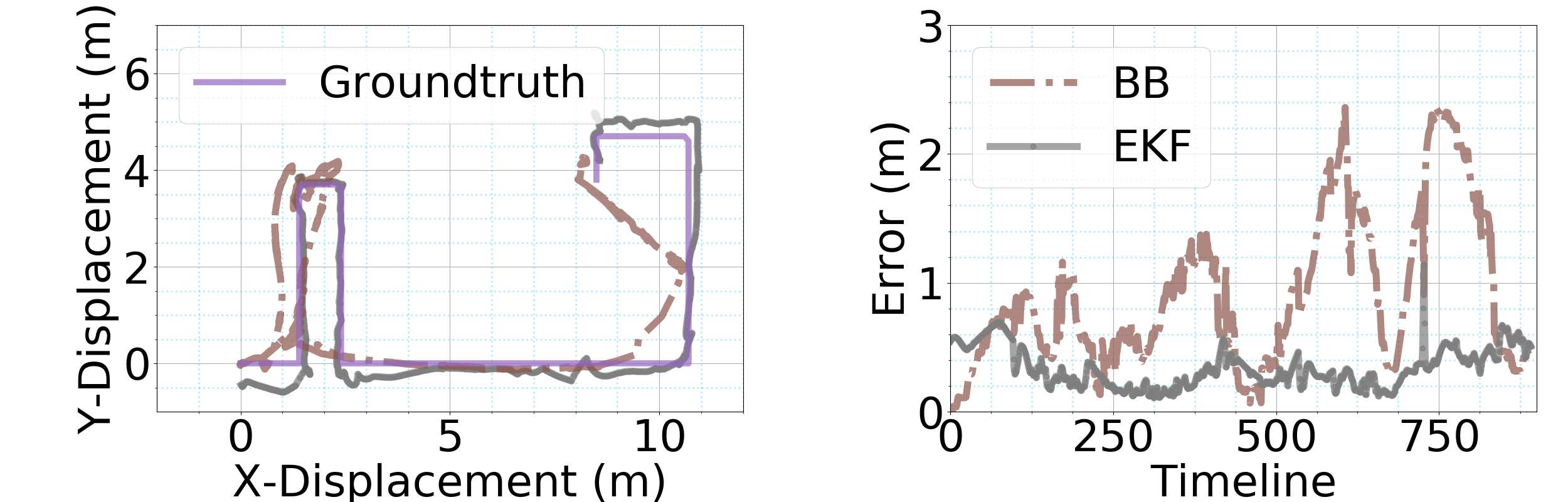}\label{bb-limitation} \label{bb-limitation}
     }
      \vspace{-0.3cm}
      \caption{Limitations of Algorithmic tracking (RF+VO using EKF) and Deep learning based blackbox (BB) solutions. While BB performs better than classical solutions in trained environment, it fails in a new unseen place.}
     \label{fig:limitations2}
\vspace{-0.5cm}
\end{figure*}

\subsection{Limitations of Current Generation Tracking \& Fusion Solutions}
\label{sec:dual-layer-fusion}
\underline{\textbf{1)} \textit{Individual limitations of multiple sensors:}} VO algorithms' {\em relative} tracking -- despite some of its 
intrinsic optimizations (e.g., loop closure detection) and high accuracy -- is vulnerable to 
error accumulation over time. 
Once an error sets in, the relative measurements of the VO propagate the error forward, resulting
in significant drifts in the estimated trajectories (Fig.~\ref{fig:vo-limitations}). UWB-based active
tracking is immune to such drifts as each location estimation is independent of the previous ones.
Errors that occur due to incorrect ranging estimates (e.g., NLoS bad anchors, see Fig. ~\ref{fig:nlosvsranging}) do not propagate, 
allowing UWB to provide better accuracy in {\em absolute} tracking over time. 
However, UWB can only provide sub-meter scale coarse location tracking compared to VO's cm-scale finer
estimates. An effective fusion between the two sensors can bring complementary 
benefits to one-another, delivering a system capable of highly accurate absolute tracking 
over a sustained period of time. Fig~\ref{fig:fusioncase}
shows the potential for fusion between UWB and VO (ORB-SLAM3) for a simple trajectory 
(details are in \S\ref{IMPL}),
part of which is dimly lit (adverse to VO), while another part contains NLoS anchors 
(adverse to UWB)\footnote{We note that there is limited prior work on UWB+VO fusion. This work takes a first attempt to showcase the benefits of fusing VO and UWB sensors.}.

\noindent \underline{\textbf{2)} \textit{Scalability across multiple users:}} Accurately tracking multiple users with respect to one another (even in NLoS) is the key to realizing multi-user collaborative applications.
However, VO algorithms track users relative to their own individual starting point (needing a common origin in case of multiple users). Unless all users start at a common origin and are in visual LoS of each other with similar device/camera capabilities 
(unrealistic expectation), it is infeasible to estimate users' locations with respect to one another. 
In contrast, RF's active tracking provides absolute location estimates of each user with
respect to a known RF anchor's physical location. 
Fusing VO's relative tracking with RF's absolute location estimates can enable accurate tracking in a
global reference point (RF anchor) as well as relative to one-another.
\noindent \underline{\textbf{3)} \textit{Limitations of today's sensor fusion methodologies:}}
Current sensor fusion methods can be broadly divided 
into two classes of solutions: 1) algorithmic or 2) data-driven models.
Algorithmic solutions (e.g., Kalman Filter (KF)~\cite{venkatnarayan2019enhancing} and Bayesian Particle Filters~\cite{wang2017imu}) aim to minimize statistical noise using time series data from individual sensors. 
However, we observe that these approaches consistently under-perform when
fusing UWB and ORB-SLAM3. The KF requires system model to be linearized and noise distribution to be Gaussian, 
which often is not true for VO and RF real-life measurements, either due to sensor hardware and/or 
environmental artifacts (e.g. scene, NLoS issues). As an example, 
Fig. \ref{efk-limitation} shows the performance of
Extended Kalman Filter (EKF) for fusing ORB-SLAM3 and UWB on one of our trajectories (\S\ref{EVAL}). The fused location is
quite erroneous (a median error of \texttt{80\,cm}), leaving a significant room for improvement. On the other hand, Particle filters require a very large number of samples to accurately estimate the posterior probability density of the system. In complex environments, this results in particle depletion and consequently poor performance. 

Deep learning based fusion solutions are shown to be effective in identifying nonlinearities, thus accurately fuse multiple sources of information. Popular recent solutions include fusing passive sensors such as VO+IMU fusion~\cite{clark2017vinet}, mmWave+IMU fusion~\cite{lu2020milliego}. These blackbox (BB) models
are trained on large-scale raw sensor data to obtain relative location or pose estimates. Porting these solutions to fuse raw RF and VO data, to predict {\em absolute} position puts a lot of burden 
on the model in capturing the complex geometric problem structure inherent to RF and camera based localization. This in-turn makes the model heavy,
and overly reliant on input data distributions resulting in 
poor performance in untrained environments. 
Fig. \ref{bb-limitation} shows the performance of a BB model (model details in \S\ref{sec:bbdesign})
that fuses UWB and camera raw data to estimate absolute location in an unseen environment. It is evident that pure deep learning (BB) models, while delivering superior performance in trained environments (see \ref{efk-limitation}), fail to 
generalize when deployed in unseen environments in the wild. More importantly, the problem with deploying these BB models in practice is that extremely slow and energy hungry on mobile devices. Because of these practical limitations, industry-grade solutions~\cite{lanham2018learn, arkitapple} still rely on classical filtering approaches for real-time tracking and energy efficiency, compromising on accuracy\cite{feigl2020localization}. 

In summary, while algorithmic solutions bring robustness to operation in various (untrained) environments but are unable to deliver effective fusion, data-driven approaches deliver 
high accuracy through superior fusion. This motivates the need for a second layer of diversity that brings together the strengths of both algorithmic (generalizability from robustness) and data-driven (high accuracy from effective fusion) approaches, while the first layer of diversity is offered by sensing modalities. \\
\section{Challenges in Realizing \system's Dual-layer Diversity}
Building a fusion approach that incorporates \system's dual-layer diversity entails addressing several critical questions.\\
\noindent \textbf{(1)} Given the pros and cons of active and passive tracking, how should their measurements be fused to automatically overcome sensor hardware and environmental artifacts and deliver both accuracy and scalability?\\
\noindent \textbf{(2)}  How should the algorithmic and data-driven approaches split the burden of the sensor fusion pipeline to complementarily bring together their accuracy and robustness benefits?\\
\noindent \textbf{(3)}  Can the resulting approach be light-weight for real-time operation on resource-constrained device platforms?

\system adopts a systematic approach towards addressing these challenges 
to deliver a \underline{\textit{robust, accurate and scalable}} tracking solution.

\section{{\Large{\system}}: Design}
\label{DESIGN}
\subsection{Overview of \system}
\system incorporates dual-layer diversity by (a) employing an algorithm-driven approach in the 
first stage to engineer features from individual RF (UWB ranges) and Visual (camera frames) 
sensor inputs; (b) followed by a data-driven (machine learning, ML) approach for effective fusion 
of these complementary sensor features through a cross-attention mechanism in the second stage.	

\system's algorithmic component, namely multi-lateration for UWB and ORB-SLAM3 for camera inputs, capture
the physical and environmental dependencies of the localization problem while providing the absolute
and relative position estimates, respectively. This, along with other sensor and environemt-specific
artifacts (RF channel quality, camera inter-frame keypoints, etc) together form the complete set of 
features that are fed into its ML component, which first employs a CNN encoder to extract dependencies
within the individual sensor features to capture the certainty of their respective location estimates.
This is followed by a cross-attention mechanism along with a 2-layer LSTM network (to capture temporal 
dependencies), to effectively fuse these disparate (absolute and relative) location-related features 
and predict the final absolute location estimate of the device with a high accuracy even under challenging 
environmental conditions. Estimating a device's {\em absolute} location estimate allows \system to 
easily scale to multi-agent collaborative applications, where all agents are tracked within a common/global frame of reference. 

Leveraging the initial location estimates from algorithms as features frees the ML module from the 
burden of capturing the physical and geometrical embeddings inherent to localization, 
allowing it to focus solely on sensor fusion. This results in two key benefits: (i) robustness and 
generalizability:  delivering accurate tracking even in unseen challenging environmental conditions, 
and (ii) real-time tracking: significantly reduced end-to-end computations making the entire pipeline
lightweight, enabling it to run in real-time even on resource-constrained mobile platforms.

\subsection{Algorithm-driven Feature Engineering}
The goal of this stage is to generate features that capture not only location estimations of the individual
sensor algorithms, but also the sensor/environmental artifacts that determine the certainty of these estimations.

\subsubsection{RF Module:} \label{sec:rfmodule} UWB sensors use time-of-flight measurement to determine the range $R_i$ between the device and an anchor $i$. Range estimations to at least 3 anchors (or fewer if AoA info. is available) along with the anchors' location information
is required for the multi-lateration algorithms to solve for the absolute location of the device.
Given a set of $n$ anchors at fixed position ($x_i$, $y_i$) with $i=1..n$,
the absolute 2D location of the device ($x, y$) 
can be estimated by minimizing the error $f_i = R_i - \sqrt{(x_i-x)^2+(y_i-y)^2}$ across the anchors. 
Among the many different optimizations that are available~\cite{jayashree2006accuracy, cheung2006constrained, hua2014geometrical},
to solve the multi-lateration problem, \system adopts the least-squares based approach.
%
%

\begin{figure}[!h]
\centering
\vspace{-0.2cm}
    \subfloat[Error in LoS vs. NLoS cases.]{%
    \includegraphics[width=0.5\linewidth]{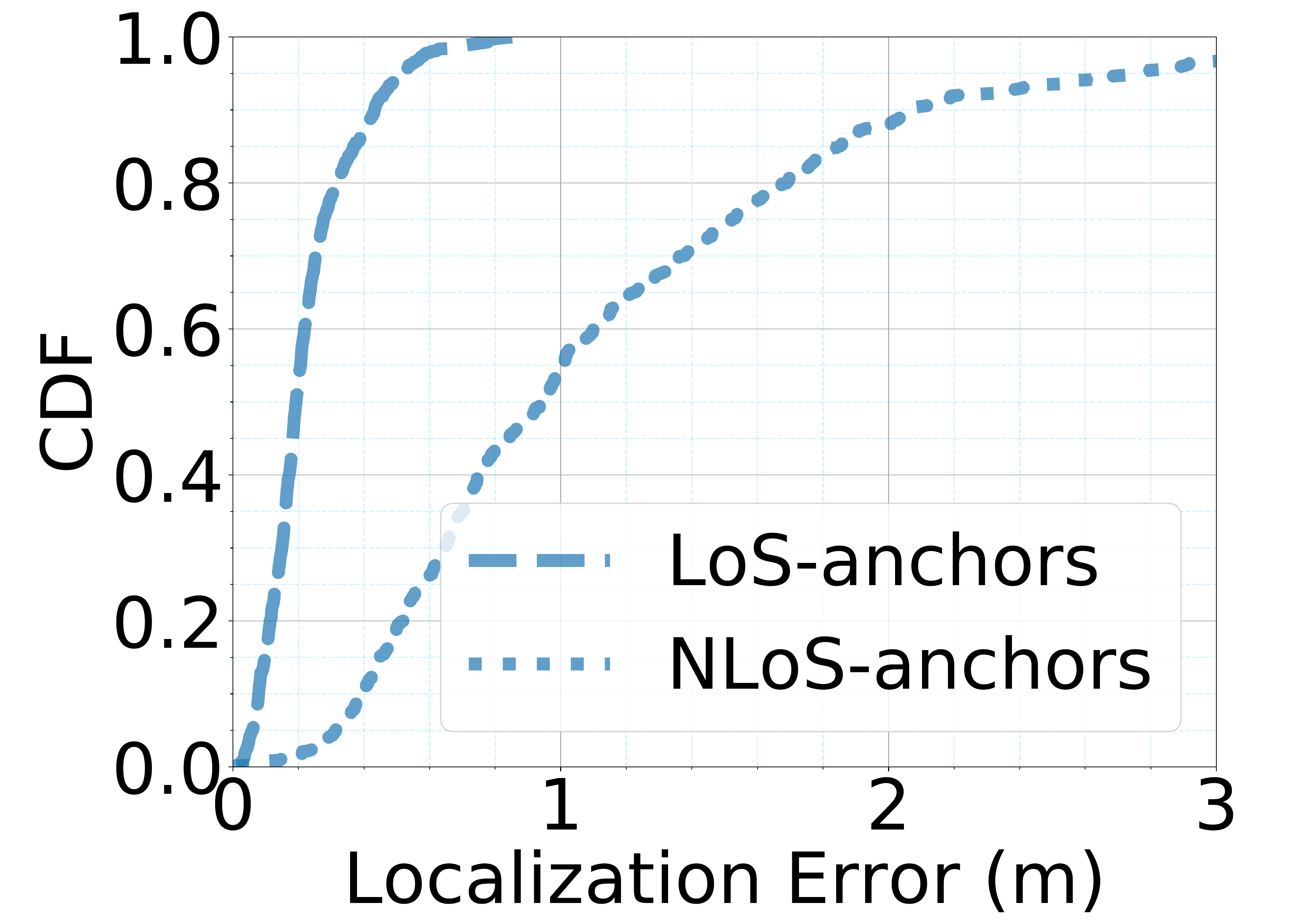} 
    }
    \subfloat[Best case vs. practical error]{%
      \includegraphics[width=0.5\linewidth]{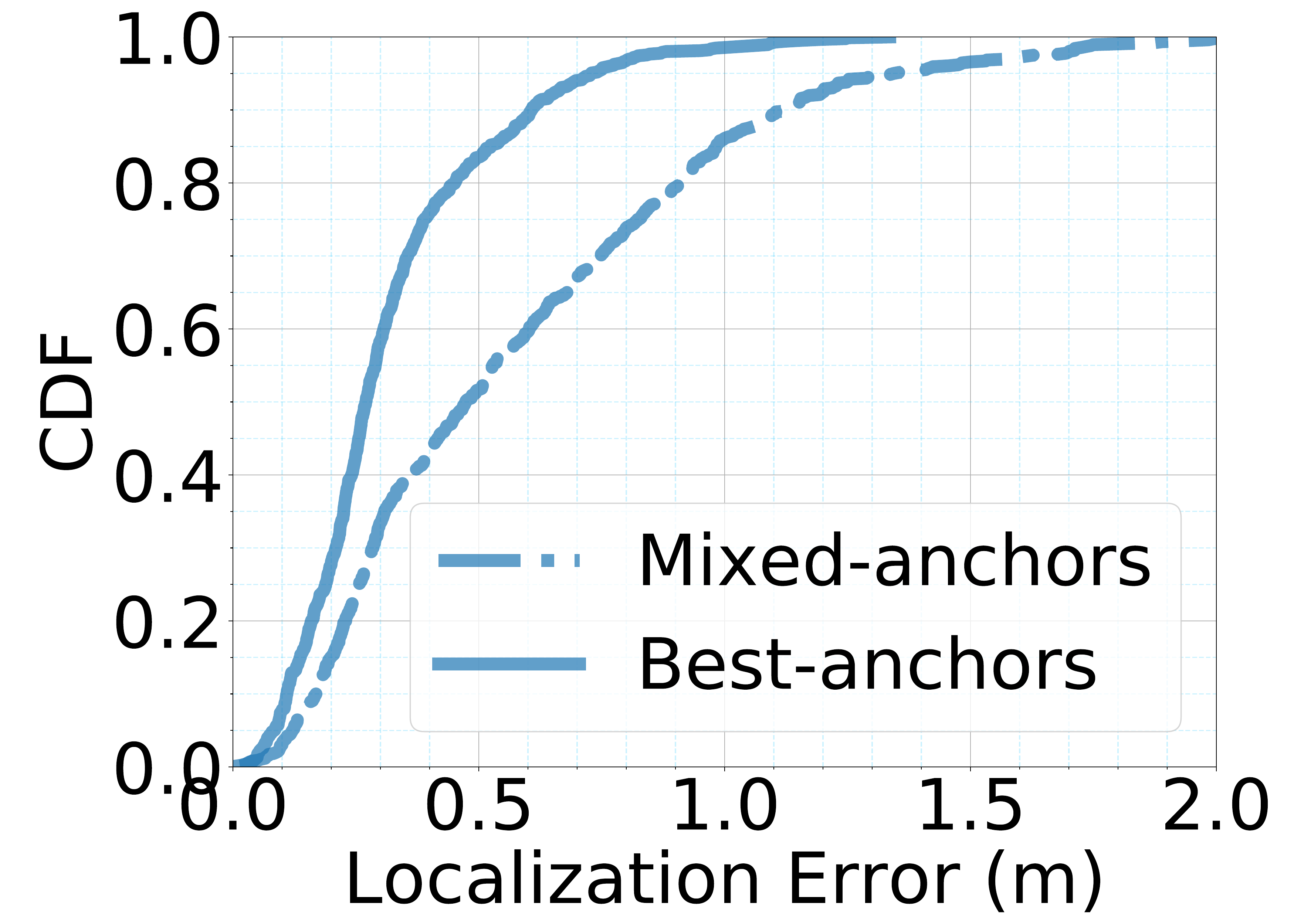} 
     }
    \vspace{-0.2cm}
    \caption{LOS and NLOS Localization errors for UWB.}
    \label{fig:errorvlosnlosrf}
    \vspace{-0.2cm}
\end{figure} 

\noindent \textbf{(1) Extracting features to capture NLoS impact:} Two critical components that impact the localization accuracy are: (i) multi-lateration optimization error, and (ii) environmental conditions (multipath, NLoS) that manifest as inaccurate range estimates (see Fig.~\ref{fig:nlosvsranging}). While the former can be estimated as a direct output of the optimization solution, the latter is 
challenging to address, and requires us to understand the impact of RF propagation and channel characteristics on UWB ranging error.

\noindent{\em \underline{Impact of NLoS anchors on ranging accuracy:}} We study the impact of NLoS anchors on ranging and localization accuracy with five anchors (from our testbed described in \S\ref{IMPL}), as the device to be located
moves in a given trajectory, exposing both LoS and NLoS paths to various anchors. Fig. \ref{fig:errorvlosnlosrf} captures the impact when all anchors are in LoS Vs. when all anchors are in NLoS. A low median error of 0.2m in the case of LoS is amplified to 1m (max. error of 3m) when all anchors are in NLoS. In a more practical setting, when the 5 anchors consist of a {\em Mix} of both LoS and NLoS, this error
is 0.55m. However, selecting the {\em Best} set of 3 (out of 5) anchors (potentially LoS anchors) would
reduce the median error to only 0.25m with the 90th percentile gain of 0.5m over the mixed anchor scheme.
While these suggest that filtering out NLoS anchors can reduce the 
error significantly, selecting the best anchors is nontrivial without knowledge of the ground truth 
environmental conditions, a challenge we address next.

\begin{figure}[t]
\centering
\vspace{-0.3cm}
    \subfloat[Range vs. received power.]{%
      \includegraphics[width=0.5\linewidth]{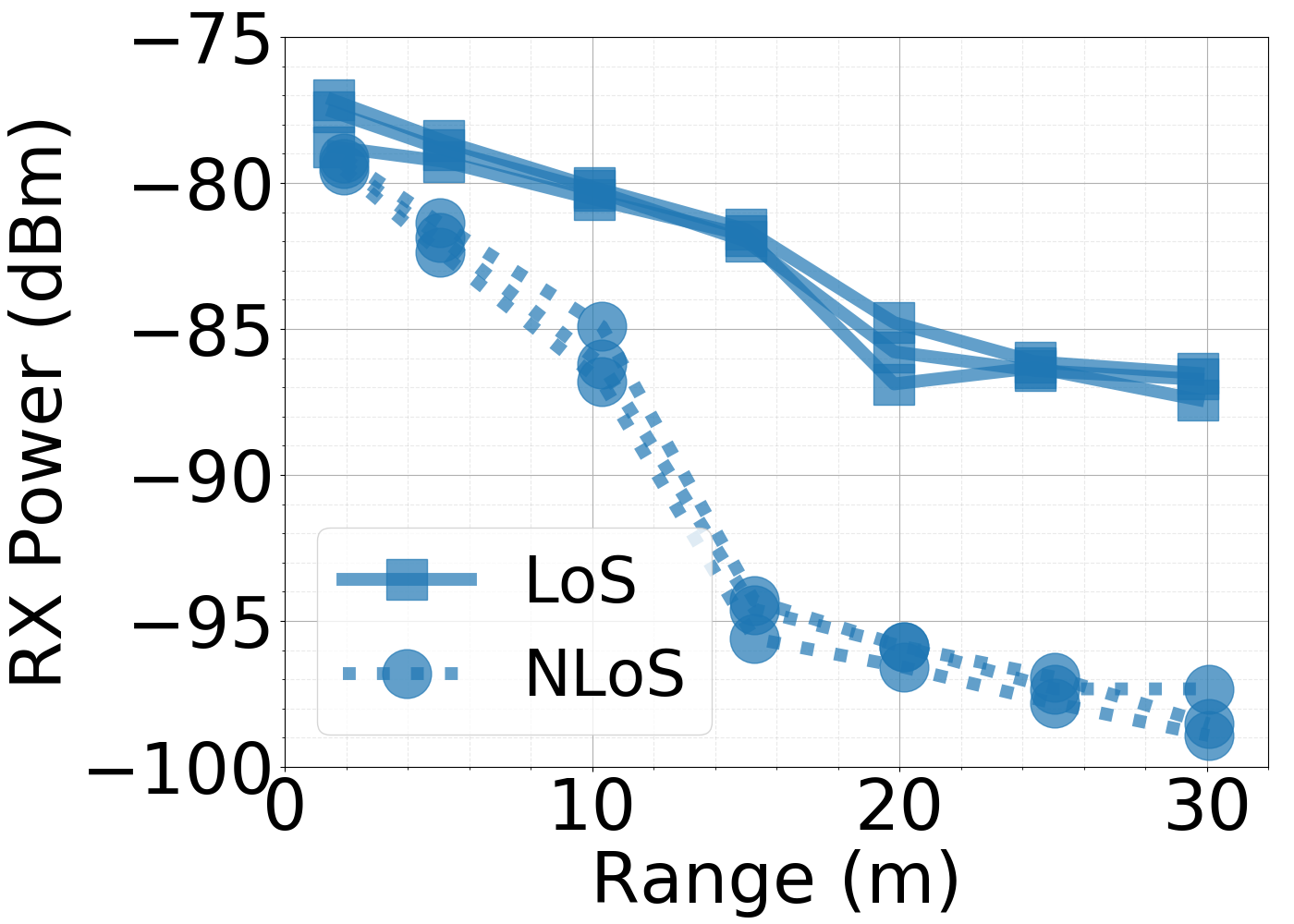} \label{fig:rangevsrxp}
     }
    \subfloat[Received power vs. Error]{%
      \includegraphics[width=0.5\linewidth]{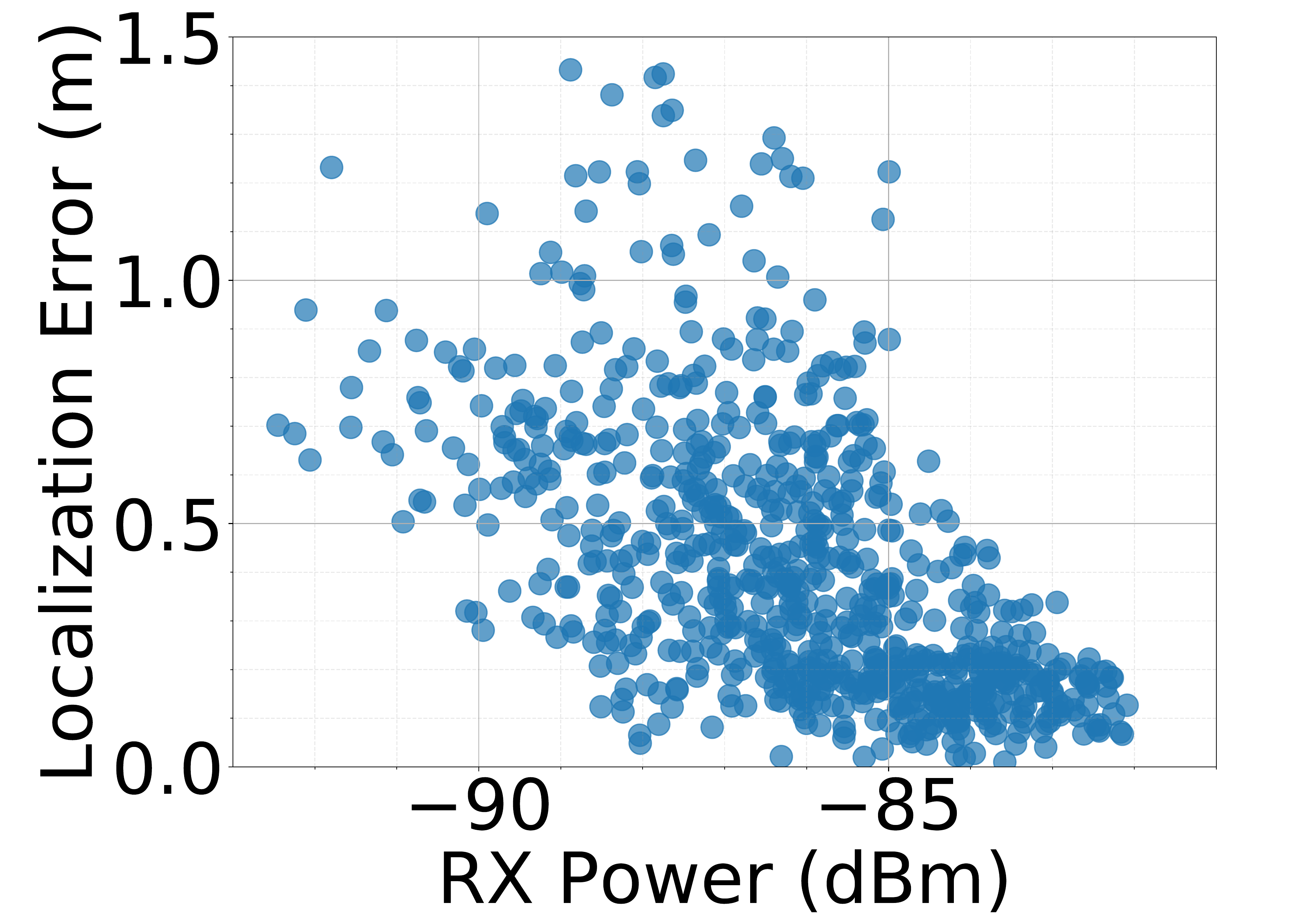} \label{fig:errorvsrxp}
     }
     \vspace{-0.4cm}
     \caption{Received power vs. localization error.}
\vspace{-0.2cm}
\end{figure}

\noindent{\em \underline{Selection of features that affect ranging:}} 
We investigate several link related metrics (e.g., first path power, amplitude, channel impulse power, etc)
and find that amongst all metrics, received signal power (estimated as shown in equation below) exhibits a highly 
discriminative  behavior with respect to range estimates.
\begin{equation}
P_{rx} = 10 \times \log_{10}\left({\frac{C \times 2^{17}}{N^2}}\right) - A \hspace{0.1cm}dBm \\
\end{equation}
where, $C$ is the Channel Impulse Response Power, $N$ is a preamble accumulation count used to normalize the
amplitude of channel
impulse responses~\cite{gururaj2017real}, $A$ is a constant determined with the pulse repetition 
frequency (details found in~\cite{dw1000}). 
Fig.~\ref{fig:rangevsrxp} shows the impact of received power on ranging under LoS and NLoS conditions for three 
different anchors, averaged over several device locations. As the range from the anchor increases, 
the received power under NLOS goes below -95dBm even within 15m, while it remains above -90dBm even 
after 30m for LoS. Consequently we find that that the received power has a strong negative correlation with localization error (Fig.~\ref{fig:errorvsrxp}). Thus, received power to an anchor ($P_i$) along with its range ($R_i$) can serve as an effective discriminative feature, capturing the impact of NLoS on the accuracy of its range and eventually location estimate, although with a nonlinear relationship.

\begin{figure}[t]
\vspace{-0.2cm}
\centering
\begin{minipage}{.47\linewidth}
\centering
  \includegraphics[width=\linewidth]{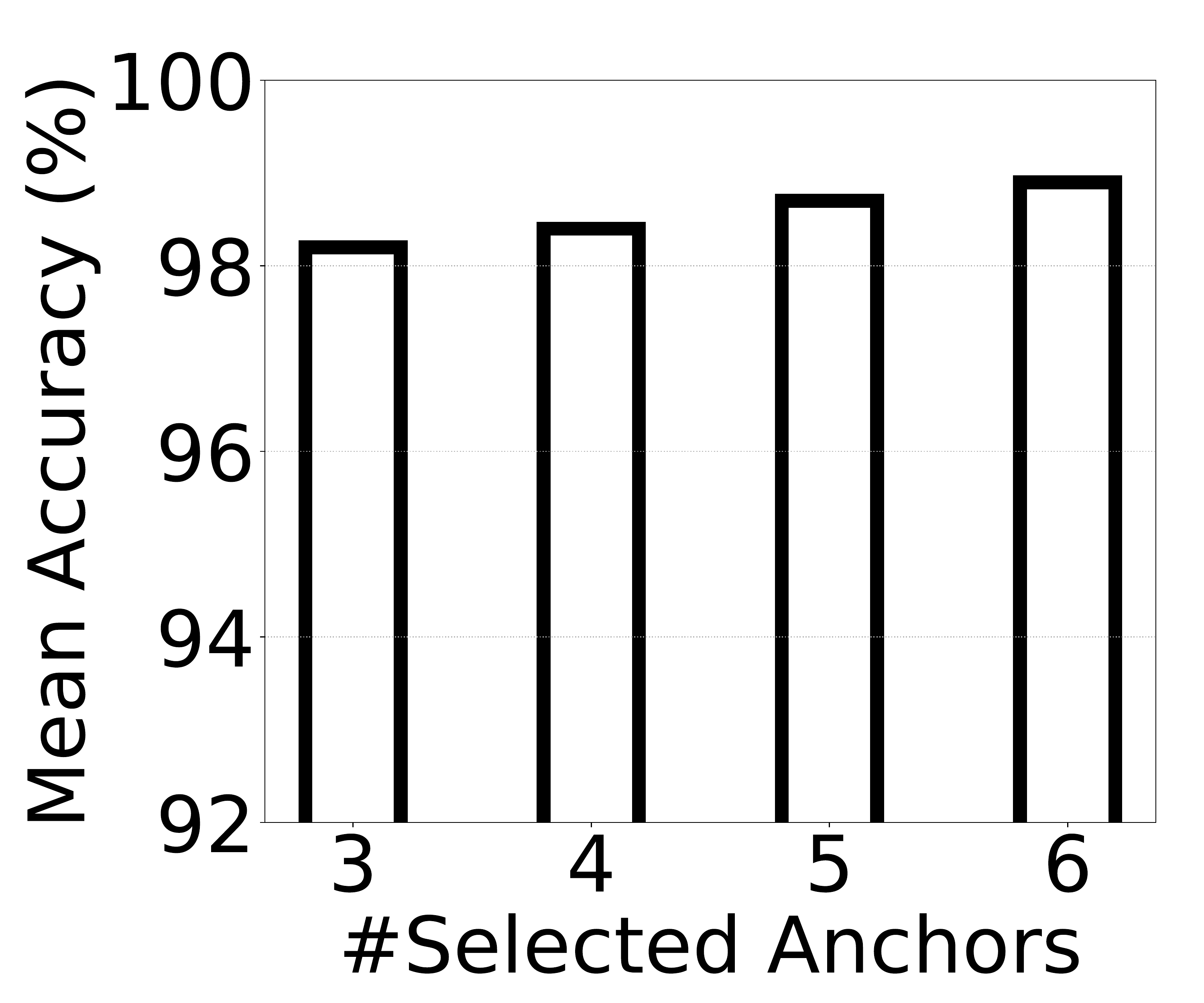}
  \vspace{-0.5cm}
  \caption{Accuracy with different no. of anchors.}
  \label{fig:anchorselection1}
\end{minipage}\hspace{0.3cm}%
\begin{minipage}{.47\linewidth}
  \includegraphics[width=\linewidth]{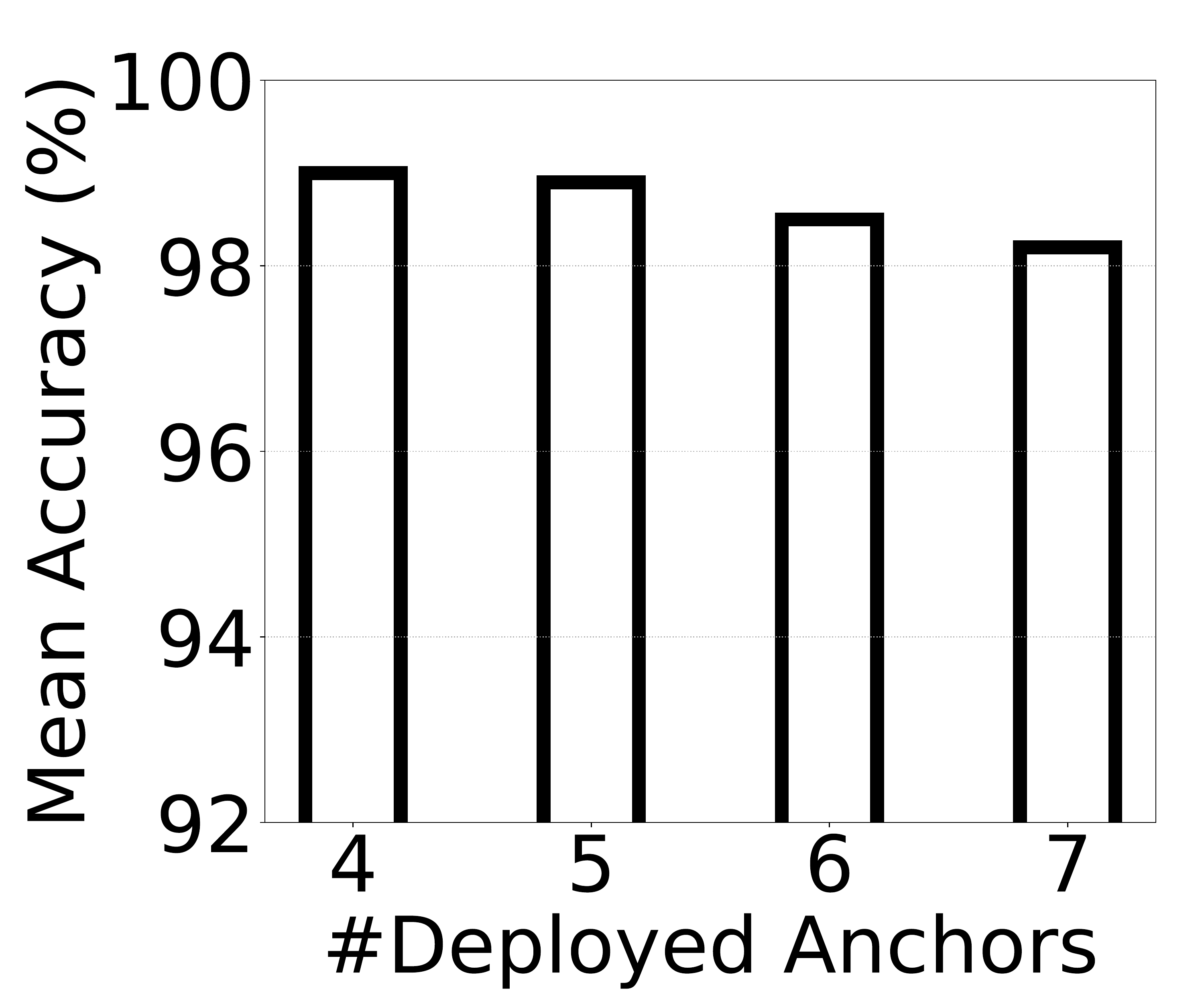}
  \vspace{-0.5cm}
  \caption{Accuracy with 3 best anchors.}
  \label{fig:anchorselection2}
\end{minipage}
\vspace{-0.6cm}
\end{figure}

\noindent \textbf{(2) Leveraging RF features to address NLoS:} 
The extracted feature $\{P,R\}$, serves two purposes in \system: (i) to filter out accurate range estimates 
used for localization when $>3$ anchors are available, and (ii) indirectly capture the certainty (variance) 
of the location estimate for subsequent fusion. Using these features, we next explain how filter out poor anchors to avoid incorrect ranges for localization.

\noindent{\underline{Anchor selection for location estimate:}} In order to select a best subset of anchors among all available anchors, we use simple ML models (e.g., SVM). This design choice keeps anchor selection light weight, and avoids burden on the fusion model. We train a classifier that selects the best $K$ anchors for localization by fitting a model with the ranges from all anchors along with their corresponding received powers as input, and the best anchor set (providing min. error compared to ground-truth) as the output binary vector. We use a multi-output classification using classifier chains~\cite{read2015scalable} to exploit the correlation among the anchors rather than independently selecting each anchor. We fit three models--- SVM, Logistic Regression, and a RandomForests classifier and perform a grid search on each of them to tune the best parameters. After the grid search, we select a best performing model (SVM in our case) and its optimal parameters from the search. Figure \ref{fig:anchorselection1} and \ref{fig:anchorselection2} shows the prediction performance in terms of mean accuracy, when predicting a subset of anchors. As shown, the model performance is highly accurate with the accuracy increasing as expected when more anchors are available for selection. Note that the model only helps filter the inputs ($K=3$ in \system), while the multi-lateration algorithm is still responsible for estimating the location estimate.  

\begin{figure}[t]
\centering
     \vspace{-0.2cm}
    \subfloat[Well-lit (568 matches)]{%
      \includegraphics[width=0.48\linewidth]{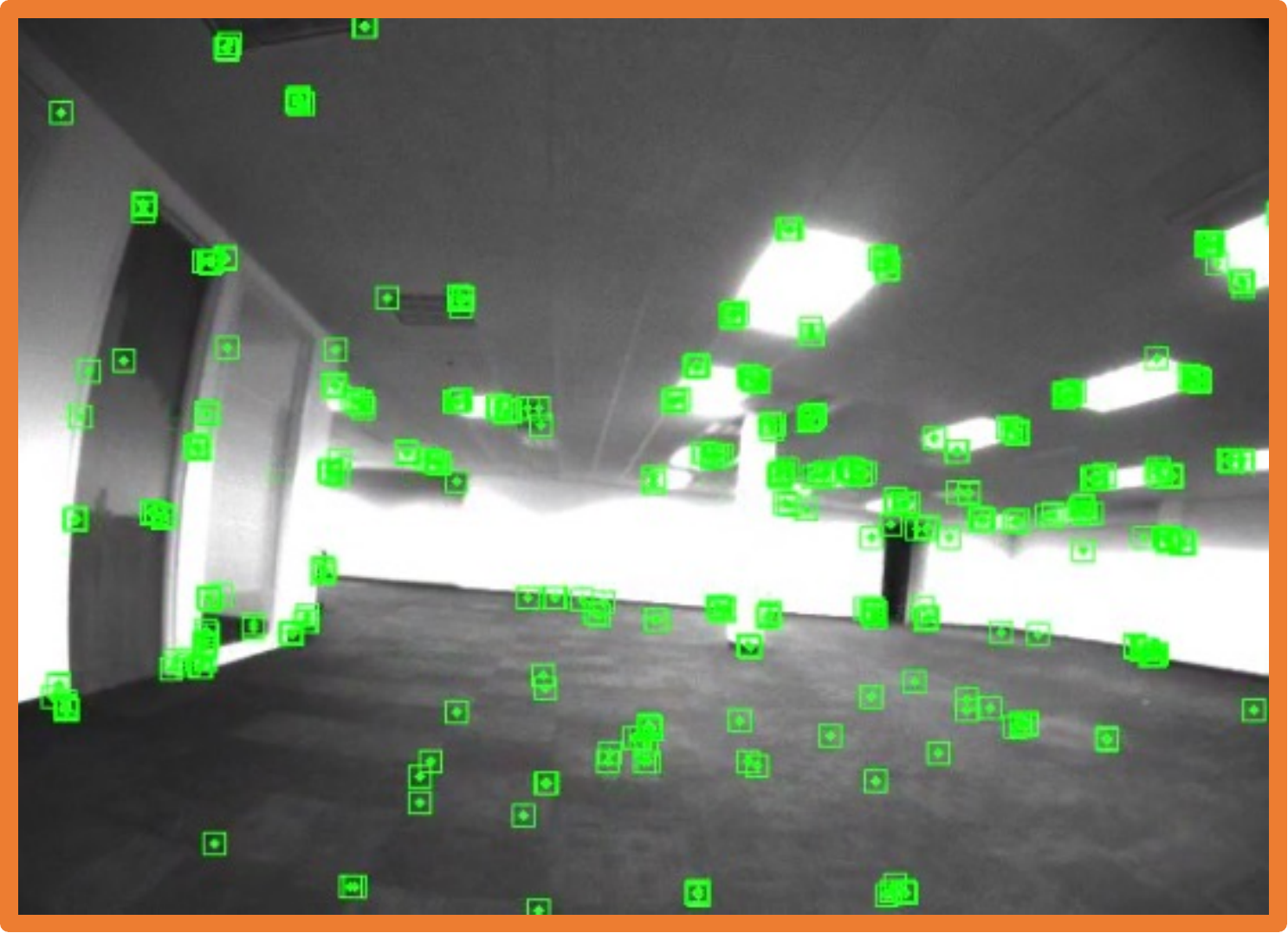} 
     }
    \subfloat[Dim-lit (252 matches)]{%
      \includegraphics[width=0.48\linewidth]{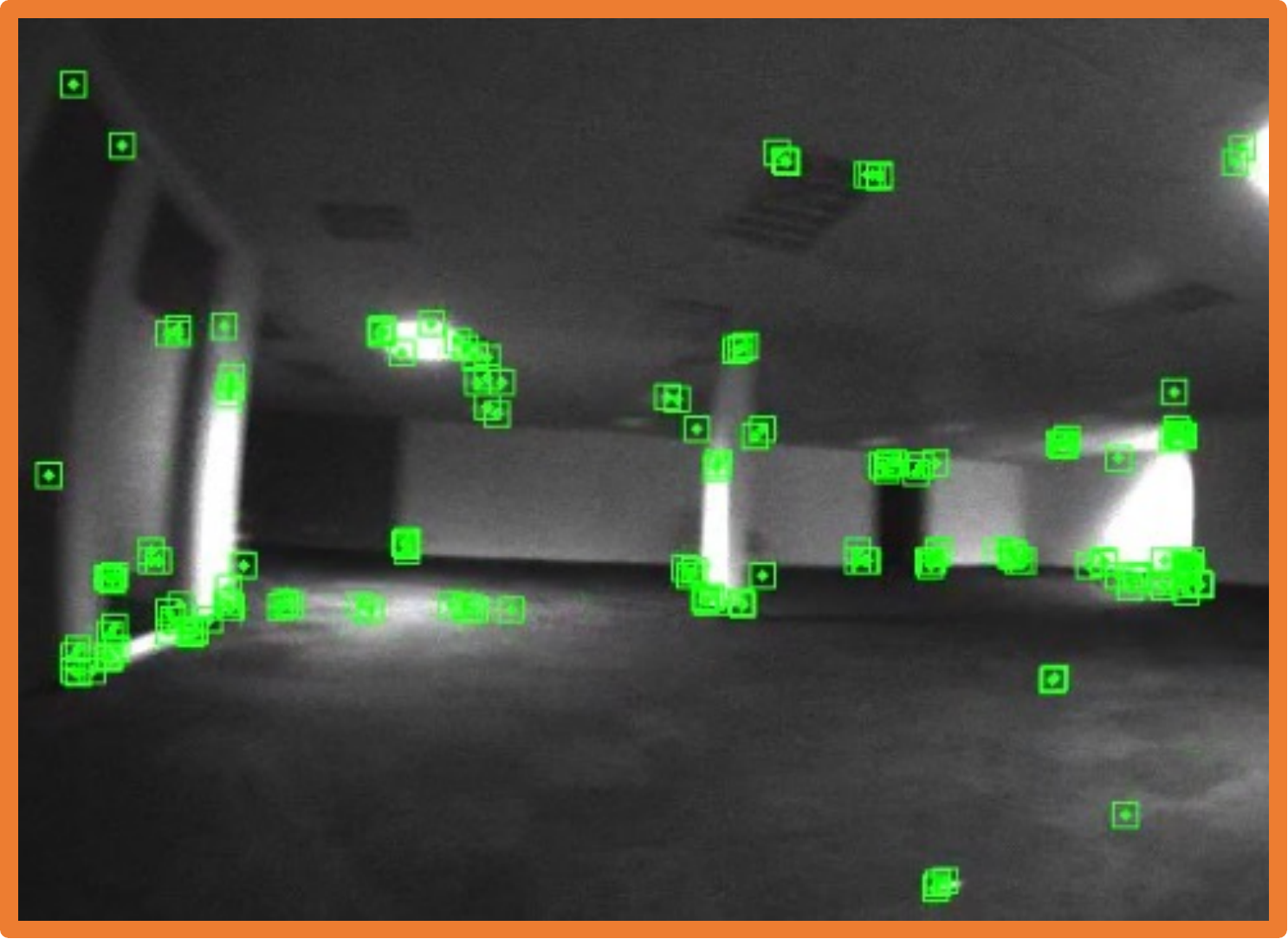}  
     }
     
     \vspace{-0.2cm}
    \subfloat[Matches vs. Error]{%
      \includegraphics[width=0.9\linewidth]{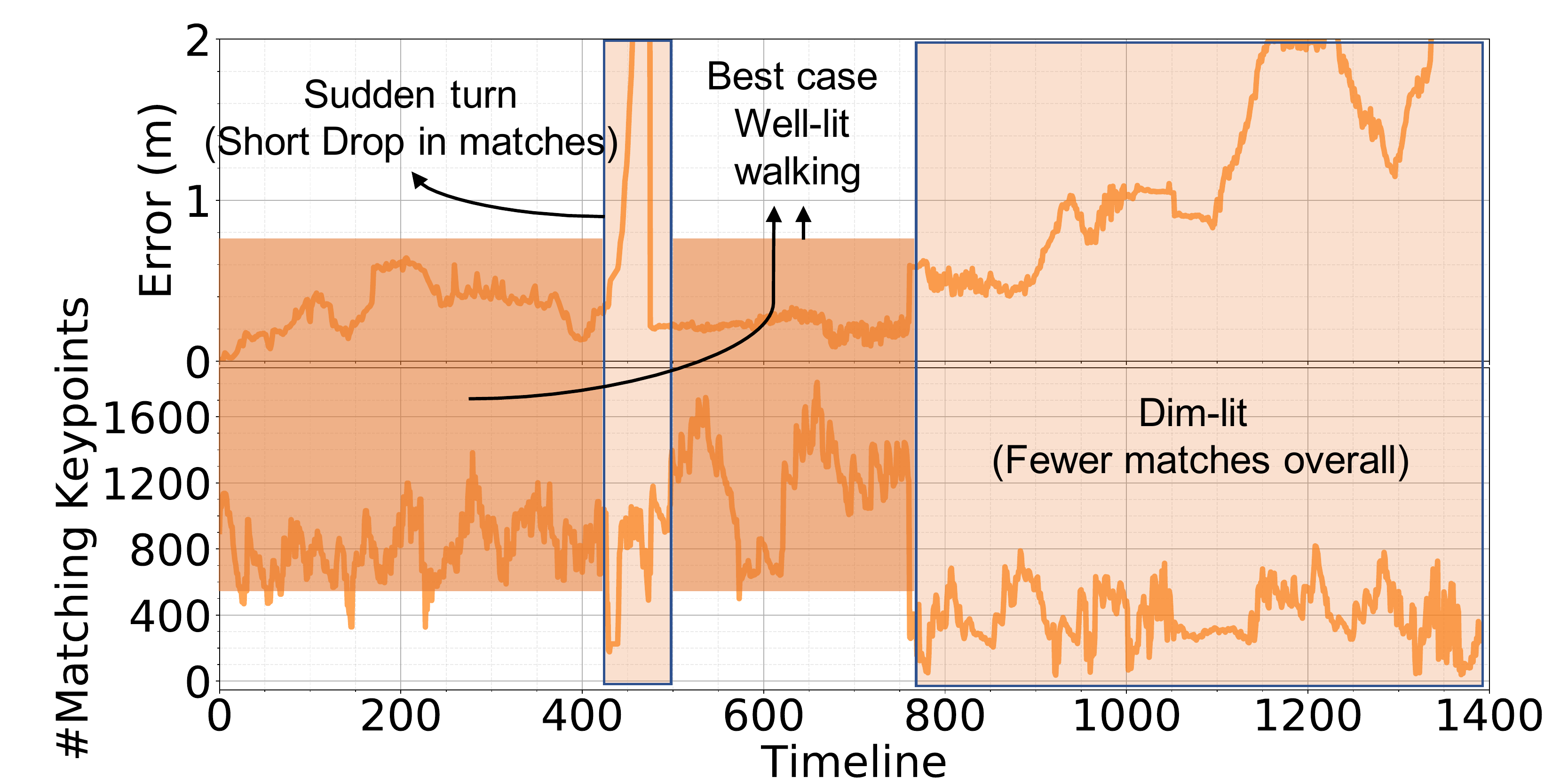}  
     } 
     \vspace{-0.4cm}
      \caption{VO matching keypoints (markers in green) vs. tracking error under different lighting conditions.}
     \label{fig:visualfeatures}
\vspace{-0.2cm}
\end{figure}

\begin{figure*}[t]
\centering
\vspace{-0.3cm}
      \includegraphics[width=\linewidth]{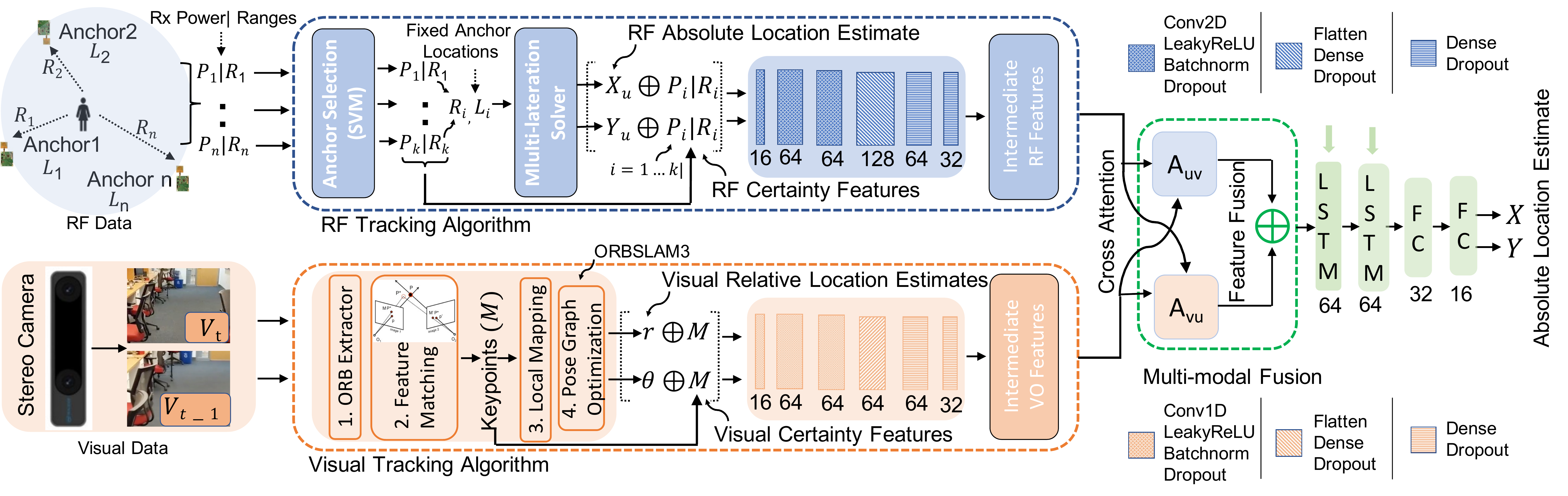}
\vspace{-0.6cm}
      \caption{\system Hybrid Fusion. \system embraces the benefits of classicial solutions at the first layer, and employs an ML pipeline for multi-modal sensor fusion to obtain absolute location estimate.}
     \label{fig:rovarfusion}
\vspace{-0.3cm}
\end{figure*} 

\noindent \textbf{(3) Feature composition for fusion model:} While anchor selection filters NLoS anchors to improve 
the location estimate, a device often might not have access to 3 LoS anchors.
As seen in Fig.~\ref{fig:errorvlosnlosrf},  the maximum localization error is still over 1m even after selecting the best set of anchors. 
Thus, feeding the absolute location estimates alone can misguide the fusion framework to adapt inefficiently to different RF conditions.
Hence, \system leverages 
 the additional feature of received powers together with the ranges as a form of certainty measure 
 (in modeling anchors as LoS and NLoS) and combines these certainty features with absolute 
 location estimates from the multi-lateration algorithm as a composed RF input to our fusion framework. 
 Formally, the input from the RF path to \system's module is given as 
 $F_r$ = $<{X_u} \oplus P_i | R_i>, <{Y_u} \oplus P_i | R_i>$, $i\in[1,K]$, where $\oplus$ is the 
 concatenation operation of location estimates with the device's range and received power to each anchor.

\subsubsection{VO Module:} \label{sec:vomodule}
Most visual tracking solutions use a stream of stereo frames to incrementally (relatively) localize the camera/device on a frame-by-frame basis, by extracting unique features from each frame. Typical features can be SIFT, SURF, ORBs~\cite{rublee2011orb}, etc. \system  (ORBSLAM3) employs ORB features that are invariant to scale, rotation, and translation properties. At a given time instance ($t$), these features are extracted from two or more camera frames ($V^s_t$), reprojected onto the real world to estimate the depth of each feature, and the scale of tracking. These matched features are used to find correspondences with a previous reference frame(s) ($V^s_{t-1}$) and create a set of matching keypoints which are then used to compute relative displacement estimates (${\Delta x}_v,{\Delta y}_v$) to get the current position ($x_v,y_v$). 

\noindent \textbf{(1) Avoiding drift by tracking translation \& heading:}
Being a relative tracking approach, even temporary environmental artifacts (limited visual features, dynamic scenes, etc.) that degrade just a few displacement estimates, result in the continuous accumulation of errors over time. 
Hence, naively combining VO's relative position estimate directly with UWB's absolute location estimate, can lead to long-term drift problems in the final fused location estimate. \system addresses this challenge by employing relative displacement in translation ($r_v$) and the heading ($\theta_v$) directly rather than using the final relative estimates of position, to drive the fusion module. Hence, even when odometry faces displacement errors temporarily, the resulting error propagation is only in displacement (heading continues tracking the absolute trajectory direction), which is transient and does not propagate. Further, even this transient error propagation is completely eliminated, when its relative estimates ($r_v, \theta_v$) are fused together with UWB's absolute location estimates ($x_u, y_u$), and result in accurate tracking.

\noindent \textbf{(2) Composing features to capture estimate certainty:}
Note that the relative estimates can themselves be erroneous even in the absence of any error drift.
Short-term environmental artifacts (e.g. occlusion lighting) can result in inaccurate
estimates with up to 1m error (see Fig. ~\ref{fig:comprehensive-seen}) even in best visually feature-rich environments. 
To compensate for such inaccurate estimates, \system  employs  additional features to capture the certainty 
of the tracking algorithm's estimate . 

Recall that, the features extracted from the images determine the VO's tracking accuracy and robustness. We dissect ORB-SLAM3's tracking component to understand its features in capturing certainty of its estimates. A straightforward feature is the number of ORB features that are matched by the algorithm across all stereo frames. However, these cannot be directly used, as there can be spurious matches and outliers. The latter can be determined based on their estimated depth and removed before feeding them to the tracking algorithm. After filtering the outliers, the final matching keypoints are the ones most relevant for tracking. We study the effect of matching keypoints on tracking performance under different environments and find that the tracking error is stongly influenced by the keypoints. Figure \ref{fig:visualfeatures} shows the number of matching keypoints under well-lit (568) and dim-lit (252) conditions, whose corresponding tracking error is shown in Figure \ref{fig:visualfeatures}c. Note that if the keypoints go below 100 for longer periods, the tracking completely fails. We also explored other features such as depth of matching features, outliers and inliers, reprojection error, etc, but we find matching keypoints ($M$) to best capture the certainty of the tracking estimates. Hence, \system combines this certainty feature $M$ with the relative tracking estimates $(r,\theta)$ as a composed VO input to our fusion framework. Formally, the input from the VO path to \system's module is given as $F_v$ = $<r \oplus M>, <{\theta} \oplus M>$, where $\oplus$ is the concatenation operation.

\subsection{Sensor Fusion through Cross-Attention}
Fig~\ref{fig:rovarfusion} shows the end-to-end architecture of \system's hybrid fusion. \system first prepares the composed features from individual sensors by passing them through a simple CNN network. This allows the location estimate and its certainty-related features to be embedded into a more representative feature 
 that can enable effective fusion. 
 
 \system then adopts an \textit{Attention} mechanism for fusion, a commonly used learning 
 technique~\cite{vaswani2017attention, wang2018non}, for adaptively weighting the features of the sensors, 
 so as to leverage their complementary nature. 
 While self attention~\cite{lee2019self} weights the features of an individual sensor to self-adapt and 
 eliminate the influence of outliers, 
 cross-attention focuses on weighting each sensor with respect to one another 
 to extract inter-sensor correlations and leverage their complementary nature. Intuitively, the model 
 would weight the RF estimates (with higher certainty) more when VO encounters unfavorable 
 environments (e.g., dim-light; lower certainty of VO), and the VO estimates more when RF 
 estimates suffer from NLoS anchors. \system directly adopts cross-attention since the features 
 engineered by its algorithms already capture self-attention (incorporating features that correlate with tracking error). 
The two cross-attention masks ($A_{rv}$, $A_{vr}$) are {\em jointly} learned using the RF 
and VO features ($F_r$, $F_v$) respectively. 
\begin{equation}
    A_{rv} = Sigmoid((W'_{rv}F_v)^T W''_{rv}F_v), 
    A_{vr} = Sigmoid((W'_{vr}F_r)^T W''_{vr}F_r)
\end{equation}
where, $W'$ and $W''$ are the weights learned during the training, which transform the extracted 
features into a lower dimension version of original RF and VO inputs, that extracts the underlying 
global topology of data. On a high level, the equation captures meaningful features through local convolutions
and long term dependencies through embedding spaces, jointly adapting the masks by capturing cross-correlations
between the two sensors. After the masks are learned, each mask is applied to its respective sensor feature 
(element-wise, $\odot$) and then merged (concatenated, $\oplus$) together to provide the fused feature
  {\em ($A = [A_{rv} \odot F_v] \oplus [A_{vr} \odot F_r]$).}
Finally, the output from cross-attention, which is a single dimension flattened array,  is  
forwarded to an LSTM network - a 2 layer RNN (with 64 hidden units per layer) to model the temporal 
dependency of the fused features, followed by 
2 Fully Connected (FC) layers that finally outputs the predicted absolute location.
The LSTM's ability to access its outputs from prior time instants (i.e., prior absolute
location estimates), enables it to effectively predict {\em absolute} location estimate
while relying on the {\em relative} location estimate from VO (when RF features have large uncertainty), 
thereby eliminating (resetting) potential VO drifts.

\begin{figure}[t]
\vspace{-0.2cm}
\centering
      \includegraphics[width=0.9\linewidth]{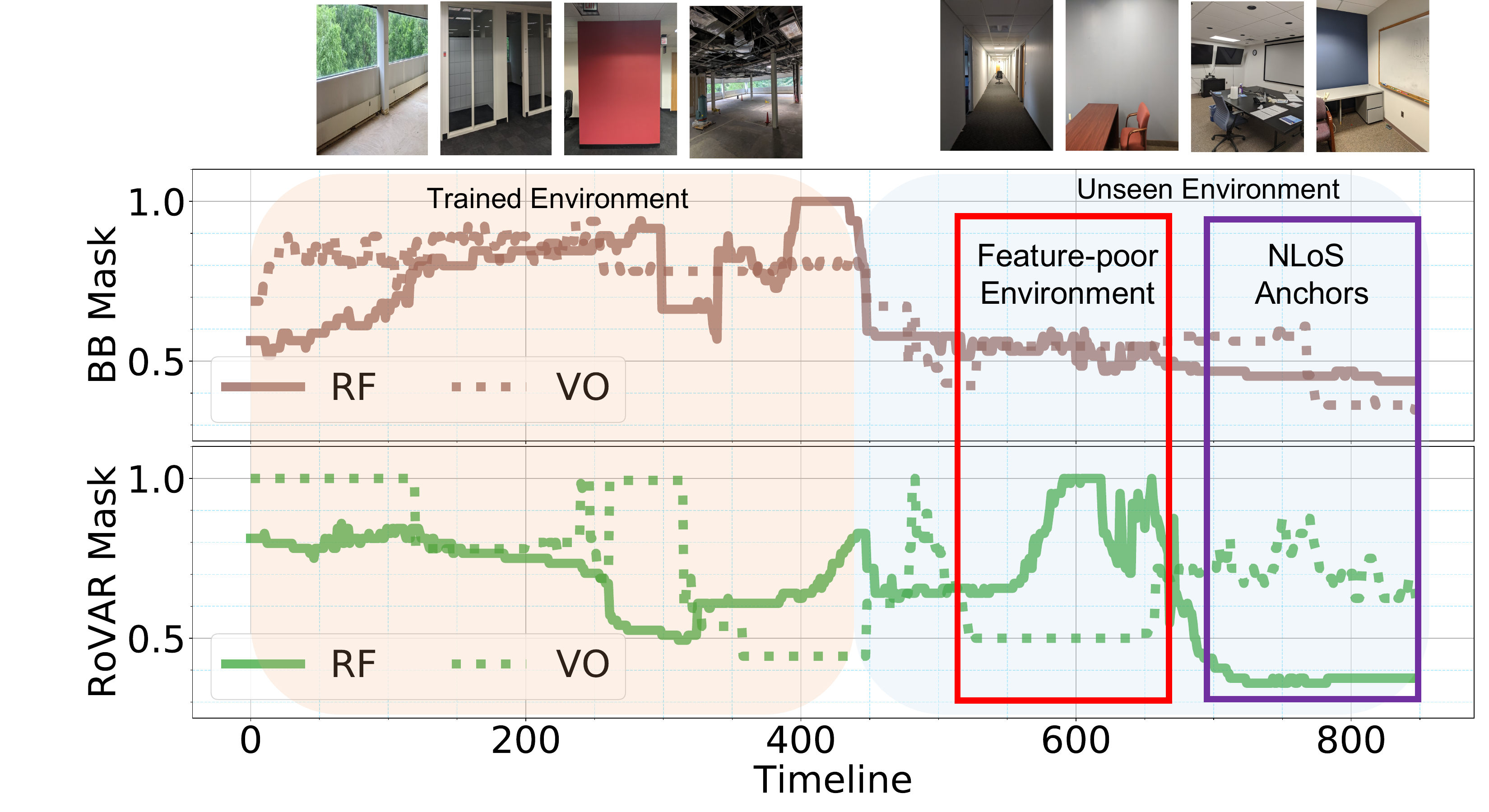}
\vspace{-0.4cm}
      \caption{Attention mask for BB vs. \system. BB's attention is comparable to \system in trained environemnt
      (orange region) while failing in unseen places (light blue region).}
     \label{fig:attexample}
\vspace{-0.4cm}
\end{figure}

\noindent \underline{\textit{Effectiveness of Cross-Attention:}}
\label{sec:CrossAttention}
Fig. \ref{fig:attexample} shows \system's attended weights for RF and VO feature vectors. As the device suffers dim-light conditions, \system
increases RF feature attention from .6 to 1, while decreasing that of the VO's features from .9 to .5, 
and in NLoS, RF features are weighted low (decreased from 1 to 0.1), while VO's weights increase from 0.5 to 0.85.
Given the independent and complementary nature of the RF and VO sensors as well as their environmental artifacts, 
it is clear that \system's cross-attention plays a valuable role in effective sensor fusion.

\subsection{\system vs. Pure Deep Learning Fusion} 
\label{sec:bbdesign}
At this point, it is natural to wonder -  {\emph{"How and why is \system better than
prior machine learning based fusion solutions?"}} 
Prior works (e.g., milliEgo~\cite{lu2020milliego}, DeepTIO~\cite{saputra2020deeptio}, VINET~\cite{clark2017vinet}, etc.) have employed 
raw sensor data to directly train a (BlackBox -- BB) ML model, compared to \system's hybrid approach, where ML is largely leveraged for fusion alone. These blackbox models tend to be more complex (e.g., additional attention layers), requiring higher computational resources (e.g., more convolution and RNN layers). Further still, having to extract the challenging problem structure inherent to {\em absolute localization}, they are unable to generalize and deliver robust performance in unseen environments (see \S\ref{EVAL}).  

For an objective comparison, we also design a pure ML-driven counterpart solution to \system (called BlackBox, BB), by instrumenting the network architecture similar to those used in~\cite{lu2020milliego} for our UWB and stereo-camera inputs. The key differences in 
BB are that, it has a deep feature extraction network for both UWB and camera streams, and employs a self-attention module. The role of self-attention 
is to enable a form of filtering of the weak input features, while emphasizing those that contribute to an accurate output prediction. After the features are self-attended, it follows a cross-attention and fusion pipeline similar to \system.

While \system's detailed performance comparison with BB is deferred to \S\ref{EVAL}, we highlight two of its key advantages:
i) much of its feature extraction burden is driven by efficient algorithms, allowing its ML-component to focus solely on fusion, enabling it to generalize and accurately track absolute location even in unseen environments (see Fig.~\ref{fig:attexample}), 
and ii) lack of a complex feature extraction network reduces its model complexity 
significantly, making it light-weight and deployable on resource-constrained platforms providing real-time operation. In our testbed (described in \S\ref{IMPL}), we show that \system has only 1.2\,M parameters while BB requires 70\,M, and when translated into raw data (after model compression), \system's memory footprint is 62$\times$ smaller than BB, directly leading to a 5 fold decrease in inference latency (see \S\ref{sec:realtime}).

\section{Implementation}
\label{IMPL}
\begin{figure}
\vspace{-0.2cm}
\centering
      \includegraphics[width=0.9\linewidth]{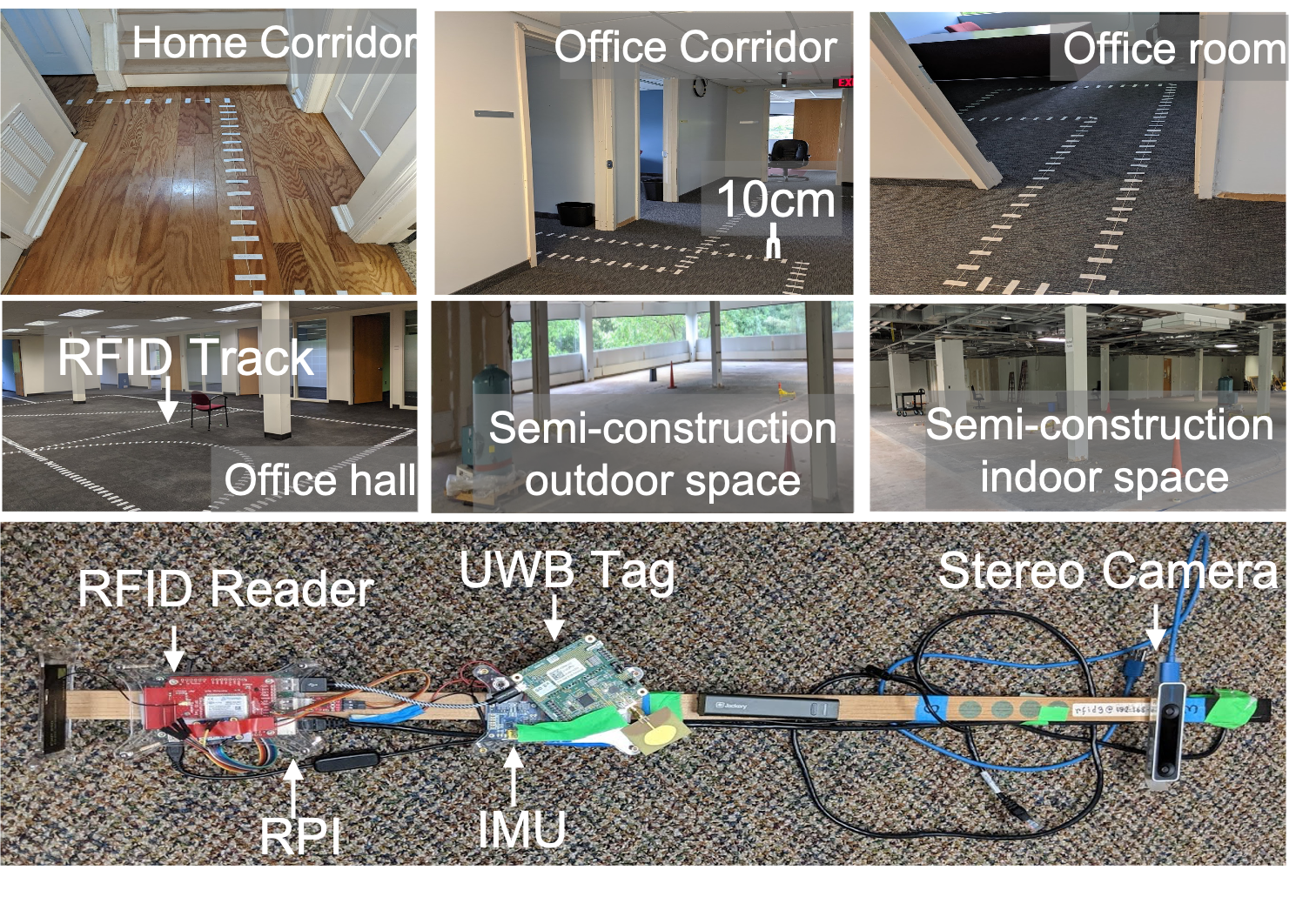}
      \vspace{-0.5cm}
      \caption{RFID (GT) tracks and \system prototype}
     \label{fig:implementation}
\vspace{-0.4cm}
\end{figure}

{\bf \system system components:}
We implement a hand held prototype of \system,
containing: 
(a) Decawave EVK1000~\cite{dw1000evk} 
UWB radios for RF localization (\S\ref{sec:rfmodule}),
(b) Intel T265 stereo camera~\cite{intelrealsenset265} for VO tracking (\S\ref{sec:vomodule}) and (c) A ThingMagic M6E-Nano RFID reader~\cite{rfidreader} for Ground Truth (GT) measurements. 
The RFID reader and the UWB radio are connected to a
RaspberryPi-4 (RPi) 
computer via the RPi's serial and Ethernet ports, respectively. A laptop (or a Jetson Nano for real-time implementation)
is connected to the Intel camera via an ethernet cable.
All the devices are time-synced using a local NTP server.
Fig.~\ref{fig:implementation} shows 
the \system prototype. 
As seen in the picture, the camera, UWB and the RFID reader are all on the same 
vertical plane, ensuring that the location determined 
by each sensor are on the same X-Y plane.

\noindent {\bf Testbeds:} We build several testbeds across multiple floors 
of an office and a home with varying light and scenary. They include
conference rooms with artificial lights, semi-constructed cemented-area, glass walled rooms,
and office corridors surrounded by obstacles (cubicles or offices) (see Fig~\ref{fig:implementation}). 
Lighting conditions in all
but one, can be controlled using light-dimmers, while one testbed has only natural sunlight.
We use a subset of testbeds (trained environments) for data collection, while the rest are kept completely unseen 
for testing the generalizability of \system.

\noindent{\bf Data collection:}
Each testbed has 2-to-3 different trajectories that contain
simple straight lines to more complex curvy-trajectories. For each trajectory, we 
collect data in: (i) bright and dim lighting conditions, and (ii) when user is walking and
running. The UWB anchors 
are often occluded due to physical structures and hence the UWB data contain both LoS and NLoS ranges.
We have collected data from 72 different traces, totalling a distance of 4.8K meters for training the \system models. \\
{\bf Ground Truth (GT):}
We use RFID tags, placed continuously every 10cm along 
a user's walking path for GT data. 
While other modalities like Lidars are conducive for single-user GT,
the physical RFIDs provide absolute GT enabling
multi-agent tracking evaluations. 
We physically record each RFID's X-Y location, and as the user walks along
the predefined trajectories holding the \system prototype in the vertical position 
(see Fig.~\ref{fig:implementation}), the RPi which is connected to the RFID reader (with the antenna 
at the bottom of the stick) records a time-stamped EPC and
RSSI information (at 50Hz) of the RFID directly below it. The Tx power of the RFID reader is set to a minimum
so that the RFID antenna reads the RFID tag only when it is a couple of inches directly
above it. Additional tag that is rarely read is filtered using RSSI information. 
Further, the velocity of the device, derived from a short sliding window of tag reads (say 20 tags), allows 
for accurate interpolation of location within the 10cm granularity.\\
\noindent {\bf UWB localization:}
We use 7 static anchors placed at known 
locations in each testbed. 
The RPi connected to the UWB radio on the \system prototype ranges
sequentially with each UWB anchor. 
\system implements an optimized ranging protocol~\cite{dw1000evk}, to reduce
the total packet exchanges with each anchor. Range estimates are
recorded with timestamps at 15Hz speed.
\system implements the geometry constrained least-squares multi-lateration 
algorithm~\cite{hua2014geometrical} for localization.\\
{\bf ORB-SLAM3:} 
We use Intel T265 camera~\cite{intelrealsenset265} for VO.
The T265 has stereo cameras with fisheye lens and records images at
848x800 pixel resolution. The recorded camera images is time-synced with the
UWB and the GT data. We use ORB-SLAM3 ROS implementation~\cite{campos2021orb}
for tracking. \\ 
{\bf Network training:}
We use 70\% of our dataset for model-training,
10\% for validation and 20\% for evaluation. We ensure that our
dataset is diverse to avoid the case of over-fitting. We use Pytorch~\cite{paszke2019pytorch}, an
open source machine-learning (ML) library to implement the \system models.
Before feeding the data to the network, we normalize the input data by subtracting 
the mean over the dataset. We use Adam optimizer with L2 norm as the loss function 
during the training. \system's  model is lightweight and takes less than an hour 
for training even on a desktop-grade Nvidia-1070 GPU.

\section{Evaluation}
\label{EVAL}
\subsection{Experimental Methodology} 
\label{label:expmethodology}
We extensively evaluate \system for both single and (collaborative) multi-user tracking scenarios under diverse (trained and unseen) environmental 
conditions for two different human-motions (running and walking) to demonstrate its robustness and efficacy 
vis-a-vis other state-of-the-art solutions (described below). Our evaluation testbeds in home and office 
environments have varying texture (rich and poor), lighting conditions (good and dim) and
healthy mix of LoS and NLoS UWB anchors (for RF) to mimic real world deployments. We use the Absolute Trajectory Error (ATE)~\cite{sturm2012benchmark}, calculated as the Euclidean distance between corresponding points on the estimated and the GT trajectories, to quantify the performance of \system. For brevity, we show a subset of the results here, and the complete result set can be found in our technical report~\cite{tech-report}. 

\begin{figure}
\vspace{-0.2cm}
\includegraphics[width=\linewidth]{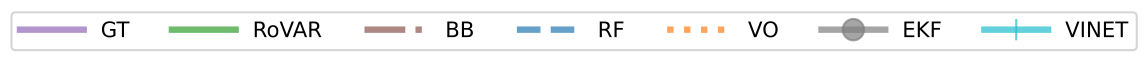}
\centering
    \subfloat[Favorable]{%
      \includegraphics[width=0.5\linewidth]{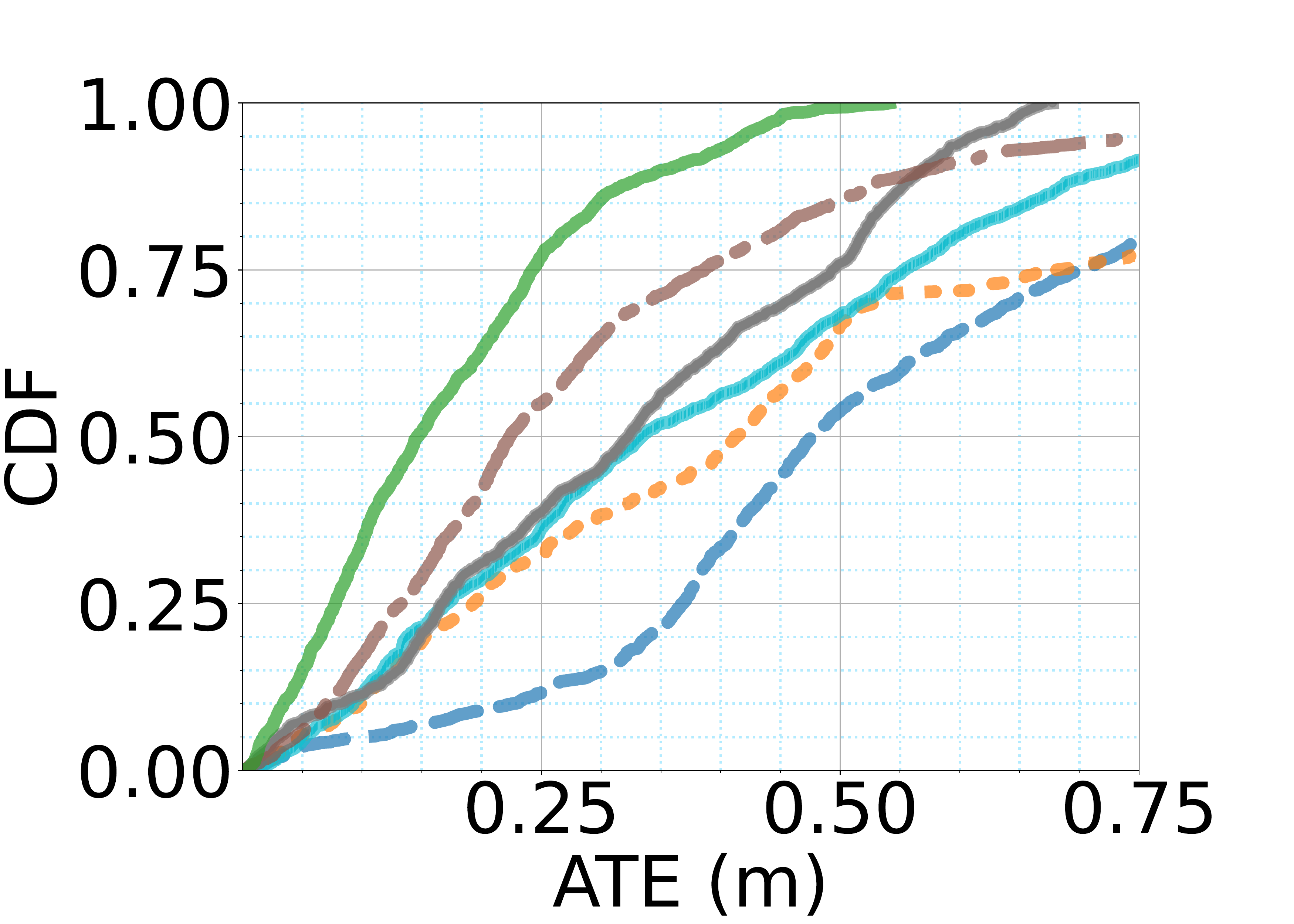} 
     }
    \subfloat[Practical]{
      \includegraphics[width=0.5\linewidth]{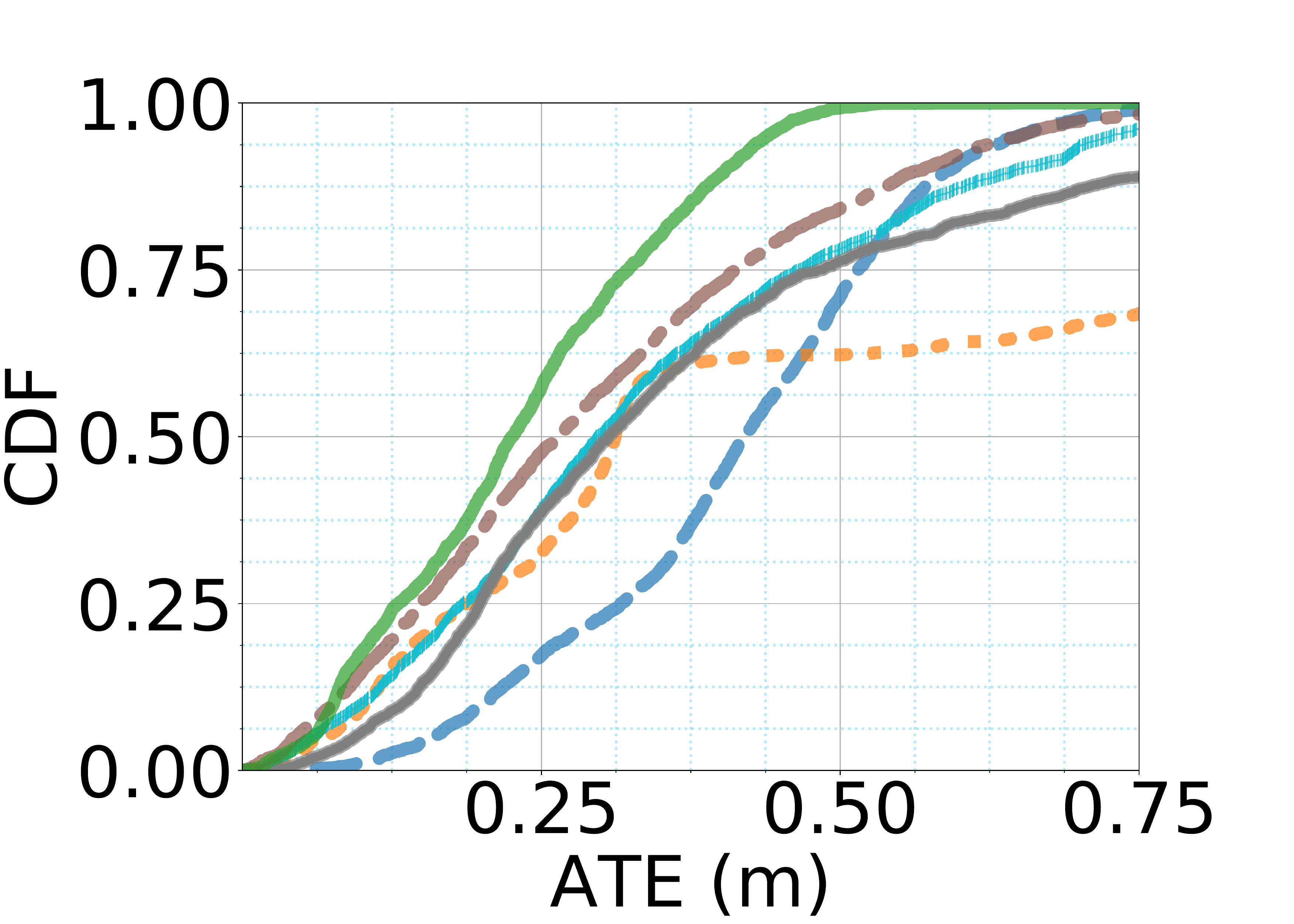}  
     }
  \caption{\system Vs. Baseline performance in Trained environments.}
\label{fig:comprehensive-seen}
\vspace{-0.5cm}
\end{figure}

\noindent \textbf{Baselines:} We compare \system with a suite of baseline solutions that include both algorithmic and ML based fusion: 1) \textbf{RF} is the multilateration-based RF localization solution described in \S\ref{sec:background}, 2) \textbf{VO} is the ORB-SLAM3 VO tracking solution described in \S\ref{sec:background}, 3) \textbf{EKF} is the Extended Kalman-filter, a commonly used algorithmic and industry-grade practical solution (e.g., ARCore~\cite{lanham2018learn} and ARKit~\cite{arkitapple}). We tune the EKF parameters to ensure its optimal performance\footnote{Tuning the EKF to find the optimal {\em process} and {\em measurement} noise covariance matrices is nontrivial in practice. We follow a general approach by iterating the filter until all the estimates until all the parameters converge.}, 4) \textbf{BB} is the Black Box ML based fusion described in \S\ref{sec:CrossAttention}. 
5) \textbf{VINET} is a state-of-the-art ML based VIO tracking solution~\cite{clark2017vinet}.

\subsection{Performance in Trained Environments}
We begin our evaluations in environments used for training the 
ML models.
\noindent \textbf{Favorable conditions:} 
We first evaluate in environments rich in texture, good lighting (good for VO) and LoS RF anchors (good for RF tracking). 
The CDFs in Fig.\ref{fig:comprehensive-seen}a show the ATE (Absolute Trajectory Error) information for each solution. \system has significantly better performance (median error 17cm) over all the alternatives with 35\% and 64\% better median-accuracy even compared to BB and VINET, respectively. While the algorithmic solutions (RF and VO) are agnostic to environmental conditions, the trained environments allow us to capture BB and VINET's performance in their most favorable scenarios. As expected, BB and VINET perform better than 
the individual algorithms, RF, VO, as well as their fused version, 
EKF. This is because VO, despite its inherent 
error-correction mechanism, suffers during trajectory-turns where the number of ORB feature matches are relatively lower than when walking in straight lines, leading 
to accumulation of error; while RF delivers a coarser accuracy compared to visual tracking 
even in the best case. The EKF fusion aims to minimize individual algorithmic errors, but 
is still vulnerable to the individual sensor errors and artifacts, while ML solution are able to mitigate algorithmic uncertainties using data-driven models. 

\begin{figure}[t]
\vspace{-0.2cm}
\includegraphics[width=\linewidth]{figs/legend.png}
\centering
    \subfloat[Walking, Home]{%
      \includegraphics[width=0.45\linewidth]{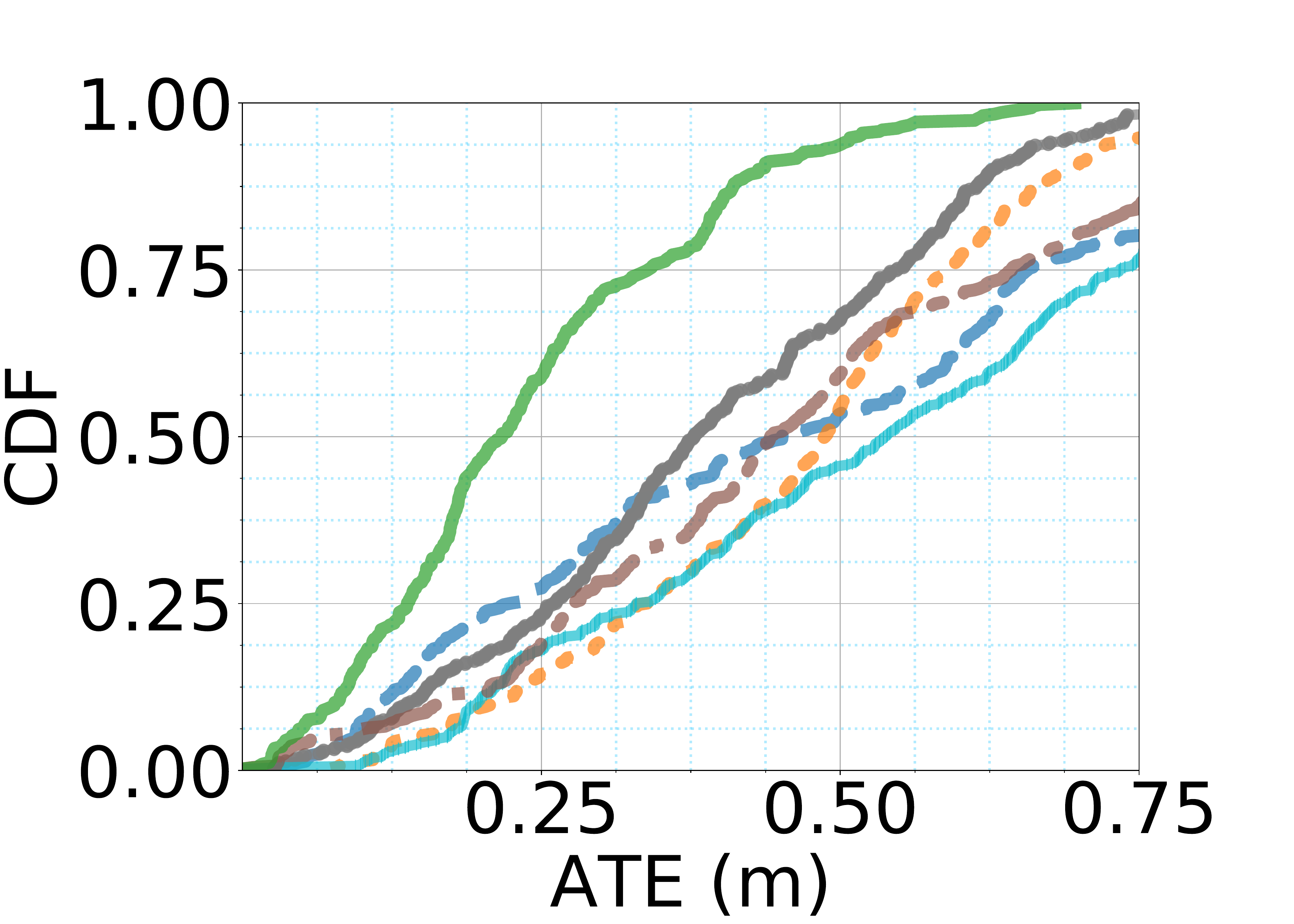} 
     }
    \subfloat[Walking, Office - 3]{%
      \includegraphics[width=0.45\linewidth]{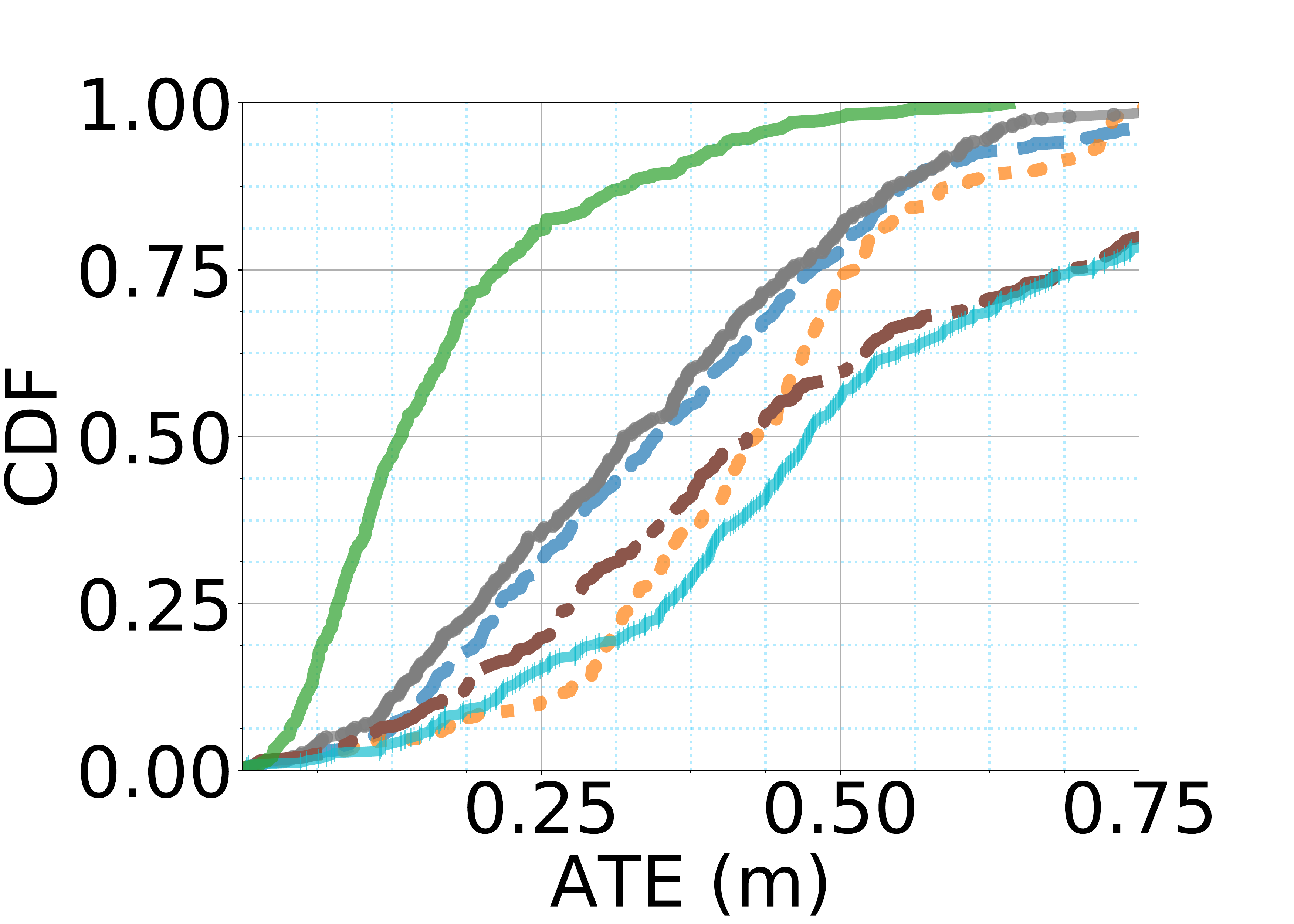} 
     }
     
    \subfloat[Walking, Home]{%
      \includegraphics[width=0.45\linewidth]{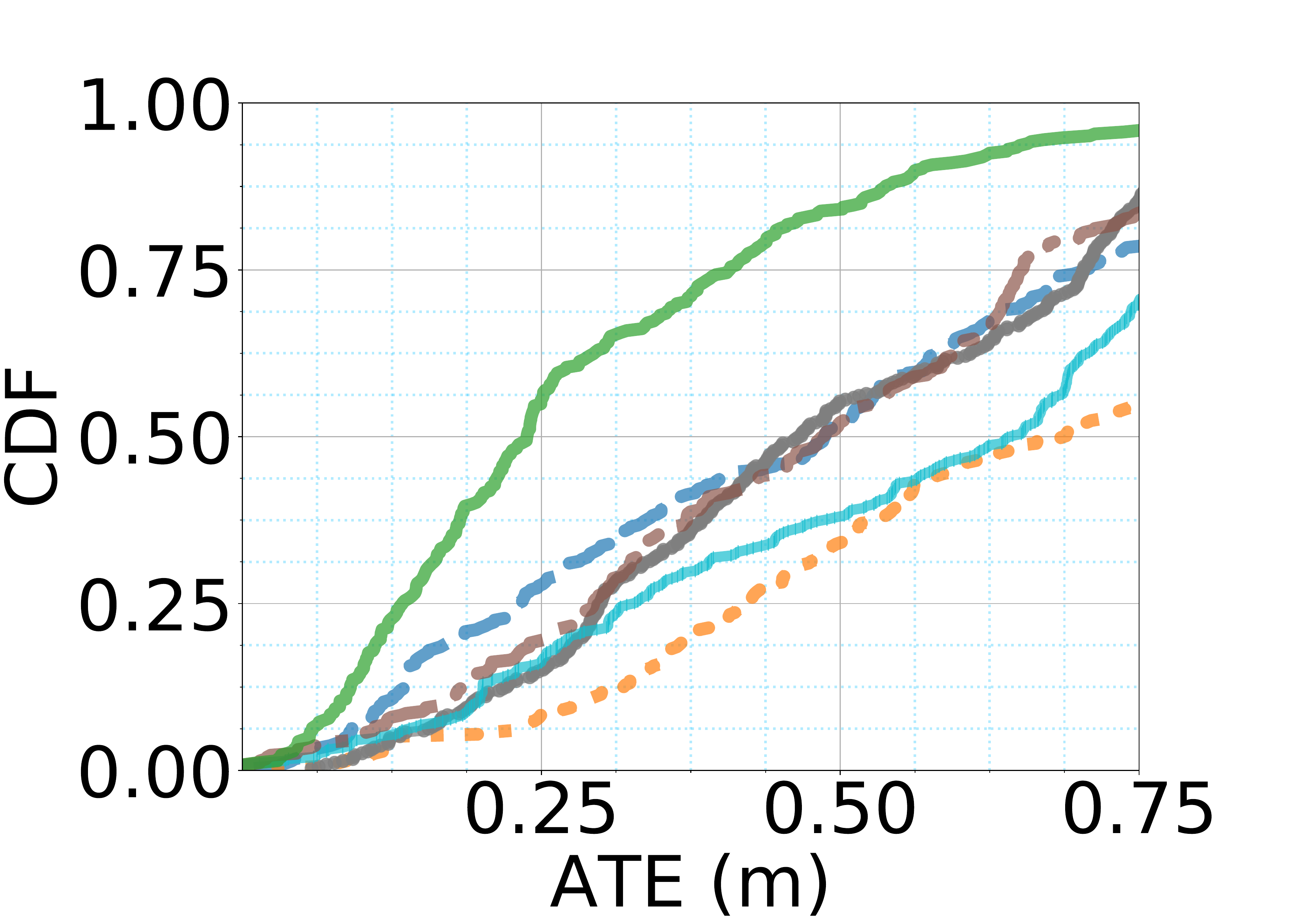} 
     }
    \subfloat[Walking, Office - 3]{%
      \includegraphics[width=0.45\linewidth]{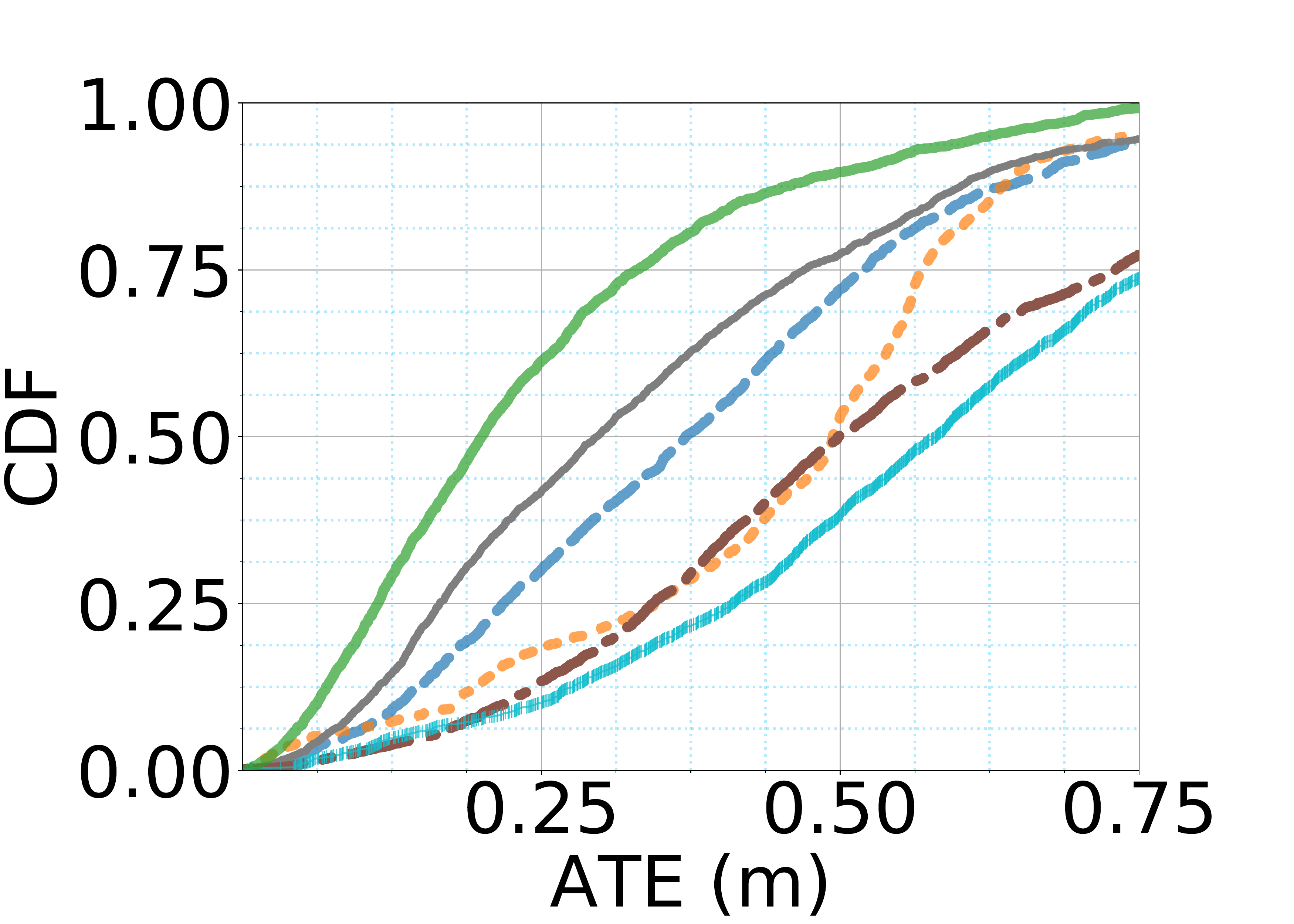} 
     }
      \vspace{-0.15in}
      \caption{\system Vs. Baseline performance in Unseen environments. Favorable conditions(a,b), Practical conditions(c,d).}
     \label{fig:comprehensive-unseen}
      \vspace{-0.2in}
\end{figure}



\noindent \textbf{Practical conditions:}
Next, we evaluate \system in the same environments, but by introducing dim lighting conditions in a room that is part of the trajectory, and moving 
a subset of RF anchors into NLoS. Results are shown in Fig.~\ref{fig:comprehensive-seen}b.
We see that RF and VO algorithms suffer due to NLoS anchors and dim lighting, respectively, with VO (median error 40cm) performing worst. VINET which relies mainly on the VO performance
seem to suffer too due to VO's degradation. BB performs
better (median error 25cm) than the algorithmic ones (EKF - median error 35cm), 
owing to its model (which leverages RF and visual raw data) capturing environmental artifacts and their impact on fusion in the trained scenarios. Finally, \system due to its RF anchor selection module increases the accuracy in RF tracking, and combined with its dual-layer diversity in fusion performs best with median ATE 17cm, a gain of 47\% and 64\% in accuracy 
over BB and VINET, respectively. 

\begin{figure}
  \vspace{-0.2cm}
\includegraphics[width=1\linewidth]{figs/legend.png}
\centering
    \subfloat[Trained Environment]{%
      \includegraphics[width=0.5\linewidth]{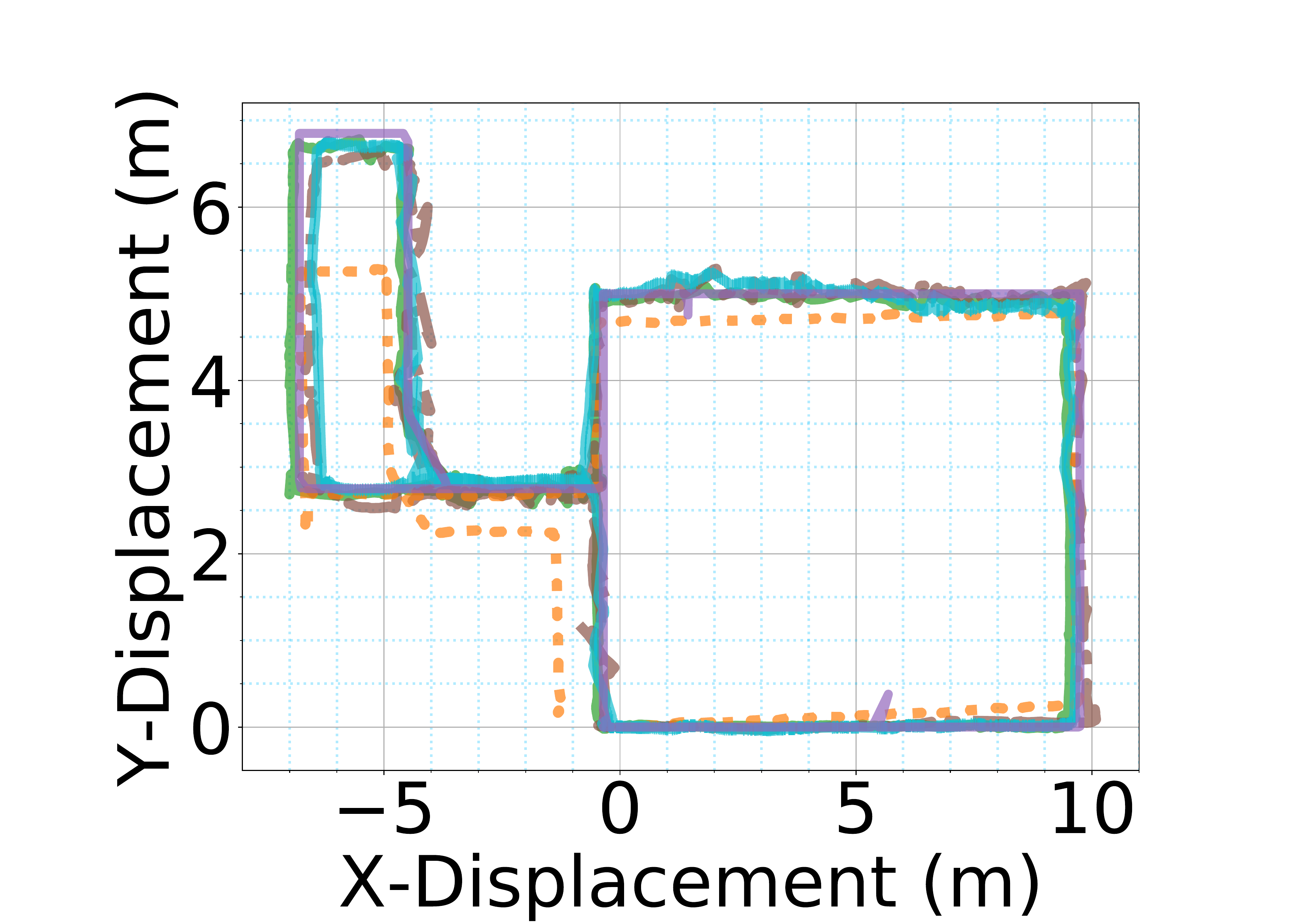} 
     }
    \subfloat[Unseen Environment]{
      \includegraphics[width=0.5\linewidth]{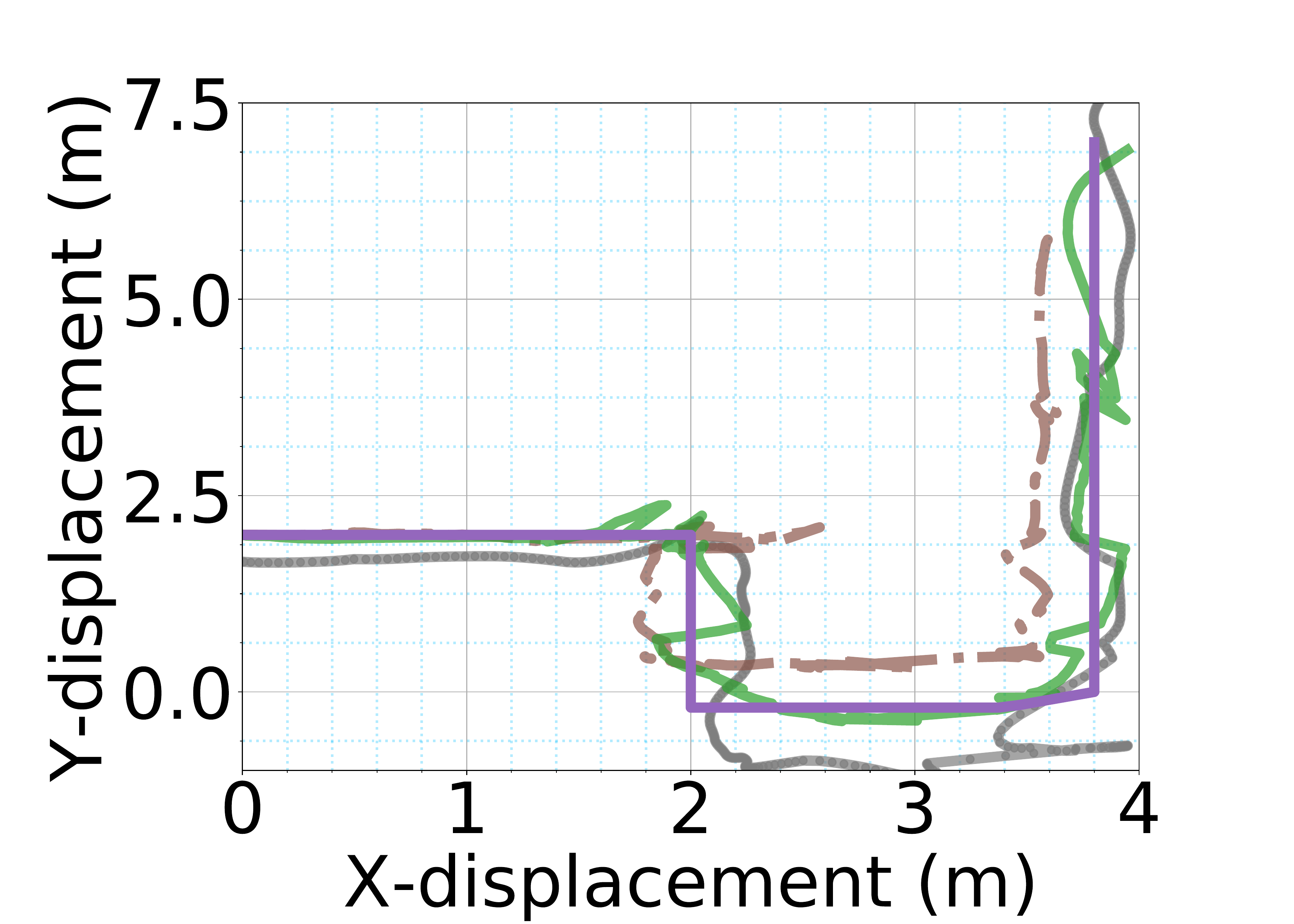}  
     }
  \caption{\system Vs. Baseline in practical, Trained (left) and Unseen environment (right).}
\label{fig:trajectory-unseen}
\vspace{-0.5cm}
\end{figure}

\subsection{Robustness in Unseen Environments} \label{sec:robustunseen}
In order to test \system's ability to generalize in untrained (unseen) environments, we evaluate its performance in completely different input distributions (new places) that are kept aside while training. We deploy two separate testbeds - one in a separate 
office environment and another in a home environment - for this purpose.
A more thorough evaluation for these testbeds can be found in our technical report~\cite{tech-report}.

\noindent \textbf{Favorable conditions:} 
As before, we begin evaluating with favorable conditions, i.e., ensure  
visual scenery has rich texture, good lighting and the RF anchors are all in LoS. Fig. \ref{fig:comprehensive-unseen}a-b shows that while RF, VO and EKF algorithms' performance
are similar to the trained environments due to their environmental-agnostic nature, 
BB and VINET solutions' performance deteriorates significantly (>$2\times$ degradation compared to trained environments), with the top 10\% of points having errors $>$1m. Clearly, BB and VINET are unable to generalize and perform robustly in untrained environments. On the other hand, \system has a median error of only {$\approx$20cm, 15cm in Home and Office - 4} (similar to its performance in trained environments) and a maximum error that is $<$ 0.75cm ($\approx$ 230\% increase in median-accuracy). {\em \system's strategy of leveraging RF and VO algorithms for extracting tracking features, allows its sensor fusion to perform robustly even in untrained environments with high accuracy.}

\noindent \textbf{Practical conditions:}
Untrained environments with practical conditions are perhaps the most challenging. To highlight \system's robustness, Fig.~\ref{fig:trajectory-unseen} compares two trajectories: one trained (left) and another unseen (right). We find that \system shows a consistent behavior in diverse scenarios (trained/unseen as well as favorable/practical conditions), while the alternatives suffer in one or the other scenarios. The corresponding error CDFs for unseen case are shown in Fig.~\ref{fig:comprehensive-unseen}c-d. All of the alternatives suffer in this case because of the noise from practical conditions for EKF (RF+VO), and limited generalizability of BB/VINET. However, \system offers a median error of only 18cm (office-3) and 23cm (Home) even in these adverse cases, showing an evidence that {\em its hybrid fusion strategy gracefully delivers in diverse environments}.

\begin{figure}
\vspace{-0.2cm}
\includegraphics[width=\linewidth]{figs/legend.png}
\centering
    \subfloat[Trained, Office, 2 Users]{%
      \includegraphics[width=0.45\linewidth]{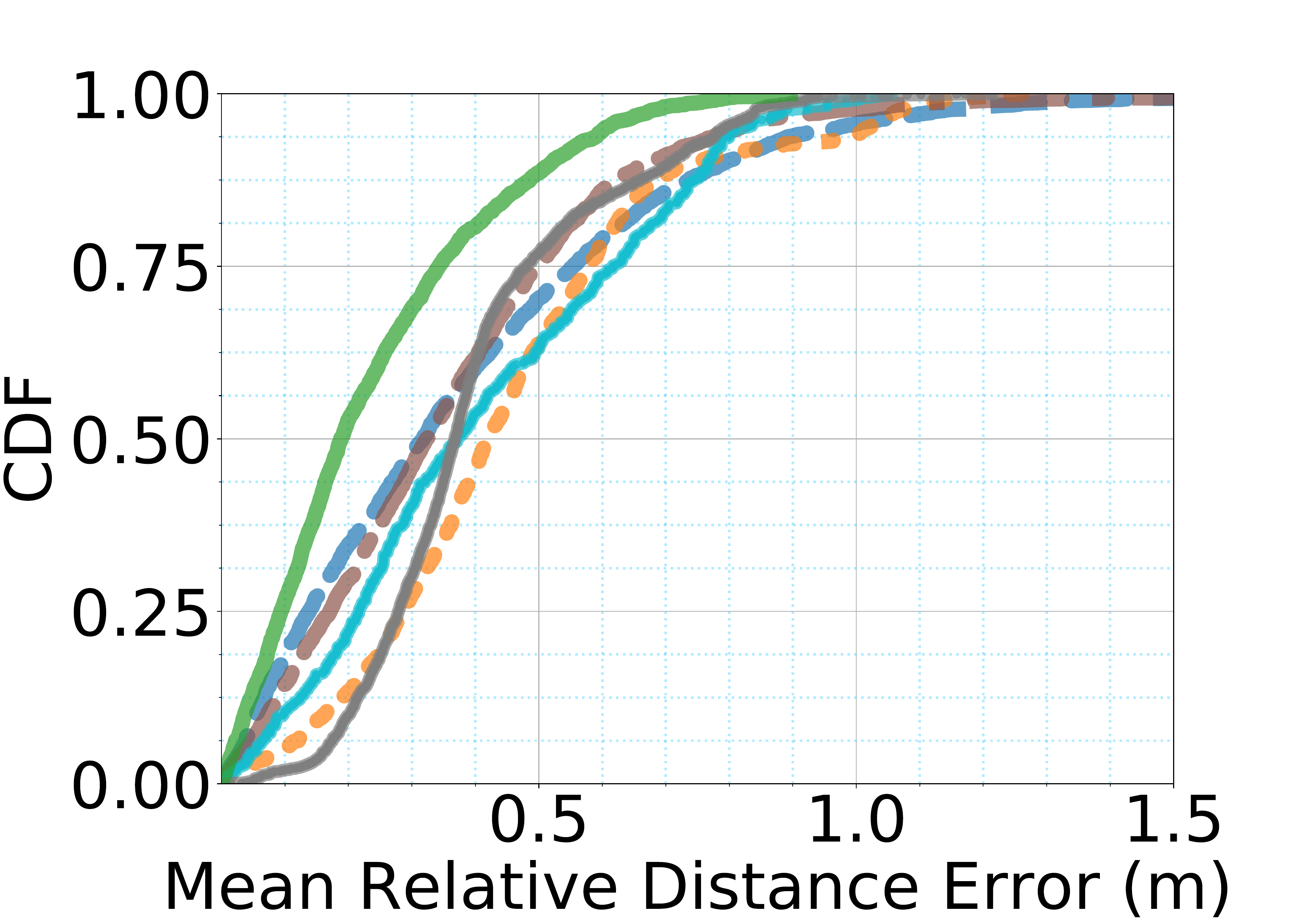} 
     }
    \subfloat[Unseen (Home), 3 Users]{
      \includegraphics[width=0.45\linewidth]{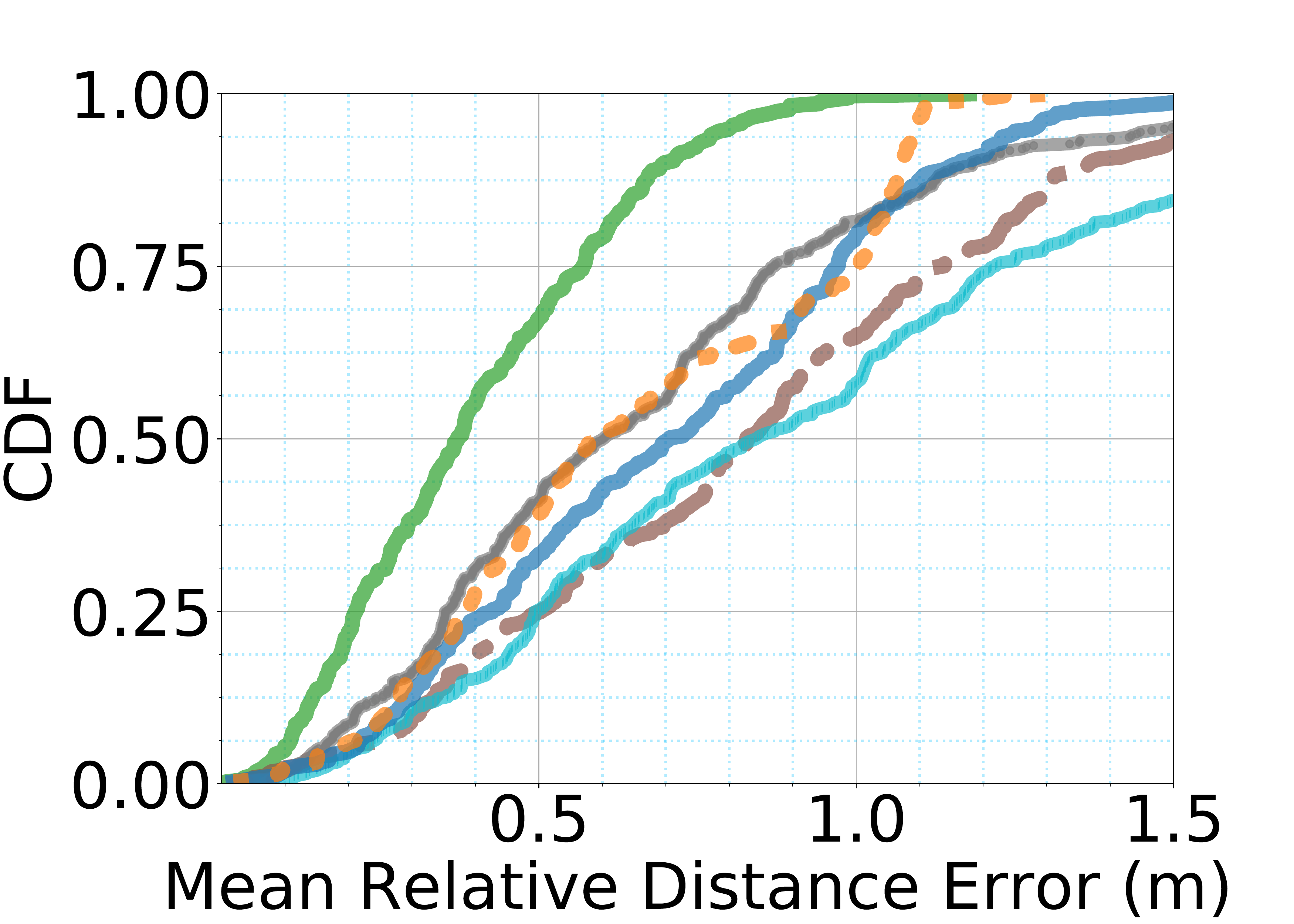}  
     }
     
    \subfloat[Trained, Office, 2 Users]{%
      \includegraphics[width=0.45\linewidth]{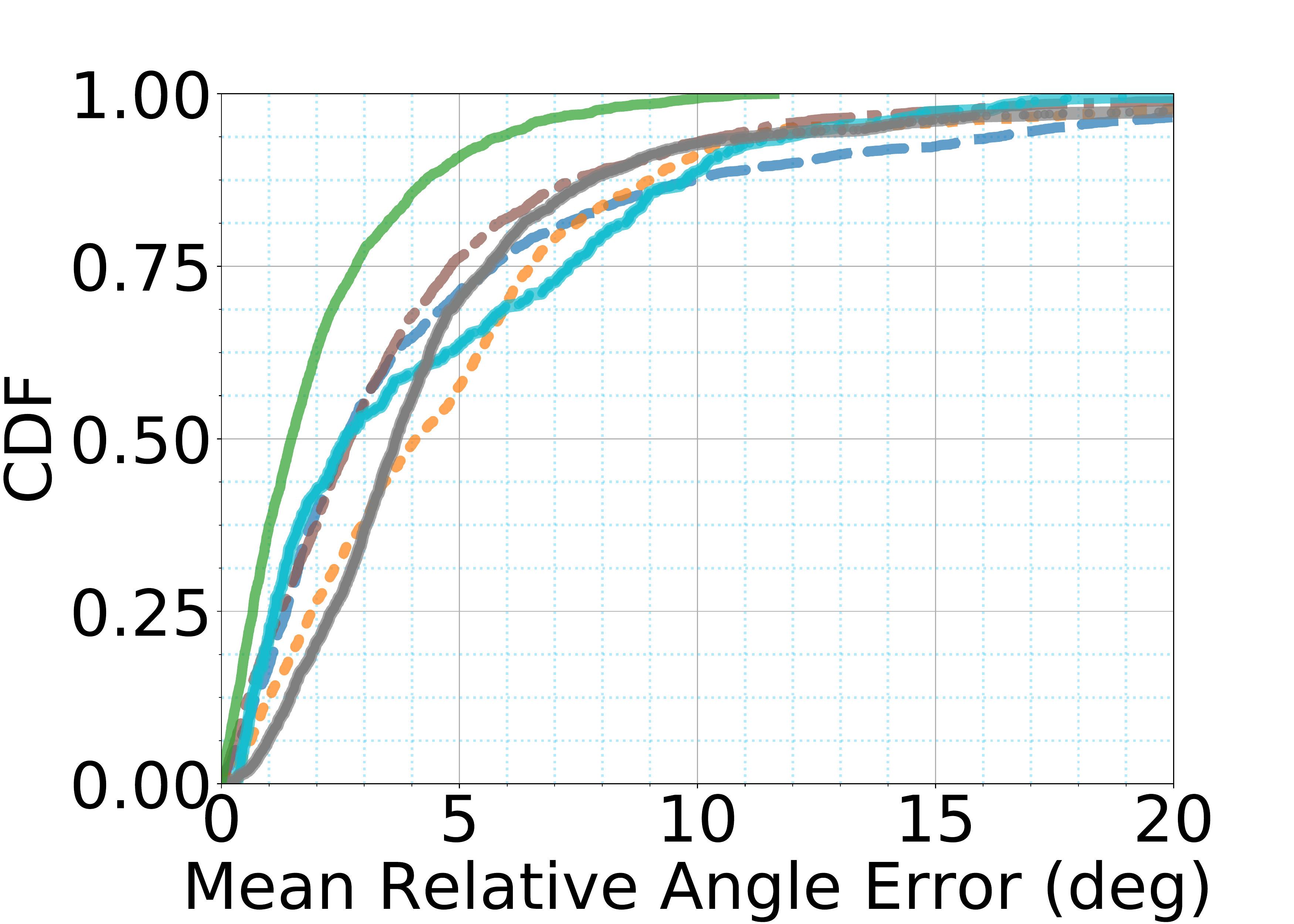} 
     }
    \subfloat[Unseen (Home), 3 Users]{%
      \includegraphics[width=0.45\linewidth]{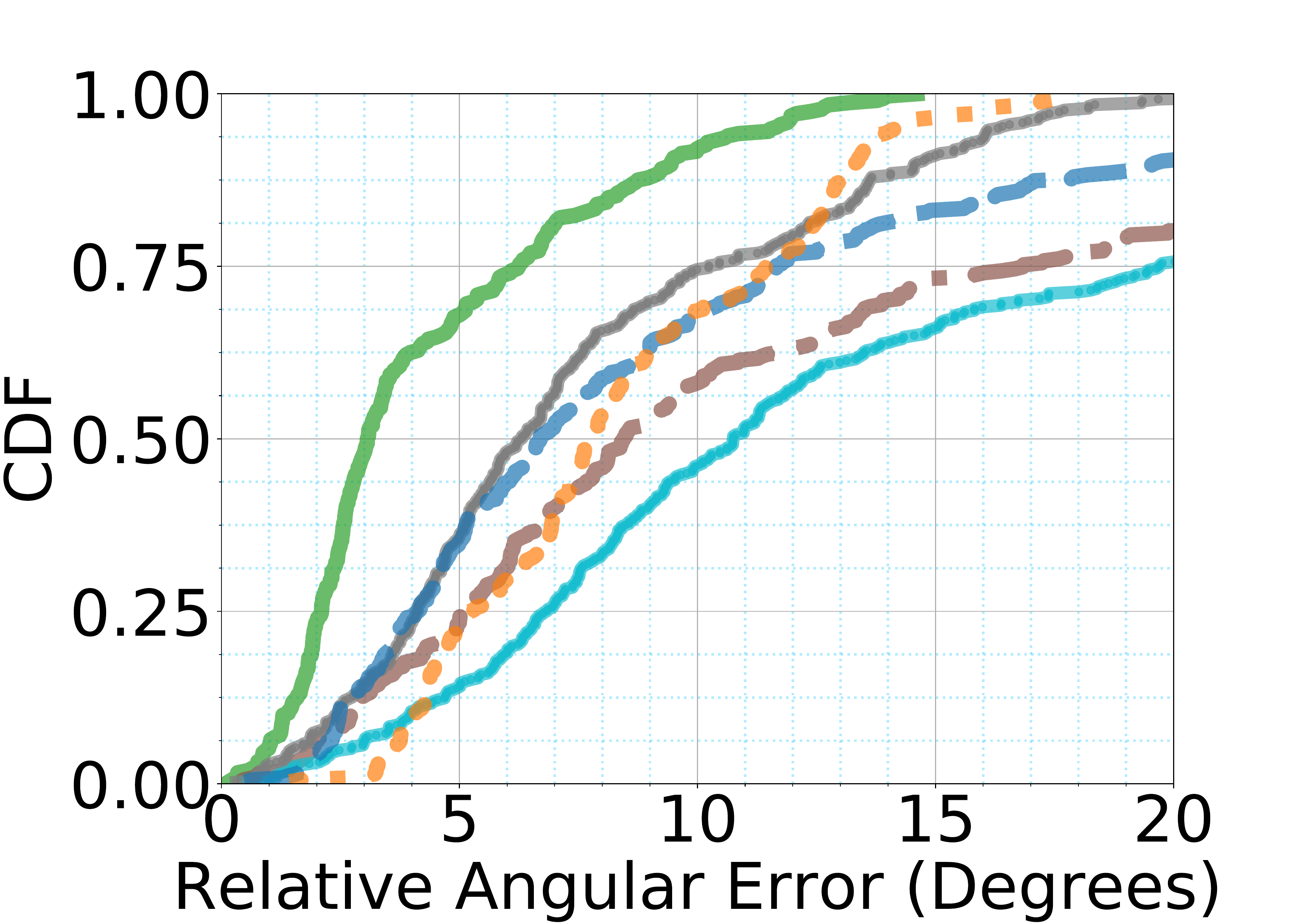} 
     }
     \vspace{-0.12in}
     \caption{\system performance for multi-users. Relative distance error (a,b) and Relative angular error (c,d)}
     \label{fig:multi-user}
    \vspace{-0.2in}
\end{figure}

\subsection{Scalability across Multiple Agents}
Keeping collaborating applications in mind, we evaluate \system's ability
to track multiple agents (users) with respect to each other. We have up to three users moving simultaneously along the same trajectory, but in different directions. To capture the tracking accuracy across multiple users simultaneously, we calculate the mean error in the relative positions for every pair of users. For every user-pair, this in turn, is captured jointly by the error in their relative distance and angle compared to ground truth. 
In the interest of space, we show the results (Fig.~\ref{fig:multi-user}) for only two environments (trained office and unseen home environment) in 
practical conditions. Results of other scenarios are in~\cite{tech-report}.
{\em \system, with its ability to absolutely localize every user (with help of anchors), and 
leverage dual-layer diversity, performs best with 30cm, 35cm relative-distance, 2.5$^{\circ}$, 3$^{\circ}$
relative-angular errors in trained and unseen environments, respectively. With gains increasing for more users,  it can scale to collaborative 
applications easily even in unseen environments.} This is a
1.2$\times$ (distance) and 1.6$\times$ (angle) improvement over BB and VINET respectively. 
RF, VO and EKF's relative-distance error is better than BB and VINET owing to better robustness, but less than \system
by 36\%, 41\% and 35\% respectively. 

\subsection{Ablation Study}
\begin{wrapfigure}{r}{0.2\textwidth}
 \vspace{-0.25in}
 \begin{center}
    \includegraphics[width=0.9\linewidth]{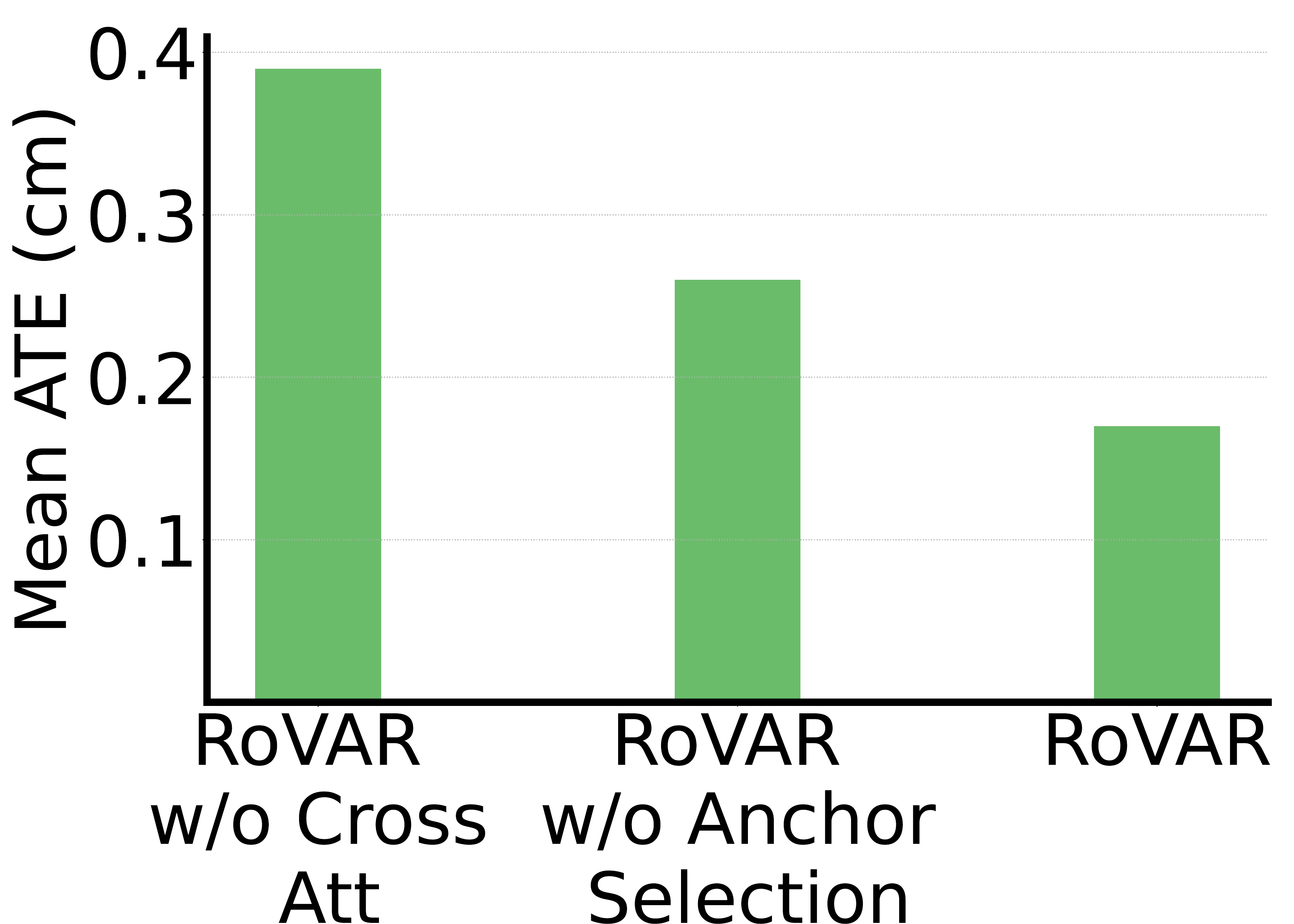}
 \end{center}
 \vspace{-0.15in}
 \caption{Breakdown of \system's individual components.}
 \vspace{-0.1in}
 \label{fig:ablation}
\end{wrapfigure}

We show an ablation study to understand the performance benefits individual components in \system. Fig. \ref{fig:ablation} shows the impact of \system's individual components: 1) \system without 
cross-attention mechanism (i.e., directly merging features to predict the final location), 2) \system without anchor filtering. The figure shows that both components are critical to the performance of \system. Similarly, \system without anchor selection performs poorly because of the influence of NLoS anchors' inaccurate ranging is affecting the localization accuracy. In case of \system without cross-attention, the system performs poorly because the features from {\em both} the sensors are merged to predict the location even when one of them is facing bad environmental conditions.

\begin{figure}
\centering
      \includegraphics[width=0.9\linewidth]{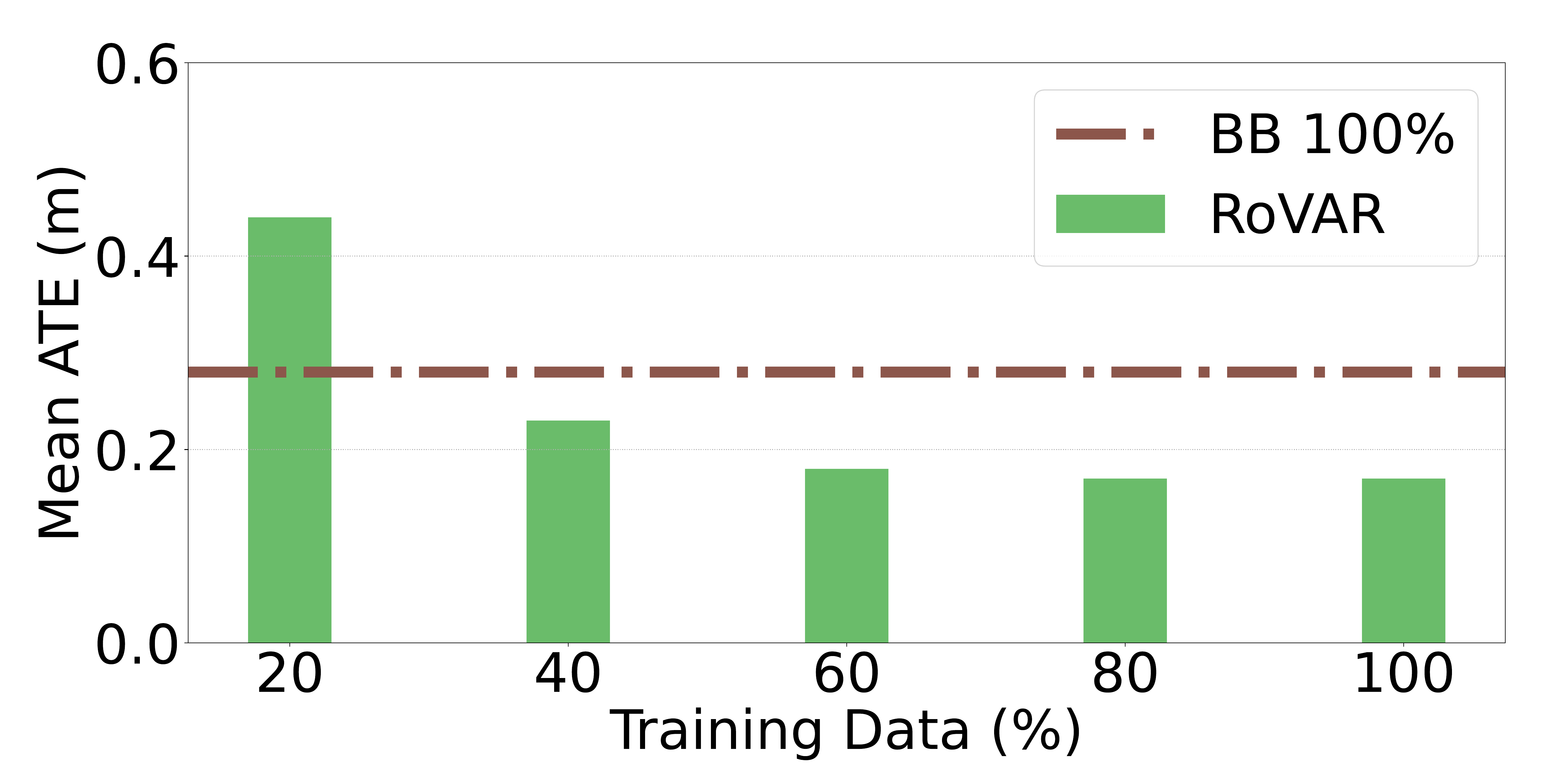}
      \vspace{-0.4cm}
      \caption{Training data size Vs. Performance.}
     \label{fig:trainingdata}
\vspace{-0.4cm}
\end{figure} 

\subsection{Low Training Requirements}
In addition to bringing robustness to new environments, {\em the reduced burden on \system's fusion module,
enables it to operate at a significantly reduced training cost compared to pure ML-driven solutions.} 
Fig.~\ref{fig:trainingdata} shows \system's performance as an increasing function of training data size (out of entire dataset available to BB).
\system is able to reduce the training size requirements by over 60\%, while still delivering a comparable performance. 

\subsection{Real-time System Performance and Practicality on Mobile Platforms}
\label{sec:realtime}

\begin{figure}[t]
\centering
\vspace{-0.2cm}
    \subfloat[Latency vs. ORB Features]{%
    \includegraphics[width=0.49\linewidth]{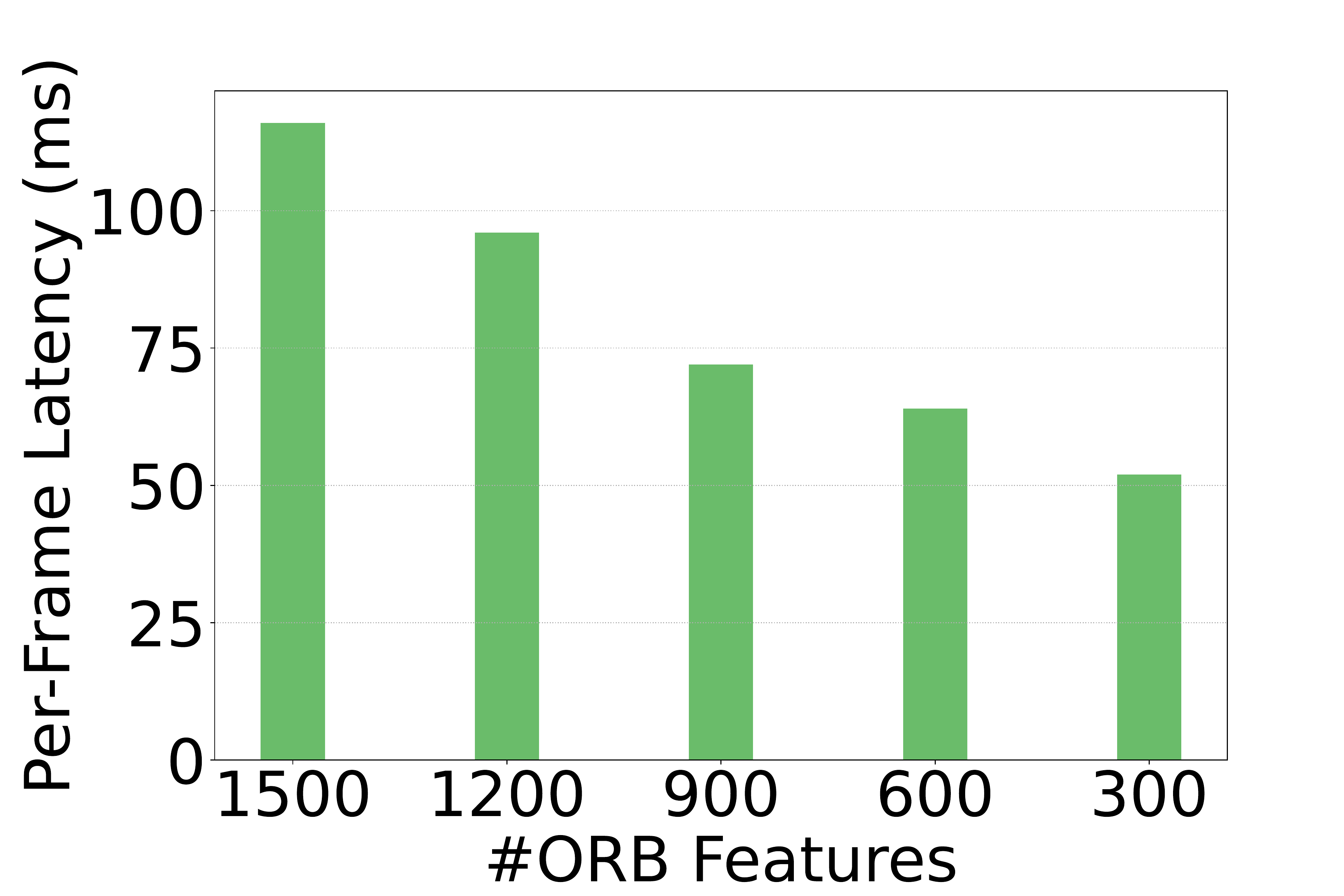}\label{fig:errorvlosnlos}
    }
    \subfloat[Accuracy vs. ORB Features]{%
      \includegraphics[width=0.49\linewidth]{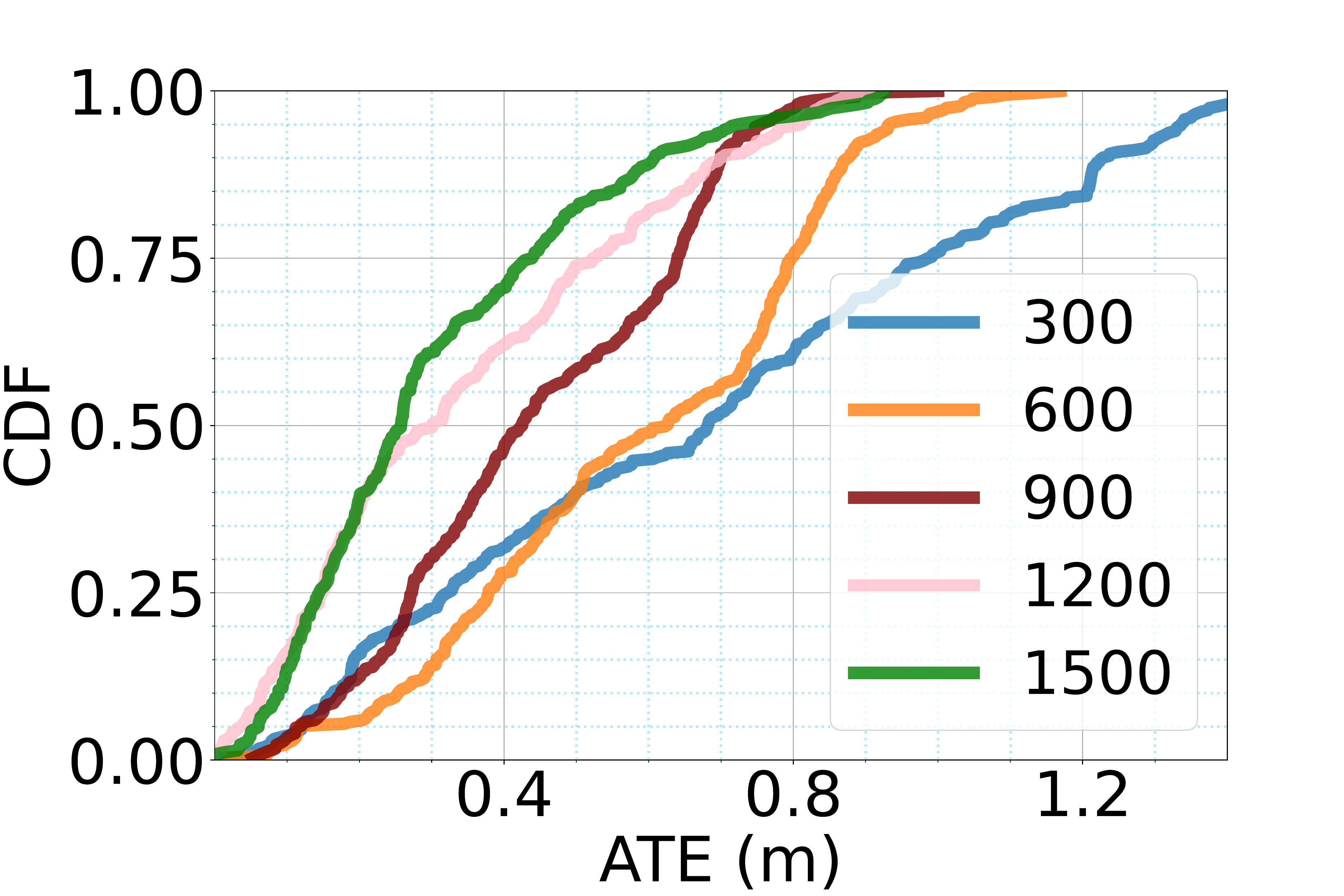} \label{fig:besterror}
     }
         \vspace{-0.2cm}
    \caption{Latency and Tracking accuracy trade-off with ORB features in ORBSLAM3.}
    \vspace{-0.4cm}
    \label{fig:tradeoff}
\end{figure} 

\system's fusion model requires
significantly less memory and processing power making it extremely lightweight compared to BB model. Table~\ref{tab:syseff} highlights the effectiveness of \system in multiple fronts--- {\em latency, memory and power consumption} on mobile platforms (Jetson Nano/TX2). Moreover, the BB model is too memory intensive to run on these low-end devices unless we downsample it to 480p resolution from 848x800. Even then, the running time of BB model is 5X more (348ms) than \system. More importantly, \system has 3$\times$ less power consumption compared to BB model, which makes \system a well suited ML-driven fusion methodology that can be deployed on power constrained mobile devices in practice. 
\begin{wrapfigure}{r}{0.25\textwidth}
 \vspace{-0.05in}
 \begin{center}
    \includegraphics[width=0.9\linewidth]{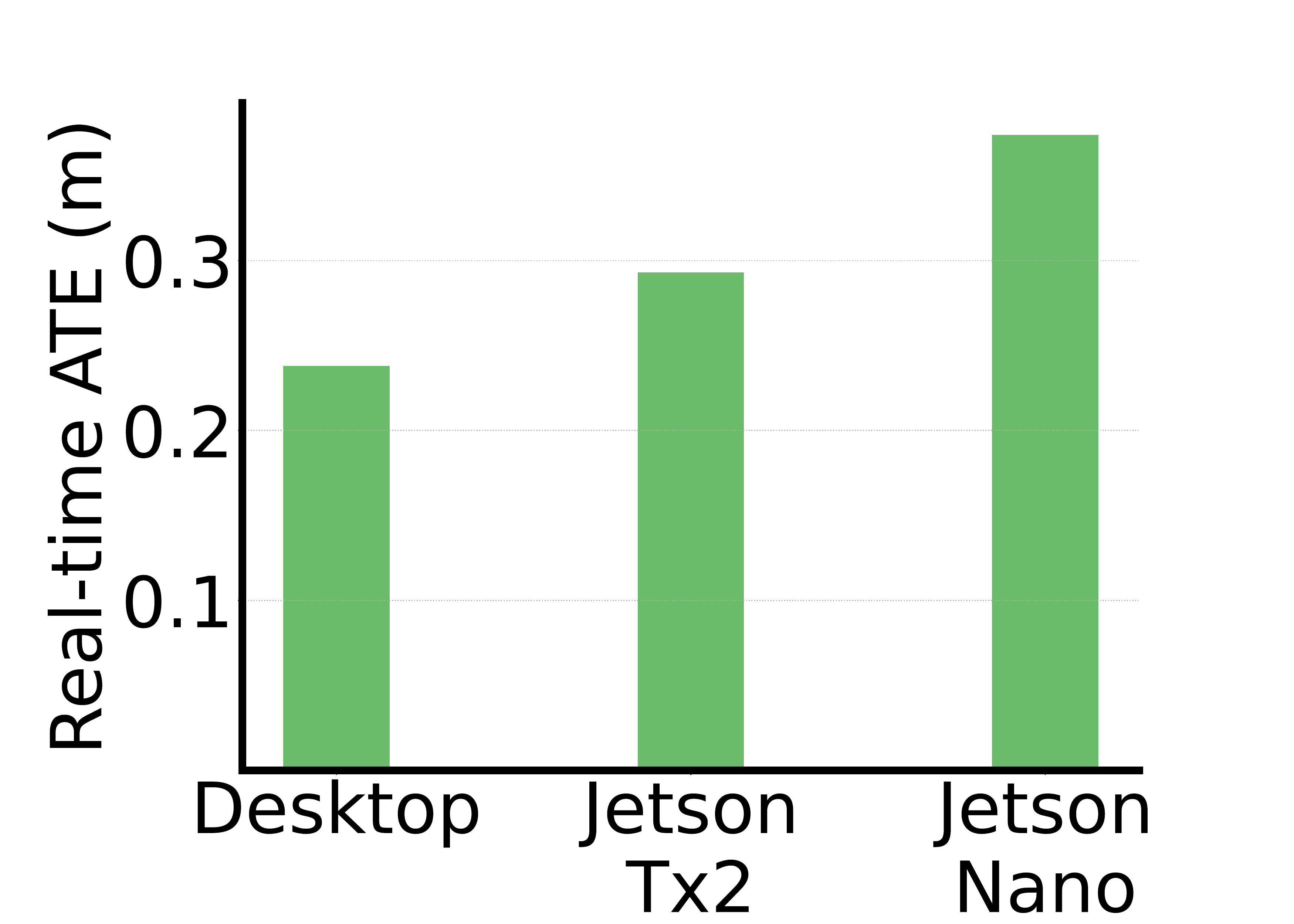}
 \end{center}
 \vspace{-0.2in}
 \caption{\system's real-time accuracy.}
 \vspace{-0.15in}
 \label{fig:realtimeaccuracy}
\end{wrapfigure}
However, the VO (ORBSLAM3) algorithm, the main overhead contributor to the \system pipeline
can afford to use 900 ORB features (for latency 75 ms -- see Fig.~\ref{fig:tradeoff}a) instead of the optimal choice of 1500 ORB features, resulting in \system's overall latency to be around 90ms (20ms more than TX2). Adopting VO optimizations such as EdgeSLAM~\cite{ben2020edge} over edge-cloud networks can further reduce the latency. Nonetheless, the drop in overall median error (Fig.~\ref{fig:besterror}) due to this setting is only 15cm resulting in \system's median ATE of 38cm (Fig. \ref{fig:realtimeaccuracy}) in untrained home environment, which is still better the alternatives (as seen in \S\ref{sec:robustunseen}). This performance highlights that \system can be deployed in practice easily compared to BB models that are difficult to realize on low-end platforms with high accuracy in unseen places.

\begin{table}[t]
  \centering
  \caption{Latency, Memory, and Energy usage ($^*$video is downsampled to 480p for BB because it cannot run on low-end devices. The reported power consumption is measured on Jetson Nano platform using its i2c interface power rails (e.g., \cite{powermonitor2022})).}
    \vspace{-0.2cm}
  \scalebox{0.83}{
  \begin{tabular}{|l|c|c|c|c|c|}
  \hline
      \textbf{ } & \textbf{RF} & \textbf{VO} & \textbf{BB} & \textbf{\system} & \textbf{\system} \\
       &   &   &   & \textbf{Fusion} & \textbf{Overall} \\
    \hline
    Desktop (ms)  & 1.80 & 43.67     & 126.91    & 2.61  & 48.08 \\ \hline
    Jetson TX2 (ms)  & 3.20 & 61.67  & 348$^*$ & 5.34 & 70.21\\ \hline
    Jetson Nano (ms)  & 4.62 & 118.34  & 596$^*$ & 9.50 & 132.46\\ \hline
    Model/Binary size & 13KB  &  100KB & 341.15MB & 5.05MB & 5.28MB \\ \hline 
    Power (Nano) & 150mW & 320mW& 2620mW$^*$& 340mW& 810mW\\ \hline
  \end{tabular}
    }
  \label{tab:syseff}
  \vspace{-0.3cm}
\end{table}


\section{Conclusion}
\label{CONCL}
We tackled the problem of bringing robustness and accuracy to multi-agent tracking in practical, indoor environments, by introducing the framework of dual-layer diversity in sensor fusion. 
\system, an embodiment of this framework, brought together the multi-agent and high-resolution tracking benefits of active (UWB) and passive (visual odometry) tracking modalities through an intelligent combination of both algorithmic and data-driven techniques to deliver robustness in everyday environments. \system's comprehensive evaluation showcased its significant benefits over prior approaches that rely on a single layer of diversity. We believe \system's hybrid, dual-layer diversity approach to sensor fusion offers an important step in opening the door for exciting multi-agent collaborative applications in everyday indoor environments.

\bibliographystyle{ACM-Reference-Format}
\bibliography{form}


\begin{thebibliography}{63}


\ifx \showCODEN    \undefined \def \showCODEN     #1{\unskip}     \fi
\ifx \showDOI      \undefined \def \showDOI       #1{#1}\fi
\ifx \showISBNx    \undefined \def \showISBNx     #1{\unskip}     \fi
\ifx \showISBNxiii \undefined \def \showISBNxiii  #1{\unskip}     \fi
\ifx \showISSN     \undefined \def \showISSN      #1{\unskip}     \fi
\ifx \showLCCN     \undefined \def \showLCCN      #1{\unskip}     \fi
\ifx \shownote     \undefined \def \shownote      #1{#1}          \fi
\ifx \showarticletitle \undefined \def \showarticletitle #1{#1}   \fi
\ifx \showURL      \undefined \def \showURL       {\relax}        \fi
\providecommand\bibfield[2]{#2}
\providecommand\bibinfo[2]{#2}
\providecommand\natexlab[1]{#1}
\providecommand\showeprint[2][]{arXiv:#2}

\bibitem[\protect\citeauthoryear{??}{dw1}{2020a}]%
        {dw1000}
 \bibinfo{year}{2020}\natexlab{a}.
\newblock \bibinfo{title}{{Decawave DW1000 USER MANUAL}}.
\newblock
  \bibinfo{howpublished}{\url{https://www.decawave.com/sites/default/files/resources/dw1000_user_manual_2.11.pdf}}.
    (\bibinfo{year}{2020}).
\newblock


\bibitem[\protect\citeauthoryear{??}{dw1}{2020b}]%
        {dw1000evk}
 \bibinfo{year}{2020}\natexlab{b}.
\newblock \bibinfo{title}{{Decawave UWB Two-Way Ranging}}.
\newblock
  \bibinfo{howpublished}{\url{https://www.decawave.com/product/evk1000-evaluation-kit/}}.
    (\bibinfo{year}{2020}).
\newblock


\bibitem[\protect\citeauthoryear{??}{int}{2020}]%
        {intelrealsenset265}
 \bibinfo{year}{2020}\natexlab{}.
\newblock \bibinfo{title}{{Intel RealSense Tracking Camera T265}}.
\newblock
  \bibinfo{howpublished}{\url{https://www.intelrealsense.com/tracking-camera-t265/}}.
    (\bibinfo{year}{2020}).
\newblock


\bibitem[\protect\citeauthoryear{??}{rfi}{2020}]%
        {rfidreader}
 \bibinfo{year}{2020}\natexlab{}.
\newblock \bibinfo{title}{{M6E Nano RFID Reader}}.
\newblock
  \bibinfo{howpublished}{\url{https://www.sparkfun.com/products/14066}}.
  (\bibinfo{year}{2020}).
\newblock


\bibitem[\protect\citeauthoryear{??}{jet}{2020}]%
        {jetsontx2}
 \bibinfo{year}{2020}\natexlab{}.
\newblock \bibinfo{title}{{Nvidia Jetson TX2 Module}}.
\newblock
  \bibinfo{howpublished}{\url{https://developer.nvidia.com/embedded/jetson-tx2}}.
    (\bibinfo{year}{2020}).
\newblock


\bibitem[\protect\citeauthoryear{??}{app}{2021}]%
        {appleu1chip}
 \bibinfo{year}{2021}\natexlab{}.
\newblock \bibinfo{title}{Apple's UWB Support and U1 chip}.
\newblock
  \bibinfo{howpublished}{\url{https://support.apple.com/en-us/HT212274}}.
  (\bibinfo{year}{2021}).
\newblock


\bibitem[\protect\citeauthoryear{??}{sam}{2021}]%
        {sammobileuwbchip}
 \bibinfo{year}{2021}\natexlab{}.
\newblock \bibinfo{title}{Samsung Galaxy S21+ and UWB}.
\newblock
  \bibinfo{howpublished}{\url{https://www.sammobile.com/news/uwb-explained-galaxy-s21-plus-s21-ultra}}.
    (\bibinfo{year}{2021}).
\newblock


\bibitem[\protect\citeauthoryear{??}{nxp}{2021}]%
        {nxpsinglechip}
 \bibinfo{year}{2021}\natexlab{}.
\newblock \bibinfo{title}{Single chip UWB SR040 UWB Module}.
\newblock
  \bibinfo{howpublished}{\url{https://www.nxp.com/products/wireless/secure-ultra-wideband-uwb/trimension-sr040-uwb-module-with-embedded-rf-connector-asmop1co0r1:ASMOP1CO0R1}}.
    (\bibinfo{year}{2021}).
\newblock


\bibitem[\protect\citeauthoryear{??}{ark}{2022}]%
        {arkitapple}
 \bibinfo{year}{2022}\natexlab{}.
\newblock \bibinfo{title}{{Apple ARKit}}.
\newblock
  \bibinfo{howpublished}{\url{https://developer.apple.com/augmented-reality/arkit/}}.
    (\bibinfo{year}{2022}).
\newblock


\bibitem[\protect\citeauthoryear{??}{pow}{2022}]%
        {powermonitor2022}
 \bibinfo{year}{2022}\natexlab{}.
\newblock \bibinfo{title}{{Jetson Nano/TX2 Power Measurements}}.
\newblock
  \bibinfo{howpublished}{\url{https://github.com/leonardopsantos/jetsonTX2Power}}.
    (\bibinfo{year}{2022}).
\newblock


\bibitem[\protect\citeauthoryear{??}{vid}{2022}]%
        {video-link}
 \bibinfo{year}{2022}\natexlab{}.
\newblock \bibinfo{title}{RoVAR - Demo Video}.
\newblock
  \bibinfo{howpublished}{{\url{https://www.dropbox.com/sh/qeuabbq1lz97bpg/AADa0G9zLEehjkny7mHtUHIOa?dl=0&preview=RoVAR.mp4}}}.
    (\bibinfo{year}{2022}).
\newblock


\bibitem[\protect\citeauthoryear{??}{tec}{2022}]%
        {tech-report}
 \bibinfo{year}{2022}\natexlab{}.
\newblock \bibinfo{title}{RoVAR - Tech Report}.
\newblock
  \bibinfo{howpublished}{{\url{https://www.dropbox.com/sh/qeuabbq1lz97bpg/AADa0G9zLEehjkny7mHtUHIOa?dl=0&preview=paper.pdf}}}.
    (\bibinfo{year}{2022}).
\newblock


\bibitem[\protect\citeauthoryear{Ayyalasomayajula, Arun, Wu, Sharma, Sethi,
  Vasisht, and Bharadia}{Ayyalasomayajula et~al\mbox{.}}{2020}]%
        {ayyalasomayajula2020deep}
\bibfield{author}{\bibinfo{person}{Roshan Ayyalasomayajula},
  \bibinfo{person}{Aditya Arun}, \bibinfo{person}{Chenfeng Wu},
  \bibinfo{person}{Sanatan Sharma}, \bibinfo{person}{Abhishek~Rajkumar Sethi},
  \bibinfo{person}{Deepak Vasisht}, {and} \bibinfo{person}{Dinesh Bharadia}.}
  \bibinfo{year}{2020}\natexlab{}.
\newblock \showarticletitle{Deep learning based wireless localization for
  indoor navigation}. In \bibinfo{booktitle}{{\em Proceedings of the 26th
  Annual International Conference on Mobile Computing and Networking}}.
  \bibinfo{pages}{1--14}.
\newblock


\bibitem[\protect\citeauthoryear{Ben~Ali, Hashemifar, and Dantu}{Ben~Ali
  et~al\mbox{.}}{2020}]%
        {ben2020edge}
\bibfield{author}{\bibinfo{person}{Ali~J Ben~Ali},
  \bibinfo{person}{Zakieh~Sadat Hashemifar}, {and} \bibinfo{person}{Karthik
  Dantu}.} \bibinfo{year}{2020}\natexlab{}.
\newblock \showarticletitle{Edge-SLAM: edge-assisted visual simultaneous
  localization and mapping}. In \bibinfo{booktitle}{{\em Proceedings of the
  18th International Conference on Mobile Systems, Applications, and
  Services}}. \bibinfo{pages}{325--337}.
\newblock


\bibitem[\protect\citeauthoryear{Bloesch, Omari, Hutter, and Siegwart}{Bloesch
  et~al\mbox{.}}{2015}]%
        {bloesch2015robust}
\bibfield{author}{\bibinfo{person}{Michael Bloesch}, \bibinfo{person}{Sammy
  Omari}, \bibinfo{person}{Marco Hutter}, {and} \bibinfo{person}{Roland
  Siegwart}.} \bibinfo{year}{2015}\natexlab{}.
\newblock \showarticletitle{Robust visual inertial odometry using a direct
  EKF-based approach}. In \bibinfo{booktitle}{{\em 2015 IEEE/RSJ international
  conference on intelligent robots and systems (IROS)}}. IEEE,
  \bibinfo{pages}{298--304}.
\newblock


\bibitem[\protect\citeauthoryear{Campos, Elvira, Rodr{\'\i}guez, Montiel, and
  Tard{\'o}s}{Campos et~al\mbox{.}}{2021}]%
        {campos2021orb}
\bibfield{author}{\bibinfo{person}{Carlos Campos}, \bibinfo{person}{Richard
  Elvira}, \bibinfo{person}{Juan J~G{\'o}mez Rodr{\'\i}guez},
  \bibinfo{person}{Jos{\'e}~MM Montiel}, {and} \bibinfo{person}{Juan~D
  Tard{\'o}s}.} \bibinfo{year}{2021}\natexlab{}.
\newblock \showarticletitle{ORB-SLAM3: An Accurate Open-Source Library for
  Visual, Visual--Inertial, and Multimap SLAM}.
\newblock \bibinfo{journal}{{\em IEEE Transactions on Robotics\/}}
  (\bibinfo{year}{2021}).
\newblock


\bibitem[\protect\citeauthoryear{Chen, Rosa, Miao, Lu, Wu, Markham, and
  Trigoni}{Chen et~al\mbox{.}}{2019}]%
        {chen2019selective}
\bibfield{author}{\bibinfo{person}{Changhao Chen}, \bibinfo{person}{Stefano
  Rosa}, \bibinfo{person}{Yishu Miao}, \bibinfo{person}{Chris~Xiaoxuan Lu},
  \bibinfo{person}{Wei Wu}, \bibinfo{person}{Andrew Markham}, {and}
  \bibinfo{person}{Niki Trigoni}.} \bibinfo{year}{2019}\natexlab{}.
\newblock \bibinfo{title}{Selective Sensor Fusion for Neural Visual-Inertial
  Odometry}.
\newblock   (\bibinfo{year}{2019}).
\newblock
\showeprint[arxiv]{cs.CV/1903.01534}


\bibitem[\protect\citeauthoryear{Cheung, So, Ma, and Chan}{Cheung
  et~al\mbox{.}}{2006}]%
        {cheung2006constrained}
\bibfield{author}{\bibinfo{person}{Ka~Wai Cheung}, \bibinfo{person}{Hing-Cheung
  So}, \bibinfo{person}{Wing-Kin Ma}, {and} \bibinfo{person}{Yiu-Tong Chan}.}
  \bibinfo{year}{2006}\natexlab{}.
\newblock \showarticletitle{A constrained least squares approach to mobile
  positioning: algorithms and optimality}.
\newblock \bibinfo{journal}{{\em EURASIP Journal on Advances in Signal
  Processing\/}}  \bibinfo{volume}{2006} (\bibinfo{year}{2006}),
  \bibinfo{pages}{1--23}.
\newblock


\bibitem[\protect\citeauthoryear{Clark, Wang, Wen, Markham, and Trigoni}{Clark
  et~al\mbox{.}}{2017}]%
        {clark2017vinet}
\bibfield{author}{\bibinfo{person}{Ronald Clark}, \bibinfo{person}{Sen Wang},
  \bibinfo{person}{Hongkai Wen}, \bibinfo{person}{Andrew Markham}, {and}
  \bibinfo{person}{Niki Trigoni}.} \bibinfo{year}{2017}\natexlab{}.
\newblock \showarticletitle{Vinet: Visual-inertial odometry as a
  sequence-to-sequence learning problem}. In \bibinfo{booktitle}{{\em
  Proceedings of the AAAI Conference on Artificial Intelligence}},
  Vol.~\bibinfo{volume}{31}.
\newblock


\bibitem[\protect\citeauthoryear{Constandache, Agarwal, Tashev, and
  Choudhury}{Constandache et~al\mbox{.}}{2014}]%
        {constandache2014daredevil}
\bibfield{author}{\bibinfo{person}{Ionut Constandache}, \bibinfo{person}{Sharad
  Agarwal}, \bibinfo{person}{Ivan Tashev}, {and} \bibinfo{person}{Romit~Roy
  Choudhury}.} \bibinfo{year}{2014}\natexlab{}.
\newblock \showarticletitle{Daredevil: indoor location using sound}.
\newblock \bibinfo{journal}{{\em ACM SIGMOBILE Mobile Computing and
  Communications Review\/}} \bibinfo{volume}{18}, \bibinfo{number}{2}
  (\bibinfo{year}{2014}), \bibinfo{pages}{9--19}.
\newblock


\bibitem[\protect\citeauthoryear{Corrales, Candelas, and Torres}{Corrales
  et~al\mbox{.}}{2008}]%
        {corrales2008hybrid}
\bibfield{author}{\bibinfo{person}{Juan~Antonio Corrales}, \bibinfo{person}{FA
  Candelas}, {and} \bibinfo{person}{Fernando Torres}.}
  \bibinfo{year}{2008}\natexlab{}.
\newblock \showarticletitle{Hybrid tracking of human operators using IMU/UWB
  data fusion by a Kalman filter}. In \bibinfo{booktitle}{{\em 2008 3rd
  ACM/IEEE International Conference on Human-Robot Interaction (HRI)}}. IEEE,
  \bibinfo{pages}{193--200}.
\newblock


\bibitem[\protect\citeauthoryear{Feigl, Porada, Steiner, L{\"o}ffler,
  Mutschler, and Philippsen}{Feigl et~al\mbox{.}}{2020}]%
        {feigl2020localization}
\bibfield{author}{\bibinfo{person}{Tobias Feigl}, \bibinfo{person}{Andreas
  Porada}, \bibinfo{person}{Steve Steiner}, \bibinfo{person}{Christoffer
  L{\"o}ffler}, \bibinfo{person}{Christopher Mutschler}, {and}
  \bibinfo{person}{Michael Philippsen}.} \bibinfo{year}{2020}\natexlab{}.
\newblock \showarticletitle{Localization Limitations of ARCore, ARKit, and
  Hololens in Dynamic Large-scale Industry Environments.}. In
  \bibinfo{booktitle}{{\em VISIGRAPP (1: GRAPP)}}. \bibinfo{pages}{307--318}.
\newblock


\bibitem[\protect\citeauthoryear{Gururaj, Rajendra, Song, Law, and Cai}{Gururaj
  et~al\mbox{.}}{2017}]%
        {gururaj2017real}
\bibfield{author}{\bibinfo{person}{Karthikeyan Gururaj},
  \bibinfo{person}{Anojh~Kumaran Rajendra}, \bibinfo{person}{Yang Song},
  \bibinfo{person}{Choi~Look Law}, {and} \bibinfo{person}{Guofa Cai}.}
  \bibinfo{year}{2017}\natexlab{}.
\newblock \showarticletitle{Real-time identification of NLOS range measurements
  for enhanced UWB localization}. In \bibinfo{booktitle}{{\em 2017
  international conference on indoor positioning and indoor navigation
  (IPIN)}}. IEEE, \bibinfo{pages}{1--7}.
\newblock


\bibitem[\protect\citeauthoryear{Hashemifar, Adhivarahan, Balakrishnan, and
  Dantu}{Hashemifar et~al\mbox{.}}{2019}]%
        {hashemifar2019augmenting}
\bibfield{author}{\bibinfo{person}{Zakieh~S Hashemifar},
  \bibinfo{person}{Charuvahan Adhivarahan}, \bibinfo{person}{Anand
  Balakrishnan}, {and} \bibinfo{person}{Karthik Dantu}.}
  \bibinfo{year}{2019}\natexlab{}.
\newblock \showarticletitle{Augmenting visual SLAM with Wi-Fi sensing for
  indoor applications}.
\newblock \bibinfo{journal}{{\em Autonomous Robots\/}} \bibinfo{volume}{43},
  \bibinfo{number}{8} (\bibinfo{year}{2019}), \bibinfo{pages}{2245--2260}.
\newblock


\bibitem[\protect\citeauthoryear{Herath and Pathirana}{Herath and
  Pathirana}{2013}]%
        {herath2013optimal}
\bibfield{author}{\bibinfo{person}{Sanvidha~CK Herath} {and}
  \bibinfo{person}{Pubudu~N Pathirana}.} \bibinfo{year}{2013}\natexlab{}.
\newblock \showarticletitle{Optimal sensor arrangements in angle of arrival
  (AoA) and range based localization with linear sensor arrays}.
\newblock \bibinfo{journal}{{\em Sensors\/}} \bibinfo{volume}{13},
  \bibinfo{number}{9} (\bibinfo{year}{2013}), \bibinfo{pages}{12277--12294}.
\newblock


\bibitem[\protect\citeauthoryear{Hua, Hang, Yue, Hang, and Kan}{Hua
  et~al\mbox{.}}{2014}]%
        {hua2014geometrical}
\bibfield{author}{\bibinfo{person}{Zhu Hua}, \bibinfo{person}{Li Hang},
  \bibinfo{person}{Li Yue}, \bibinfo{person}{Long Hang}, {and}
  \bibinfo{person}{Zheng Kan}.} \bibinfo{year}{2014}\natexlab{}.
\newblock \showarticletitle{Geometrical constrained least squares estimation in
  wireless location systems}. In \bibinfo{booktitle}{{\em 2014 4th IEEE
  International Conference on Network Infrastructure and Digital Content}}.
  IEEE, \bibinfo{pages}{159--163}.
\newblock


\bibitem[\protect\citeauthoryear{Ibrahim, Liu, Jawahar, Nguyen, Gruteser,
  Howard, Yu, and Bai}{Ibrahim et~al\mbox{.}}{2018}]%
        {ibrahim2018verification}
\bibfield{author}{\bibinfo{person}{Mohamed Ibrahim}, \bibinfo{person}{Hansi
  Liu}, \bibinfo{person}{Minitha Jawahar}, \bibinfo{person}{Viet Nguyen},
  \bibinfo{person}{Marco Gruteser}, \bibinfo{person}{Richard Howard},
  \bibinfo{person}{Bo Yu}, {and} \bibinfo{person}{Fan Bai}.}
  \bibinfo{year}{2018}\natexlab{}.
\newblock \showarticletitle{Verification: Accuracy evaluation of WiFi fine time
  measurements on an open platform}. In \bibinfo{booktitle}{{\em Proceedings of
  the 24th Annual International Conference on Mobile Computing and
  Networking}}. \bibinfo{pages}{417--427}.
\newblock


\bibitem[\protect\citeauthoryear{Jayashree, Arumugam, Anusha, and
  Hariny}{Jayashree et~al\mbox{.}}{2006}]%
        {jayashree2006accuracy}
\bibfield{author}{\bibinfo{person}{LS Jayashree}, \bibinfo{person}{S Arumugam},
  \bibinfo{person}{M Anusha}, {and} \bibinfo{person}{AB Hariny}.}
  \bibinfo{year}{2006}\natexlab{}.
\newblock \showarticletitle{On the accuracy of centroid based multilateration
  procedure for location discovery in wireless sensor networks}. In
  \bibinfo{booktitle}{{\em 2006 IFIP international conference on wireless and
  optical communications networks}}. IEEE, \bibinfo{pages}{6--pp}.
\newblock


\bibitem[\protect\citeauthoryear{Jia, Jin, and Spanos}{Jia
  et~al\mbox{.}}{2014}]%
        {jia2014soundloc}
\bibfield{author}{\bibinfo{person}{Ruoxi Jia}, \bibinfo{person}{Ming Jin},
  {and} \bibinfo{person}{Costas~J Spanos}.} \bibinfo{year}{2014}\natexlab{}.
\newblock \showarticletitle{Soundloc: Acoustic method for indoor localization
  without infrastructure}.
\newblock \bibinfo{journal}{{\em arXiv preprint arXiv:1407.4409\/}}
  (\bibinfo{year}{2014}).
\newblock


\bibitem[\protect\citeauthoryear{Jung and Woo}{Jung and Woo}{2004}]%
        {jung2004ubitrack}
\bibfield{author}{\bibinfo{person}{Seokmin Jung} {and}
  \bibinfo{person}{Woontack Woo}.} \bibinfo{year}{2004}\natexlab{}.
\newblock \showarticletitle{UbiTrack: Infrared-based user Tracking System for
  indoor environment}.
\newblock \bibinfo{journal}{{\em International Conferece on Artificial Reality
  and Telexisitence (ICAT04)\/}} (\bibinfo{year}{2004}),
  \bibinfo{pages}{1345--1278}.
\newblock


\bibitem[\protect\citeauthoryear{Kotaru, Joshi, Bharadia, and Katti}{Kotaru
  et~al\mbox{.}}{2015}]%
        {kotaru2015spotfi}
\bibfield{author}{\bibinfo{person}{Manikanta Kotaru}, \bibinfo{person}{Kiran
  Joshi}, \bibinfo{person}{Dinesh Bharadia}, {and} \bibinfo{person}{Sachin
  Katti}.} \bibinfo{year}{2015}\natexlab{}.
\newblock \showarticletitle{Spotfi: Decimeter level localization using wifi}.
  In \bibinfo{booktitle}{{\em Proceedings of the 2015 ACM Conference on Special
  Interest Group on Data Communication}}. \bibinfo{pages}{269--282}.
\newblock


\bibitem[\protect\citeauthoryear{Kumar, Marks, and Jones}{Kumar
  et~al\mbox{.}}{2014}]%
        {kumar2014improving}
\bibfield{author}{\bibinfo{person}{Suren Kumar}, \bibinfo{person}{Tim~K Marks},
  {and} \bibinfo{person}{Michael Jones}.} \bibinfo{year}{2014}\natexlab{}.
\newblock \showarticletitle{Improving person tracking using an inexpensive
  thermal infrared sensor}. In \bibinfo{booktitle}{{\em Proceedings of the IEEE
  Conference on Computer Vision and Pattern Recognition Workshops}}.
  \bibinfo{pages}{217--224}.
\newblock


\bibitem[\protect\citeauthoryear{Lanham}{Lanham}{2018}]%
        {lanham2018learn}
\bibfield{author}{\bibinfo{person}{Micheal Lanham}.}
  \bibinfo{year}{2018}\natexlab{}.
\newblock \bibinfo{booktitle}{{\em Learn ARCore-Fundamentals of Google ARCore:
  Learn to build augmented reality apps for Android, Unity, and the web with
  Google ARCore 1.0}}.
\newblock \bibinfo{publisher}{Packt Publishing Ltd}.
\newblock


\bibitem[\protect\citeauthoryear{Lee, Lee, and Kang}{Lee et~al\mbox{.}}{2019}]%
        {lee2019self}
\bibfield{author}{\bibinfo{person}{Junhyun Lee}, \bibinfo{person}{Inyeop Lee},
  {and} \bibinfo{person}{Jaewoo Kang}.} \bibinfo{year}{2019}\natexlab{}.
\newblock \showarticletitle{Self-attention graph pooling}. In
  \bibinfo{booktitle}{{\em International Conference on Machine Learning}}.
  PMLR, \bibinfo{pages}{3734--3743}.
\newblock


\bibitem[\protect\citeauthoryear{Li, Wang, and Gu}{Li et~al\mbox{.}}{2020}]%
        {li2020deepslam}
\bibfield{author}{\bibinfo{person}{Ruihao Li}, \bibinfo{person}{Sen Wang},
  {and} \bibinfo{person}{Dongbing Gu}.} \bibinfo{year}{2020}\natexlab{}.
\newblock \showarticletitle{Deepslam: A robust monocular slam system with
  unsupervised deep learning}.
\newblock \bibinfo{journal}{{\em IEEE Transactions on Industrial
  Electronics\/}} \bibinfo{volume}{68}, \bibinfo{number}{4}
  (\bibinfo{year}{2020}), \bibinfo{pages}{3577--3587}.
\newblock


\bibitem[\protect\citeauthoryear{L{\"o}llmann, Evers, Schmidt, Mellmann,
  Barfuss, Naylor, and Kellermann}{L{\"o}llmann et~al\mbox{.}}{2018}]%
        {lollmann2018locata}
\bibfield{author}{\bibinfo{person}{Heinrich~W L{\"o}llmann},
  \bibinfo{person}{Christine Evers}, \bibinfo{person}{Alexander Schmidt},
  \bibinfo{person}{Heinrich Mellmann}, \bibinfo{person}{Hendrik Barfuss},
  \bibinfo{person}{Patrick~A Naylor}, {and} \bibinfo{person}{Walter
  Kellermann}.} \bibinfo{year}{2018}\natexlab{}.
\newblock \showarticletitle{The LOCATA challenge data corpus for acoustic
  source localization and tracking}. In \bibinfo{booktitle}{{\em 2018 IEEE 10th
  Sensor Array and Multichannel Signal Processing Workshop (SAM)}}. IEEE,
  \bibinfo{pages}{410--414}.
\newblock


\bibitem[\protect\citeauthoryear{Lu, Saputra, Zhao, Almalioglu, de~Gusmao,
  Chen, Sun, Trigoni, and Markham}{Lu et~al\mbox{.}}{2020}]%
        {lu2020milliego}
\bibfield{author}{\bibinfo{person}{Chris~Xiaoxuan Lu}, \bibinfo{person}{Muhamad
  Risqi~U Saputra}, \bibinfo{person}{Peijun Zhao}, \bibinfo{person}{Yasin
  Almalioglu}, \bibinfo{person}{Pedro~PB de Gusmao}, \bibinfo{person}{Changhao
  Chen}, \bibinfo{person}{Ke Sun}, \bibinfo{person}{Niki Trigoni}, {and}
  \bibinfo{person}{Andrew Markham}.} \bibinfo{year}{2020}\natexlab{}.
\newblock \showarticletitle{milliEgo: single-chip mmWave radar aided egomotion
  estimation via deep sensor fusion}. In \bibinfo{booktitle}{{\em Proceedings
  of the 18th Conference on Embedded Networked Sensor Systems}}.
  \bibinfo{pages}{109--122}.
\newblock


\bibitem[\protect\citeauthoryear{Luo, Fan, and Li}{Luo et~al\mbox{.}}{2017}]%
        {luo2017indoor}
\bibfield{author}{\bibinfo{person}{Junhai Luo}, \bibinfo{person}{Liying Fan},
  {and} \bibinfo{person}{Husheng Li}.} \bibinfo{year}{2017}\natexlab{}.
\newblock \showarticletitle{Indoor positioning systems based on visible light
  communication: State of the art}.
\newblock \bibinfo{journal}{{\em IEEE Communications Surveys \& Tutorials\/}}
  \bibinfo{volume}{19}, \bibinfo{number}{4} (\bibinfo{year}{2017}),
  \bibinfo{pages}{2871--2893}.
\newblock


\bibitem[\protect\citeauthoryear{Neuhold, Bettstetter, and Molisch}{Neuhold
  et~al\mbox{.}}{2019}]%
        {neuhold2019hipr}
\bibfield{author}{\bibinfo{person}{Daniel Neuhold}, \bibinfo{person}{Christian
  Bettstetter}, {and} \bibinfo{person}{Andreas~F Molisch}.}
  \bibinfo{year}{2019}\natexlab{}.
\newblock \showarticletitle{HiPR: High-precision UWB ranging for sensor
  networks}. In \bibinfo{booktitle}{{\em Proceedings of the 22nd International
  ACM Conference on Modeling, Analysis and Simulation of Wireless and Mobile
  Systems}}. \bibinfo{pages}{103--107}.
\newblock


\bibitem[\protect\citeauthoryear{Ni, Wang, Tang, Yin, and Shen}{Ni
  et~al\mbox{.}}{2017}]%
        {ni2017accurate}
\bibfield{author}{\bibinfo{person}{Lei Ni}, \bibinfo{person}{Yuxin Wang},
  \bibinfo{person}{Haoyang Tang}, \bibinfo{person}{Zhao Yin}, {and}
  \bibinfo{person}{Yanming Shen}.} \bibinfo{year}{2017}\natexlab{}.
\newblock \showarticletitle{Accurate localization using LTE signaling data}. In
  \bibinfo{booktitle}{{\em 2017 IEEE International Conference on Computer and
  Information Technology (CIT)}}. IEEE, \bibinfo{pages}{268--273}.
\newblock


\bibitem[\protect\citeauthoryear{Palacios, Bielsa, Casari, and Widmer}{Palacios
  et~al\mbox{.}}{2018}]%
        {palacios2018communication}
\bibfield{author}{\bibinfo{person}{Joan Palacios}, \bibinfo{person}{Guillermo
  Bielsa}, \bibinfo{person}{Paolo Casari}, {and} \bibinfo{person}{Joerg
  Widmer}.} \bibinfo{year}{2018}\natexlab{}.
\newblock \showarticletitle{Communication-driven localization and mapping for
  millimeter wave networks}. In \bibinfo{booktitle}{{\em IEEE INFOCOM 2018-IEEE
  Conference on Computer Communications}}. IEEE, \bibinfo{pages}{2402--2410}.
\newblock


\bibitem[\protect\citeauthoryear{Palacios, Casari, and Widmer}{Palacios
  et~al\mbox{.}}{2017}]%
        {palacios2017jade}
\bibfield{author}{\bibinfo{person}{Joan Palacios}, \bibinfo{person}{Paolo
  Casari}, {and} \bibinfo{person}{Joerg Widmer}.}
  \bibinfo{year}{2017}\natexlab{}.
\newblock \showarticletitle{JADE: Zero-knowledge device localization and
  environment mapping for millimeter wave systems}. In \bibinfo{booktitle}{{\em
  IEEE INFOCOM 2017-IEEE Conference on Computer Communications}}. IEEE,
  \bibinfo{pages}{1--9}.
\newblock


\bibitem[\protect\citeauthoryear{Paszke, Gross, Massa, Lerer, Bradbury, Chanan,
  Killeen, Lin, Gimelshein, Antiga, et~al\mbox{.}}{Paszke
  et~al\mbox{.}}{2019}]%
        {paszke2019pytorch}
\bibfield{author}{\bibinfo{person}{Adam Paszke}, \bibinfo{person}{Sam Gross},
  \bibinfo{person}{Francisco Massa}, \bibinfo{person}{Adam Lerer},
  \bibinfo{person}{James Bradbury}, \bibinfo{person}{Gregory Chanan},
  \bibinfo{person}{Trevor Killeen}, \bibinfo{person}{Zeming Lin},
  \bibinfo{person}{Natalia Gimelshein}, \bibinfo{person}{Luca Antiga},
  {et~al\mbox{.}}} \bibinfo{year}{2019}\natexlab{}.
\newblock \showarticletitle{Pytorch: An imperative style, high-performance deep
  learning library}.
\newblock \bibinfo{journal}{{\em arXiv preprint arXiv:1912.01703\/}}
  (\bibinfo{year}{2019}).
\newblock


\bibitem[\protect\citeauthoryear{Peng, Shen, Zhang, Li, and Tan}{Peng
  et~al\mbox{.}}{2007}]%
        {chunyi2007}
\bibfield{author}{\bibinfo{person}{Chunyi Peng}, \bibinfo{person}{Guobin Shen},
  \bibinfo{person}{Yongguang Zhang}, \bibinfo{person}{Yanlin Li}, {and}
  \bibinfo{person}{Kun Tan}.} \bibinfo{year}{2007}\natexlab{}.
\newblock \showarticletitle{BeepBeep: A High Accuracy Acoustic Ranging System
  Using COTS Mobile Devices}. In \bibinfo{booktitle}{{\em Proceedings of the
  5th International Conference on Embedded Networked Sensor Systems}} {\em
  (\bibinfo{series}{SenSys '07})}. \bibinfo{publisher}{Association for
  Computing Machinery}, \bibinfo{address}{New York, NY, USA},
  \bibinfo{pages}{1–14}.
\newblock
\showISBNx{9781595937636}
\showDOI{%
\url{https://doi.org/10.1145/1322263.1322265}}


\bibitem[\protect\citeauthoryear{Ragot, Khemmar, Pokala, Rossi, and
  Ertaud}{Ragot et~al\mbox{.}}{2019}]%
        {ragot2019benchmark}
\bibfield{author}{\bibinfo{person}{Nicolas Ragot}, \bibinfo{person}{Redouane
  Khemmar}, \bibinfo{person}{Adithya Pokala}, \bibinfo{person}{Romain Rossi},
  {and} \bibinfo{person}{Jean-Yves Ertaud}.} \bibinfo{year}{2019}\natexlab{}.
\newblock \showarticletitle{Benchmark of visual slam algorithms: Orb-slam2 vs
  rtab-map}. In \bibinfo{booktitle}{{\em 2019 Eighth International Conference
  on Emerging Security Technologies (EST)}}. IEEE, \bibinfo{pages}{1--6}.
\newblock


\bibitem[\protect\citeauthoryear{Read, Martino, Olmos, and Luengo}{Read
  et~al\mbox{.}}{2015}]%
        {read2015scalable}
\bibfield{author}{\bibinfo{person}{Jesse Read}, \bibinfo{person}{Luca Martino},
  \bibinfo{person}{Pablo~M Olmos}, {and} \bibinfo{person}{David Luengo}.}
  \bibinfo{year}{2015}\natexlab{}.
\newblock \showarticletitle{Scalable multi-output label prediction: From
  classifier chains to classifier trellises}.
\newblock \bibinfo{journal}{{\em Pattern Recognition\/}} \bibinfo{volume}{48},
  \bibinfo{number}{6} (\bibinfo{year}{2015}), \bibinfo{pages}{2096--2109}.
\newblock


\bibitem[\protect\citeauthoryear{Rublee, Rabaud, Konolige, and Bradski}{Rublee
  et~al\mbox{.}}{2011}]%
        {rublee2011orb}
\bibfield{author}{\bibinfo{person}{Ethan Rublee}, \bibinfo{person}{Vincent
  Rabaud}, \bibinfo{person}{Kurt Konolige}, {and} \bibinfo{person}{Gary
  Bradski}.} \bibinfo{year}{2011}\natexlab{}.
\newblock \showarticletitle{ORB: An efficient alternative to SIFT or SURF}. In
  \bibinfo{booktitle}{{\em 2011 International conference on computer vision}}.
  Ieee, \bibinfo{pages}{2564--2571}.
\newblock


\bibitem[\protect\citeauthoryear{Ruiz and Granja}{Ruiz and Granja}{2017}]%
        {ruiz2017comparing}
\bibfield{author}{\bibinfo{person}{Antonio Ram{\'o}n~Jim{\'e}nez Ruiz} {and}
  \bibinfo{person}{Fernando~Seco Granja}.} \bibinfo{year}{2017}\natexlab{}.
\newblock \showarticletitle{Comparing ubisense, bespoon, and decawave uwb
  location systems: Indoor performance analysis}.
\newblock \bibinfo{journal}{{\em IEEE Transactions on instrumentation and
  Measurement\/}} \bibinfo{volume}{66}, \bibinfo{number}{8}
  (\bibinfo{year}{2017}), \bibinfo{pages}{2106--2117}.
\newblock


\bibitem[\protect\citeauthoryear{Saputra, de~Gusmao, Lu, Almalioglu, Rosa,
  Chen, Wahlstr{\"o}m, Wang, Markham, and Trigoni}{Saputra
  et~al\mbox{.}}{2020}]%
        {saputra2020deeptio}
\bibfield{author}{\bibinfo{person}{Muhamad Risqi~U Saputra},
  \bibinfo{person}{Pedro~PB de Gusmao}, \bibinfo{person}{Chris~Xiaoxuan Lu},
  \bibinfo{person}{Yasin Almalioglu}, \bibinfo{person}{Stefano Rosa},
  \bibinfo{person}{Changhao Chen}, \bibinfo{person}{Johan Wahlstr{\"o}m},
  \bibinfo{person}{Wei Wang}, \bibinfo{person}{Andrew Markham}, {and}
  \bibinfo{person}{Niki Trigoni}.} \bibinfo{year}{2020}\natexlab{}.
\newblock \showarticletitle{Deeptio: A deep thermal-inertial odometry with
  visual hallucination}.
\newblock \bibinfo{journal}{{\em IEEE Robotics and Automation Letters\/}}
  \bibinfo{volume}{5}, \bibinfo{number}{2} (\bibinfo{year}{2020}),
  \bibinfo{pages}{1672--1679}.
\newblock


\bibitem[\protect\citeauthoryear{Saputra, Markham, and Trigoni}{Saputra
  et~al\mbox{.}}{2018}]%
        {saputra2018visual}
\bibfield{author}{\bibinfo{person}{Muhamad Risqi~U Saputra},
  \bibinfo{person}{Andrew Markham}, {and} \bibinfo{person}{Niki Trigoni}.}
  \bibinfo{year}{2018}\natexlab{}.
\newblock \showarticletitle{Visual SLAM and structure from motion in dynamic
  environments: A survey}.
\newblock \bibinfo{journal}{{\em ACM Computing Surveys (CSUR)\/}}
  \bibinfo{volume}{51}, \bibinfo{number}{2} (\bibinfo{year}{2018}),
  \bibinfo{pages}{1--36}.
\newblock


\bibitem[\protect\citeauthoryear{Solin, Cortes, Rahtu, and Kannala}{Solin
  et~al\mbox{.}}{2018}]%
        {solin2018inertial}
\bibfield{author}{\bibinfo{person}{Arno Solin}, \bibinfo{person}{Santiago
  Cortes}, \bibinfo{person}{Esa Rahtu}, {and} \bibinfo{person}{Juho Kannala}.}
  \bibinfo{year}{2018}\natexlab{}.
\newblock \showarticletitle{Inertial odometry on handheld smartphones}. In
  \bibinfo{booktitle}{{\em 2018 21st International Conference on Information
  Fusion (FUSION)}}. IEEE, \bibinfo{pages}{1--5}.
\newblock


\bibitem[\protect\citeauthoryear{Song, Chandraker, and Guest}{Song
  et~al\mbox{.}}{2015}]%
        {song2015high}
\bibfield{author}{\bibinfo{person}{Shiyu Song}, \bibinfo{person}{Manmohan
  Chandraker}, {and} \bibinfo{person}{Clark~C Guest}.}
  \bibinfo{year}{2015}\natexlab{}.
\newblock \showarticletitle{High accuracy monocular SFM and scale correction
  for autonomous driving}.
\newblock \bibinfo{journal}{{\em IEEE transactions on pattern analysis and
  machine intelligence\/}} \bibinfo{volume}{38}, \bibinfo{number}{4}
  (\bibinfo{year}{2015}), \bibinfo{pages}{730--743}.
\newblock


\bibitem[\protect\citeauthoryear{Sturm, Engelhard, Endres, Burgard, and
  Cremers}{Sturm et~al\mbox{.}}{2012}]%
        {sturm2012benchmark}
\bibfield{author}{\bibinfo{person}{J{\"u}rgen Sturm}, \bibinfo{person}{Nikolas
  Engelhard}, \bibinfo{person}{Felix Endres}, \bibinfo{person}{Wolfram
  Burgard}, {and} \bibinfo{person}{Daniel Cremers}.}
  \bibinfo{year}{2012}\natexlab{}.
\newblock \showarticletitle{A benchmark for the evaluation of RGB-D SLAM
  systems}. In \bibinfo{booktitle}{{\em 2012 IEEE/RSJ International Conference
  on Intelligent Robots and Systems}}. IEEE, \bibinfo{pages}{573--580}.
\newblock


\bibitem[\protect\citeauthoryear{Taketomi, Uchiyama, and Ikeda}{Taketomi
  et~al\mbox{.}}{2017}]%
        {taketomi2017visual}
\bibfield{author}{\bibinfo{person}{Takafumi Taketomi}, \bibinfo{person}{Hideaki
  Uchiyama}, {and} \bibinfo{person}{Sei Ikeda}.}
  \bibinfo{year}{2017}\natexlab{}.
\newblock \showarticletitle{Visual SLAM algorithms: a survey from 2010 to
  2016}.
\newblock \bibinfo{journal}{{\em IPSJ Transactions on Computer Vision and
  Applications\/}} \bibinfo{volume}{9}, \bibinfo{number}{1}
  (\bibinfo{year}{2017}), \bibinfo{pages}{1--11}.
\newblock


\bibitem[\protect\citeauthoryear{Tarzia, Dinda, Dick, and Memik}{Tarzia
  et~al\mbox{.}}{2011}]%
        {tarzia2011indoor}
\bibfield{author}{\bibinfo{person}{Stephen~P Tarzia}, \bibinfo{person}{Peter~A
  Dinda}, \bibinfo{person}{Robert~P Dick}, {and} \bibinfo{person}{Gokhan
  Memik}.} \bibinfo{year}{2011}\natexlab{}.
\newblock \showarticletitle{Indoor localization without infrastructure using
  the acoustic background spectrum}. In \bibinfo{booktitle}{{\em Proceedings of
  the 9th international conference on Mobile systems, applications, and
  services}}. \bibinfo{pages}{155--168}.
\newblock


\bibitem[\protect\citeauthoryear{Vaswani, Shazeer, Parmar, Uszkoreit, Jones,
  Gomez, Kaiser, and Polosukhin}{Vaswani et~al\mbox{.}}{2017}]%
        {vaswani2017attention}
\bibfield{author}{\bibinfo{person}{Ashish Vaswani}, \bibinfo{person}{Noam
  Shazeer}, \bibinfo{person}{Niki Parmar}, \bibinfo{person}{Jakob Uszkoreit},
  \bibinfo{person}{Llion Jones}, \bibinfo{person}{Aidan~N Gomez},
  \bibinfo{person}{Lukasz Kaiser}, {and} \bibinfo{person}{Illia Polosukhin}.}
  \bibinfo{year}{2017}\natexlab{}.
\newblock \showarticletitle{Attention is All you Need}. In
  \bibinfo{booktitle}{{\em NIPS}}.
\newblock


\bibitem[\protect\citeauthoryear{Venkatnarayan and Shahzad}{Venkatnarayan and
  Shahzad}{2019a}]%
        {venkatnarayan2019enhancing}
\bibfield{author}{\bibinfo{person}{Raghav~H Venkatnarayan} {and}
  \bibinfo{person}{Muhammad Shahzad}.} \bibinfo{year}{2019}\natexlab{a}.
\newblock \showarticletitle{Enhancing indoor inertial odometry with wifi}.
\newblock \bibinfo{journal}{{\em Proceedings of the ACM on Interactive, Mobile,
  Wearable and Ubiquitous Technologies\/}} \bibinfo{volume}{3},
  \bibinfo{number}{2} (\bibinfo{year}{2019}), \bibinfo{pages}{1--27}.
\newblock


\bibitem[\protect\citeauthoryear{Venkatnarayan and Shahzad}{Venkatnarayan and
  Shahzad}{2019b}]%
        {venkatnarayan2019measuring}
\bibfield{author}{\bibinfo{person}{Raghav~Hampapur Venkatnarayan} {and}
  \bibinfo{person}{Muhammad Shahzad}.} \bibinfo{year}{2019}\natexlab{b}.
\newblock \showarticletitle{Measuring Distance Traveled by an Object using
  WiFi-CSI and IMU Fusion}. In \bibinfo{booktitle}{{\em 2019 IEEE 27th
  International Conference on Network Protocols (ICNP)}}. IEEE,
  \bibinfo{pages}{1--2}.
\newblock


\bibitem[\protect\citeauthoryear{Wan and Van Der~Merwe}{Wan and Van
  Der~Merwe}{2000}]%
        {wan2000unscented}
\bibfield{author}{\bibinfo{person}{Eric~A Wan} {and} \bibinfo{person}{Rudolph
  Van Der~Merwe}.} \bibinfo{year}{2000}\natexlab{}.
\newblock \showarticletitle{The unscented Kalman filter for nonlinear
  estimation}. In \bibinfo{booktitle}{{\em Proceedings of the IEEE 2000
  Adaptive Systems for Signal Processing, Communications, and Control Symposium
  (Cat. No. 00EX373)}}. Ieee, \bibinfo{pages}{153--158}.
\newblock


\bibitem[\protect\citeauthoryear{Wang, Girshick, Gupta, and He}{Wang
  et~al\mbox{.}}{2018}]%
        {wang2018non}
\bibfield{author}{\bibinfo{person}{Xiaolong Wang}, \bibinfo{person}{Ross
  Girshick}, \bibinfo{person}{Abhinav Gupta}, {and} \bibinfo{person}{Kaiming
  He}.} \bibinfo{year}{2018}\natexlab{}.
\newblock \showarticletitle{Non-local neural networks}. In
  \bibinfo{booktitle}{{\em Proceedings of the IEEE conference on computer
  vision and pattern recognition}}. \bibinfo{pages}{7794--7803}.
\newblock


\bibitem[\protect\citeauthoryear{Wang and Li}{Wang and Li}{2017}]%
        {wang2017imu}
\bibfield{author}{\bibinfo{person}{Yan Wang} {and} \bibinfo{person}{Xin Li}.}
  \bibinfo{year}{2017}\natexlab{}.
\newblock \showarticletitle{The IMU/UWB fusion positioning algorithm based on a
  particle filter}.
\newblock \bibinfo{journal}{{\em ISPRS International Journal of
  Geo-Information\/}} \bibinfo{volume}{6}, \bibinfo{number}{8}
  (\bibinfo{year}{2017}), \bibinfo{pages}{235}.
\newblock


\bibitem[\protect\citeauthoryear{Zeng, Pathak, Yang, and Mohapatra}{Zeng
  et~al\mbox{.}}{2016}]%
        {zeng2016human}
\bibfield{author}{\bibinfo{person}{Yunze Zeng}, \bibinfo{person}{Parth~H
  Pathak}, \bibinfo{person}{Zhicheng Yang}, {and} \bibinfo{person}{Prasant
  Mohapatra}.} \bibinfo{year}{2016}\natexlab{}.
\newblock \showarticletitle{Human tracking and activity monitoring using 60 GHz
  mmWave}. In \bibinfo{booktitle}{{\em 2016 15th ACM/IEEE International
  Conference on Information Processing in Sensor Networks (IPSN)}}. IEEE,
  \bibinfo{pages}{1--2}.
\newblock


\bibitem[\protect\citeauthoryear{Zhang and Singh}{Zhang and Singh}{2014}]%
        {zhang2014loam}
\bibfield{author}{\bibinfo{person}{Ji Zhang} {and} \bibinfo{person}{Sanjiv
  Singh}.} \bibinfo{year}{2014}\natexlab{}.
\newblock \showarticletitle{LOAM: Lidar Odometry and Mapping in Real-time.}. In
  \bibinfo{booktitle}{{\em Robotics: Science and Systems}},
  Vol.~\bibinfo{volume}{2}.
\newblock


\end{thebibliography}

\normalsize
\end{document}